\documentclass[10pt,twocolumn,letterpaper]{article}

\usepackage{cvpr}
\usepackage{times}
\usepackage{epsfig}
\usepackage{graphicx}
\usepackage{amsmath}
\usepackage{amssymb}
\usepackage{amsbsy}
\usepackage{authblk}
\usepackage{mathabx}

\makeatletter
\def\ps@myheadings{%
	\let\@oddfoot\@empty\let\@evenfoot\@empty
	\def\@evenhead{\thepage\hfil\slshape\leftmark}%
	\def\@oddhead{{\slshape\rightmark}\hfil\thepage}%
	\let\@mkboth\@gobbletwo
	\let\sectionmark\@gobble
	\let\subsectionmark\@gobble
}
\if@titlepage
\renewcommand\maketitle{\begin{titlepage}%
		\let\footnotesize\small
		\let\footnoterule\relax
		\let \footnote \thanks
		\null\vfil
		\vskip 60\p@
		\begin{center}%
			{\LARGE \@title \par}%
			\vskip 3em%
			{\large
				\lineskip .75em%
				\begin{tabular}[t]{c}%
					\@author
				\end{tabular}\par}%
			\vskip 1.5em%
			{\large \@date \par}%       % Set date in \large size.
		\end{center}\par
		\@thanks
		\vfil\null
	\end{titlepage}%
	\setcounter{footnote}{0}%
}
\else
\renewcommand\maketitle{\par
	\begingroup
	\renewcommand\thefootnote{\@fnsymbol\c@footnote}%
	\def\@makefnmark{\rlap{\@textsuperscript{\normalfont\@thefnmark}}}%
	\long\def\@makefntext##1{\parindent 1em\noindent
		\hb@xt@1.8em{%
			\hss\@textsuperscript{\normalfont\@thefnmark}}##1}%
	\if@twocolumn
	\ifnum \col@number=\@ne
	\@maketitle
	\else
	\twocolumn[\@maketitle]%
	\fi
	\else
	\newpage
	\global\@topnum\z@   % Prevents figures from going at top of page.
	\@maketitle
	\fi
	\thispagestyle{plain}\@thanks
	\endgroup
	\setcounter{footnote}{0}%
}
\makeatother

\usepackage{booktabs}
\usepackage{multirow}
\usepackage[normalem]{ulem}

\usepackage{tabularx}
\usepackage{makecell}

\usepackage{subfig}
\usepackage{float}
\usepackage{array}
\newcolumntype{P}[1]{>{\centering\arraybackslash}p{#1}}
\usepackage{nicefrac}
\usepackage{soul}
\usepackage[sort,nocompress]{cite}
\usepackage{xcolor}
\newcommand{\specialcell}[2][c]{%
	\begin{tabular}[#1]{@{}c@{}}#2\end{tabular}}
\newcolumntype{C}[1]{>{\centering\arraybackslash}m{#1}}

\definecolor{DarkGreen}{rgb}{0.0, 0.2, 0.0}

\newcommand{\comment}[1]{}

\usepackage[pagebackref=true,breaklinks=true,letterpaper=true,colorlinks,bookmarks=false]{hyperref}

\newcommand{\datasetname}{HO-3D\xspace}

\DeclareMathOperator{\argmin}{arg\,min}

\cvprfinalcopy % *** Uncomment this line for the final submission

\def\cvprPaperID{2206} % *** Enter the CVPR Paper ID here

\ifcvprfinal\fi
\begin{document}

%%%%%%%%% TITLE
\title{HOnnotate: A method for 3D Annotation of Hand and Object Poses}

  %% \bolddatasetname: A Multi-User, Multi-Object Dataset\\ for Joint 3D Hand-Object Pose Estimation}

%\author{First Author\\
%Institution1\\
%Institution1 address\\
%{\tt\small firstauthor@i1.org}
%% For a paper whose authors are all at the same institution,
%% omit the following lines up until the closing ``}''.
%% Additional authors and addresses can be added with ``\and'',
%% just like the second author.
%% To save space, use either the email address or home page, not both
%\and
%Second Author\\
%Institution2\\
%First line of institution2 address\\
%{\tt\small secondauthor@i2.org}
%}

\author[1]{Shreyas Hampali}
\author[1]{Mahdi Rad}
\author[1]{Markus Oberweger}
\author[2,1]{Vincent Lepetit}
\affil[1]{Institute for Computer Graphics and Vision, Graz University of Technology, Austria}
\affil[2]{LIGM, Ecole des Ponts, Univ Gustave Eiffel, CNRS, Marne-la-Vallée, France}
\affil[ ]{\textit {\{hampali, rad, oberweger, lepetit\}@icg.tugraz.at}\endgraf \\
	{\fontsize{10}{10}\selectfont Project Page:  \href{https://www.tugraz.at/index.php?id=40231}{https://www.tugraz.at/index.php?id=40231}}}

\maketitle
%\thispagestyle{empty}

%%%%%%%%% ABSTRACT
\begin{abstract}

We propose a method for annotating images  of a hand manipulating an object with
the 3D poses  of both the hand  and the object, together with  a dataset created
using this method.  Our motivation is  the current lack of annotated real images
for this problem,  as estimating the 3D poses is  challenging, mostly because of
the  mutual  occlusions  between  the  hand and  the  object.   To  tackle  this
challenge, we capture  sequences with one or several RGB-D  cameras and jointly
optimize   the   3D    hand   and   object   poses   over    all   the   frames
\emph{simultaneously}.  This  method allows us to  \emph{automatically} annotate
each  frame  with  accurate  estimates   of  the  poses,  despite  large  mutual
occlusions.  With  this method,  we created  \datasetname, the  first markerless
dataset  of color  images with  3D annotations  for both the hand and  object.  This
dataset is  currently made of  77,558 frames, 68  sequences, 10 persons,  and 10
objects.   Using our  dataset, we  develop a  single RGB  image-based  method to
predict the hand pose when interacting  with objects under severe occlusions and
show it generalizes to objects not seen in the dataset.

\end{abstract}

%------------------------------------------------------------------------

\section{Introduction}

%% -*- mode: latex; mode: flyspell -*-

\begin{figure*}[ht]
  \begin{center}
    \begin{tabular}{ccccc|cc}
      \includegraphics[width=0.14\linewidth]{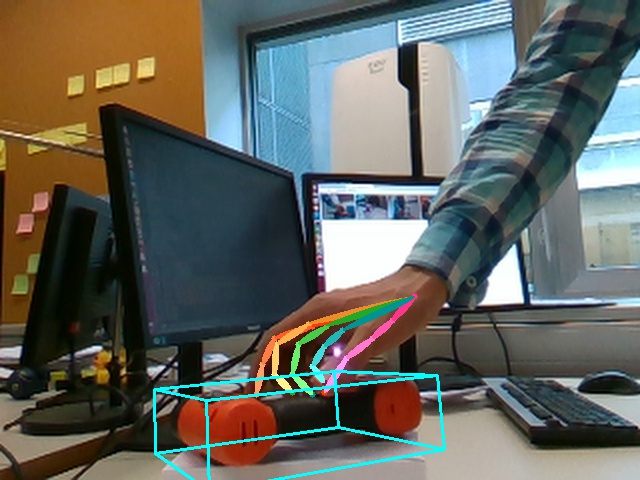} & \hspace{-4mm}
      \includegraphics[width=0.14\linewidth]{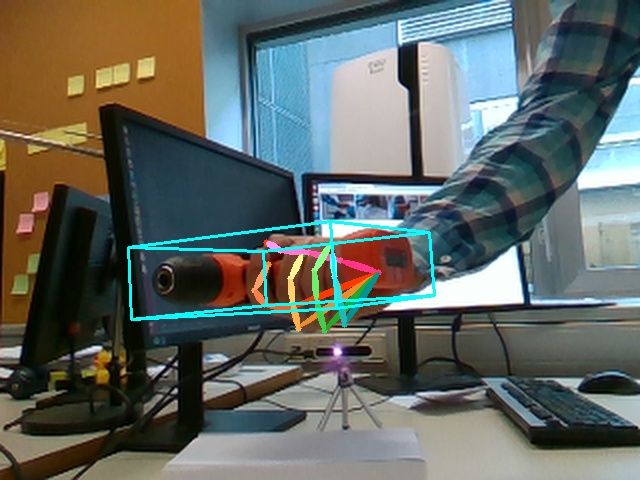} & \hspace{-4mm}
      \includegraphics[width=0.14\linewidth]{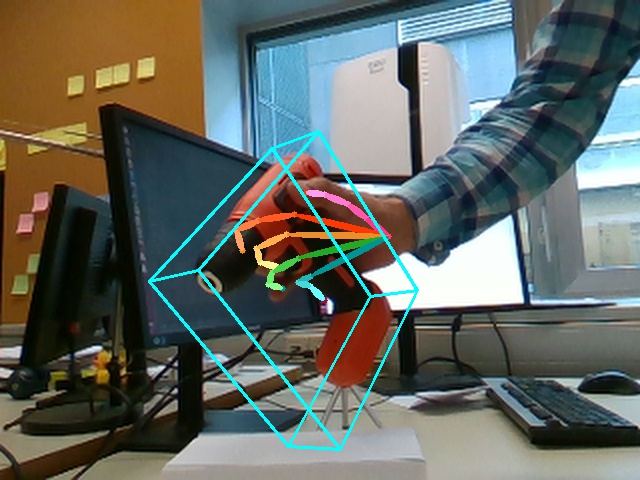} & \hspace{-4mm}
      \includegraphics[width=0.14\linewidth]{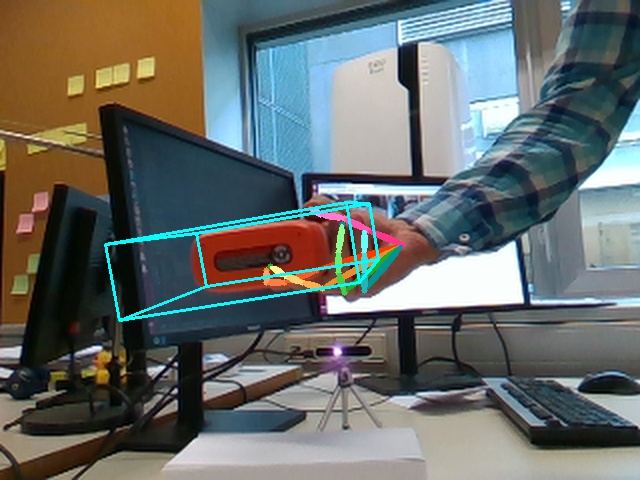} & \hspace{-4mm}
      \includegraphics[width=0.14\linewidth]{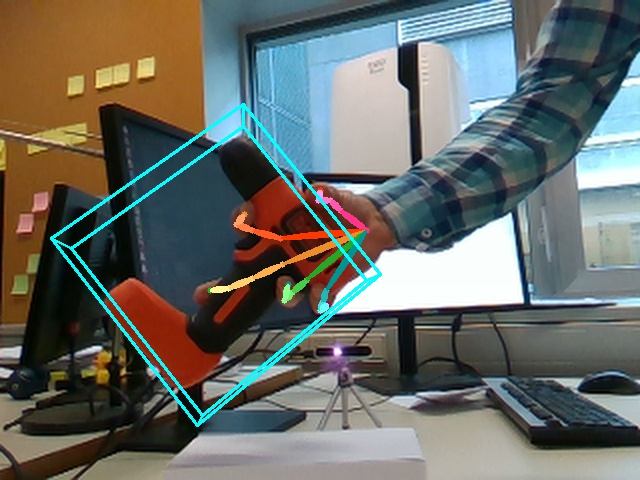} & \hspace{-1mm}

      \begin{picture}(70,45)
        \put(0,10){\includegraphics[trim=0cm 1cm 0cm 1cm, clip=true,width=0.11\linewidth]{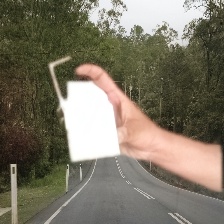}}
        \put(20,0){\cite{zimmermann2019freihand}}
      \end{picture} & \hspace{-9mm}

      \begin{picture}(70,45)
        \put(0,10){\includegraphics[width=0.11\linewidth]{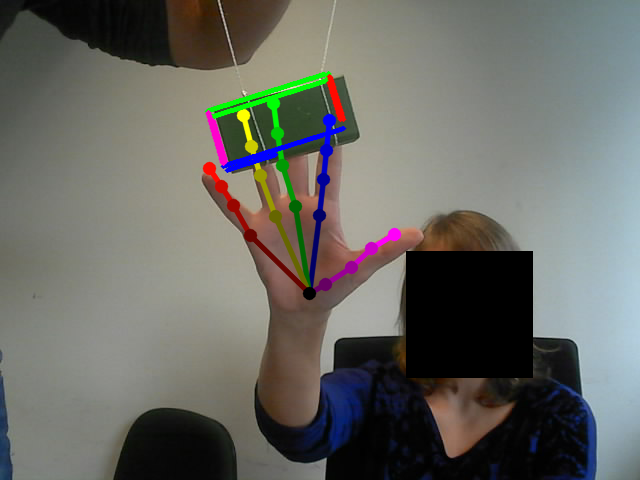}}
        \put(20,0){\cite{RealtimeHO_ECCV2016}}
      \end{picture} \\

      % &  &  & & & \cite{zimmermann2019freihand} & \cite{RealtimeHO_ECCV2016} \\
      %      \includegraphics[width=0.15\linewidth]{figures/ours_mask_0129.jpg} &
      %      \includegraphics[width=0.15\linewidth]{figures/example_icvl_col_0020.jpeg} \\
      \includegraphics[width=0.14\linewidth]{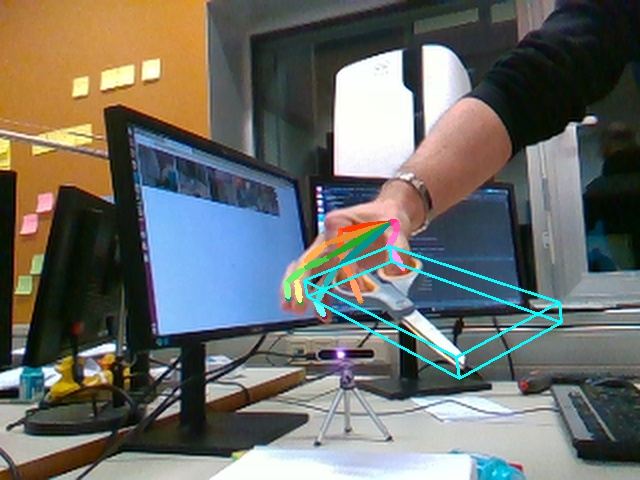} & \hspace{-4mm}
      \includegraphics[width=0.14\linewidth]{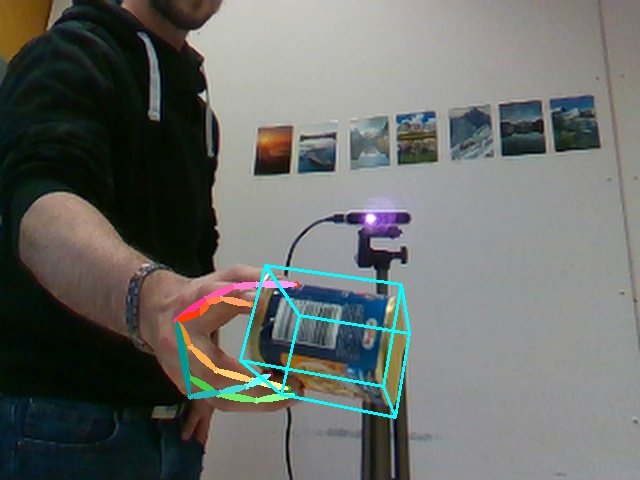} & \hspace{-4mm}
      \includegraphics[width=0.14\linewidth]{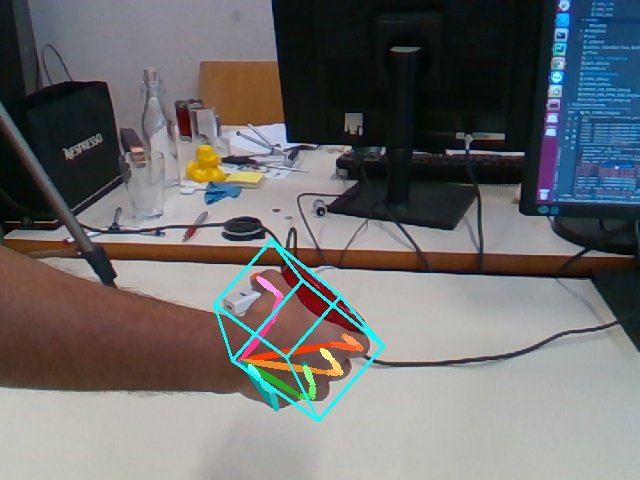} & \hspace{-4mm}
      \includegraphics[width=0.14\linewidth]{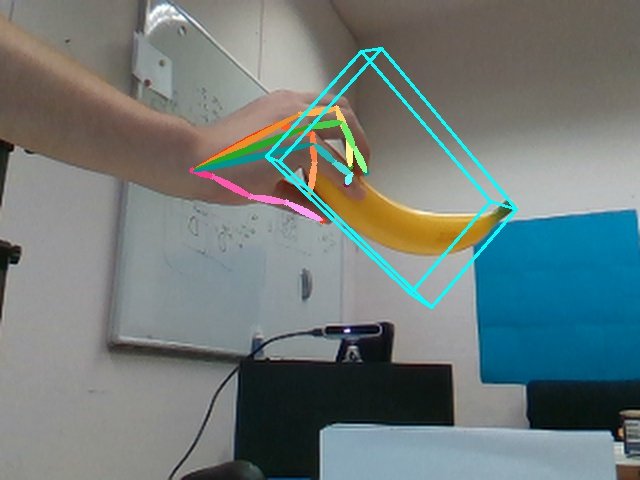} & \hspace{-4mm}
      \includegraphics[width=0.14\linewidth]{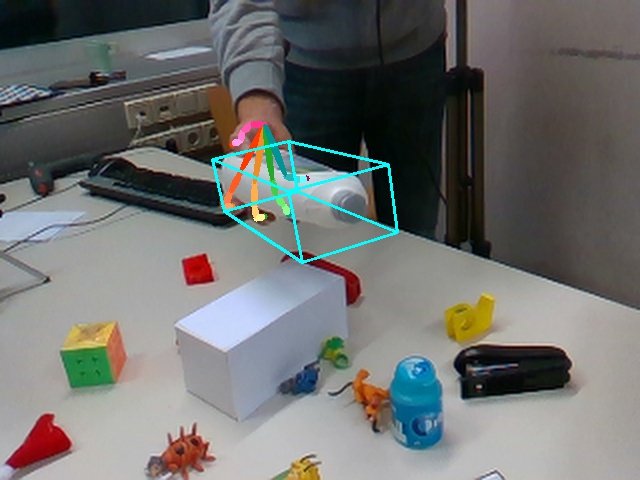} & \hspace{-1mm}

      \begin{picture}(70,45)
        \put(0,10){\includegraphics[trim=0cm 1.3cm 0cm 1.35cm, clip=true,width=0.11\linewidth]{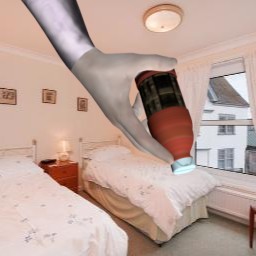}}
        \put(20,0){\cite{hasson2019learning}}
      \end{picture} & \hspace{-9mm}

      \begin{picture}(70,45)
        \put(0,10){\includegraphics[trim=4cm 0cm 10cm 0cm, clip=true,width=0.11\linewidth]{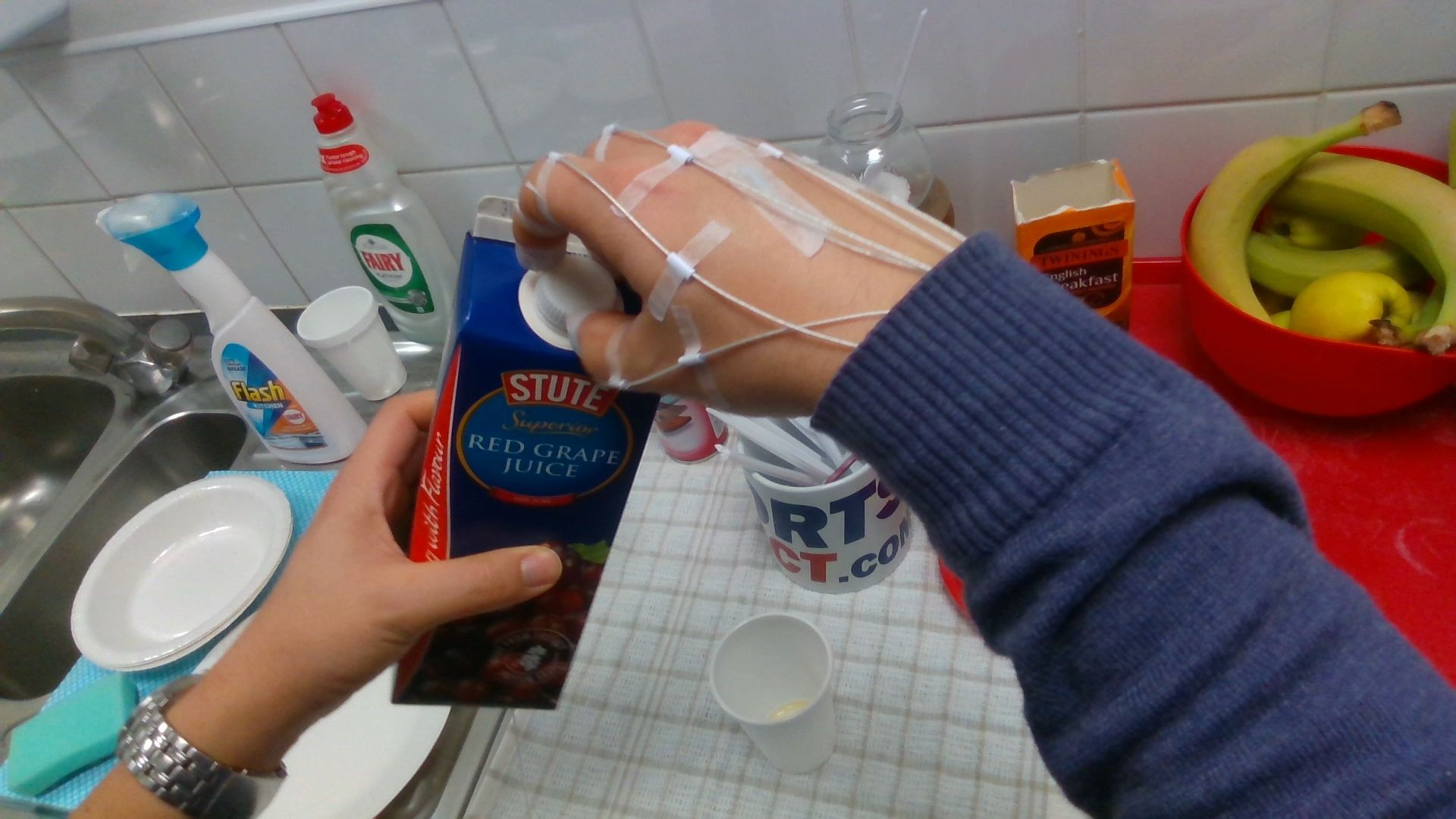}}
        \put(20,0){\cite{garcia2018first}}
      \end{picture} \\

      % &  & & &  & \cite{hasson2019learning} & \cite{garcia2018first} \\
      \multicolumn{5}{c|}{Our proposed \datasetname dataset} & \multicolumn{2}{c}{Existing datasets}
    \end{tabular}
  \end{center}\vspace{-0.5cm}
  \caption{\label{fig:one} We introduce a method for labelling real images
    of hand+object  interaction with the 3D  poses of the hand  and of the
    object.  With this method, we  automatically created a dataset made of
    more than 75,000 frames, 10  different objects and 10 different users.
    In  comparison, existing  datasets  have several  limitations: The  3D
    objects are very simple, the  interaction is not realistic, the images
    are synthetic, corrupted by sensors, and/or the number
    of samples is limited. More illustrations of annotations in our
    dataset are shown in the supplementary material.}
  \vspace{-2mm}
\end{figure*}

Methods for 3D pose estimation of  rigid objects and hands from monocular images
have  made significant  progress recently,  thanks  to the  development of  Deep
Learning, and the creation of large datasets  or the use of synthetic images for
training~\cite{sundermeyer2018implicit,Rad17,Zimmermann2017,Oberweger17,mueller2017real,Yuan2017}.
However, these recent  methods still fail when a hand  interacts with an object,
mostly  because of  large  mutual occlusions,  and of  the  absence of  datasets
specific to 3D pose estimation  for hand+object interaction. Breaking this limit
is highly desirable though, as 3D hand  and object poses would be very useful in
augmented reality  applications, or  for learning-by-imitation in  robotics, for
example.

Several  pioneer  works  have  already considered  this  problem,  sometimes  with
impressive
success~\cite{RealtimeHO_ECCV2016,kyriazis2013physically,Tzionas2016}.     These
works typically  rely on  tracking algorithms  to exploit  temporal constraints,
often also considering physical constraints between  the hand and the object to
improve the  pose estimates.   While these  temporal and  physical constraints
remain relevant,  we would like  to also benefit  from the power  of data-driven
methods for 3D  hand+object pose estimation from a single  image. Being able to
estimate these poses  from a single frame would avoid  manual initialization and
drift of  tracking algorithms.  A  data-driven approach, however,  requires real
or synthetic images annotated  with the  3D poses of  the object and  the hand.
Unfortunately,  creating annotated  data for  the hand+object
problem is very  challenging.  Both common options for  creating 3D annotations,
annotating  real  images  and  generating synthetic  images,  raise  challenging
problems.

{\bf Annotating  real images.}   One can  rely on  some algorithm  for automated
annotation,   as   was  done   for   current   benchmarks   in  3D   hand   pose
estimation~\cite{Tang2014,Qian2014realtime,Sun2015,Yuan2017,zimmermann2019freihand},
where  the  ``ground  truth''  annotations are  obtained  automatically  with  a
tracking algorithm.   Though these annotations are noisy, they are  usually taken for granted  and used
for training and evaluation~\cite{Oberweger16}.  Another
approach    is     to    use    sensors     attached    to    the     hand    as
in~\cite{garcia2018first}~(bottom  right  image  of  Fig.~\ref{fig:one}).   This
directly provides the 3D poses, however,  the sensors are visible in the images,
and thus  bias learning methods.  Significant  effort is still  required in
developing algorithms for automated annotation of real images.

{\bf Generating synthetic images.  }   Relying on synthetic images is appealing,
as the  3D poses are known  perfectly.  Realistic rendering and  domain transfer
can    be    used    to    train     3D    pose    estimation    on    synthetic
images~\cite{Mueller2018,Rad2018b,Zimmermann2017}.      Generating    physically
correct       grasps      is       possible~\cite{Miller04},      as       shown
in~\cite{hasson2019learning}, but complex manipulation is difficult to simulate.
However, real  images with accurate 3D  annotations would still be  needed to
evaluate the generalizability of the method to real data.

We therefore  propose a method  to automatically  annotate real images  of hands
grasping objects  with their  3D poses.   Our method works  with a  single RGB-D
camera, but  can exploit  more cameras  if available  for better  robustness and
accuracy.  The  single-camera setup  works under the  assumption that  the grasp
pose  varies marginally  over the  sequence; the  multi-camera setup  can handle
complex  hand+object  interaction  scenarios.   Instead of  tracking  the  poses
frame-by-frame, our  method optimizes jointly all  the 3D poses of  the hand and
the  object over  the sequence.   As  our evaluations  show, this  allows us  to
exploit temporal consistency in a stronger way than a tracking algorithm.  Using
differentiable  rendering,  we can  optimize  a  complex objective  function  by
exploiting the  new powerful gradient  descent methods originally  developed for
Deep Learning~\cite{Kingma15}.  We see this approach as the equivalent of bundle
adjustment for SLAM algorithms, where we track objects instead of points.

We rely  on the MANO hand  model~\cite{romero2017embodied}, and the 3D  model of
the objects.   We use objects  from the YCB-Video dataset~\cite{posecnn2018},  as they
have various shapes and materials,  and can be bought online~\cite{ycbonline} by
researchers interested in performing their own experiments.  Being able to use a
single camera also enables easier expansion  of the dataset by other researchers
with a larger  variety of objects and grasping poses  as multi-camera capture is
often complex to setup.

Using  our method,  we created  a dataset,  depicted in  Fig.~\ref{fig:one}, which  we call  \datasetname. In  addition,  we used  this dataset  to learn  to
predict  from  a  single RGB  image  the  3D  pose  of a  hand  manipulating  an
object. More exactly, we train a Deep  Network to predict the 2D joint locations
of the hand along with the joint direction  vectors and lift them to 3D by fitting
a MANO model  to these predictions.  This  validates the fact that  the 3D poses
estimated by our annotation method can  actually be used in a data-driven method
for hand pose estimation.  By comparing  with an existing method for hand+object
pose   estimation~\cite{hasson2019learning}   that   directly   estimates   MANO
parameters, we show  that predicting 2D keypoints and lifting  them to 3D performs
more accurately.

\section{Related Work}

%% -*- mode: latex; mode: flyspell -*-

The literature on hand and object  pose estimation is extremely broad, and we
review only some works here.

\subsection{3D Object Pose Estimation}

% I removed some citations to our work, there was many :)
% We can still add them back later :)

Estimating the  3D pose of  an object from  a single frame  is still one  of the
fundamental problems of Computer Vision. Some  methods are now robust to partial
occlusions~\cite{Oberweger2018,peng2019pvnet,hu2019segpose}, but many works rely on RGB-D data to
handle this problem~\cite{Buch2017,Mitash2017,Zhang2017,Kehl16},  by fitting the
3D object model to depth data. This  can fail when a hand grasps the object,
since the surface of the hand can be mistaken for the surface of the object.

%%  In order to be  robust to occlusion, \cite{Oberweger2018} estimates
%% the 2D  projections by averaging  heatmaps predicted from image  patches. 

%%%%%%%%%%%%%%%%%%%%%%%%%%%%%%%%%%%%%%%%%%%%%%%%%%%%%%%%%%%%%%%%%%%%%%%%%%%%%%%%

\subsection{3D Hand Pose Estimation}

Single image hand pose estimation is also a very popular problem in Computer Vision, and approaches can be divided into discriminative and generative methods. Discriminative approaches directly predict the joint locations from RGB or RGB-D images.  Recent works based on Deep Networks \cite{Tompson14b,Xu16,Neverova17,Oberweger17,mueller2017real,Zimmermann2017,ge2018hand} show remarkable performance, compared to early works based on Random Forests such as \cite{keskin2012hand}.  However, discriminative methods perform poorly in case of partial occlusion.

Generative approaches take advantage of a hand model and its kinematic structure to generate hand pose hypotheses that are physically plausible~\cite{sridhar2013interactive,Qian2014realtime,sharp2015accurate,Tzionas2014capturing,melax2013dynamics,de2011model,ye2016spatial}.  \cite{Mueller2018,Panteleris2018} predict 2D joint locations and then lift them to 3D.  Generative approaches are usually accurate and can be made robust to partial occlusions.  They typically rely on some pose prior, which may require manual initialization or result in drift when tracking.

Our work is  related to both discriminative and generative  approaches: we use a
generative approach within a global  optimization framework to generate the pose
annotations,  and  use  a  discriminative  method  to  initialize  this  complex
optimization.  \cite{Wan_2019_CVPR}  also combines generative  and discriminative
approaches to train  a network in a self-supervised setting.   However, they only
consider hands.   We also train  a discriminative  method using our  dataset, to
predict the hand poses which are robust to occlusions from interacting objects.

%%%%%%%%%%%%%%%%%%%%%%%%%%%%%%%%%%%%%%%%%%%%%%%%%%%%%%%%%%%%%%%%%%%%%%%%%%%%%%%%

\subsection{Synthetic Images for 3D Pose Estimation}

Being able to train discriminative methods on synthetic data is valuable as it is difficult to acquire annotations for real images~\cite{Zimmermann2017}.  \cite{hasson2019learning,Rad2018b} show that because of the domain gap between synthetic and real images, training on synthetic images only results in sub-optimal performance.  A GAN method was used in \cite{Mueller2018} to make synthetic images of hands more realistic.  While using synthetic images remains appealing for many problems, creating the virtual scenes can be expensive and time-consuming.  Generating animated realistic hand grasps of various objects, as it would be required to solve the problem considered in this paper remains challenging. Being able to use real sequences for training has thus also its advantages. Moreover, evaluation should be performed on real images.

%% Being able to  train discriminative methods on synthetic data  is valuable as it
%% is  difficult  to  acquire annotations  for  real  images~\cite{Zimmermann2017}.
%% \cite{hasson2019learning,Rad2018b} show  that because of the  domain gap between
%% synthetic  and  real  images,  training  on synthetic  images  only  results  in
%% sub-optimal performance. A sophisticated GAN  is used by \cite{Mueller2018}, but
%% this still  requires renderings of  high-quality synthetic color  images.  While
%% using synthetic images remains attractive for many problems, creating the virtual
%% scenes can also be expensive  and time consuming.  Generating animated realistic
%% hand grasps  of various objects,  as it would be  required to solve  the problem
%% considered in this  paper remains challenging. Being able to  use real sequences
%% for  training has  thus  also its  advantages. Moreover,  evaluation  has to  be
%% performed on real images.

\subsection{Joint Hand+Object Pose Estimation}

Early         approaches        for         joint        hand+object         pose
estimation~\cite{Oikonomidis2011full,Wang11d,Ballan2012motion}  typically relied
on  multi-view camera  setups, and  frame-by-frame tracking  methods, which  may
require      careful      initialization      and     drift      over      time.
\cite{panteleris20153d,Tzionas_2015_ICCV}  propose generative  methods to  track
finger   contact    points   for   in-hand   RGB-D    object   shape   scanning.
\cite{pham2015capturing,Pham_2015_CVPR} consider sensing from vision to estimate
contact forces during hand+object interactions  using a single RGB-D camera, and
then estimate the hand and the  object pose.  However, these methods are limited
to small occlusions.

\cite{kyriazis2013physically,Tzionas2016} propose to use a physics simulator and
a  3D renderer  for  frame-to-frame tracking  of hands  and  objects from  RGB-D.
\cite{Kyriazis14} uses  an ensemble  of Collaborative Trackers  for multi-object
and multiple  hand tracking from  RGB-D images.   The accuracy of  these methods
seems to  be qualitatively  high, but  as the ground  truth acquisition in
a real-world is known to be hard, they evaluate the proposed method on
synthetic datasets, or by measuring the  standard deviation of the difference in
hand/object poses during a grasping scenario.

\cite{Tsoli_2018_ECCV} considers the problem of  tracking a deformable object in
interaction with a hand, by optimizing  an energy function on the appearance and
the kinematics  of the hand,  together with hand+object  contact configurations.
However, it is  evaluated quantitatively only on synthetic  images, which points
to the difficulty  of evaluation on real data.  In  addition, they only consider
scenarios where the  hand is visible from  a top view, restricting  the range of
the hand poses and not allowing occlusions.

Very recently,  \cite{kokic2019learning} uses a  coarse hand pose  estimation to
retrieve the 3D pose and shape of hand-held objects. However, they only consider
a   specific  type   of   object  and   do  not   estimate   the  object   pose.
\cite{hasson2019learning}  presents a  model  with contact  loss that  considers
physically feasible hand+object interaction  to improve grasp quality.  However,
to  estimate 3D  hand pose,  they  predict PCA  components for  the pose,  which
results  in  lower   accuracy  compared  to  ours,   as  our  experiments
show. \cite{tekin2019h}  proposes a deep model  to jointly predict 3D  hand and
object poses from egocentric views, but the absence of physical constraints might
result in infeasible grasps.

\subsection{Hand+Object Datasets}

Several datasets for  hand+object interactions have already  been proposed. Many
works    provide    egocentric   RGB    or    RGB-D    sequences   for    action
recognition~\cite{bullock2015yale,cai2015scalable,fathi2011learning,bambach2015lending,rogez2015understanding}\comment{,luo2017scene,Tzionas_2015_ICCV}.
However,
they  focus  on   grasp  and  action  labels  and  do   not  provide  3D  poses.
\cite{choi2017robust,rogez20143d,mueller2017real,Tsoli_2018_ECCV}  
generate synthetic datasets with 3D hand pose  annotations, but fine interaction between a
hand and an object remains difficult to generate accurately.

\cite{Tzionas2016,Tzionas2014capturing}  captured sequences  in  the context  of
hand+hand  and   hand+object  interaction,   with  2D  hand   annotations  only.
\cite{Myanganbayar18} collected  a dataset of  real RGB images of  hands holding
objects.  They  also provide 2D joint  annotations of pairs of  non-occluded and
occluded hands,  by removing  the object  from the grasp  of the  subject, while
maintaining  their hand  in  the same  pose.   \cite{goudie20173d} proposes  two
datasets, a hand+object  segmentation dataset, and a  hand+object pose estimation
dataset. However,  for both  datasets, the  background pixels  have been  set to
zero, and the training  images only consist of a hand  interacting with a tennis
ball.   They provide  hand  pose  annotations and  object  positions, by  manually
labelling the joints and using a generative method to refine the joint positions.
\cite{JooSS18} generate a large scale dataset with full body pose and hand pose annotations in a multi-view setup. They use a generative approach to fit the body and hand models to 3D keypoints and point cloud. However, their dataset focuses on total body pose annotation and not hand+object interactions exclusively and does not provide object pose annotations.

\cite{RealtimeHO_ECCV2016} proposed  an RGB-D dataset  of a hand  manipulating a
cube, which  contains manual ground  truth for  both fingertip positions  and 3D
poses of the  cube.  \cite{pham2018hand} collected a dataset  where they measure
motion and force under different  object-grasp configurations using sensors, but
do   not   provide  3D   poses.    In   contrast   to  these   previous   works,
\cite{garcia2018first} provides a dataset of hand and object interactions with 3D annotations
for both hand joints and object pose.  They used a motion capture system made of
magnetic sensors  attached to  the user's hand and  to the  object in  order to
obtain hand 3D pose annotations in  RGB-D video sequences.  However, this changes
the appearance of the hand in the color images as the sensors and the tape attaching
them are visible.

Very recently,  \cite{hasson2019learning} introduced  ObMan, a large  dataset of
images of  hands grasping  objects. The images in ObMan dataset are synthetic  and the
grasps  are generated  using an  algorithm from  robotics.  Even  more recently,
\cite{zimmermann2019freihand} proposed a multi-view RGB dataset, FreiHAND, which
includes hand-object interactions.  However, the annotations are  limited to the
3D poses and shapes of  the hand.  Further, \cite{zimmermann2019freihand} uses a
human-in-the-loop method  to obtain annotations  from multiple RGB cameras  in a
green-screen  background environment.  Our method,  on  the other  hand is  fully
automatic, capable of working even on a single RGBD-camera setup and does not make
any assumption  on the background.  The objects in  our dataset are  also larger
than those  in FreiHAND, thus  resulting in a  more challenging scenario  as the
occlusions are larger. The annotation  accuracy of our method is comparable
to \cite{zimmermann2019freihand} as described in Section~\ref{sec:evalAnno}.

As illustrated  in Fig.~\ref{fig:one} and  Table~\ref{tab:hand_obj_dataset}, our
\datasetname  dataset is  the first  markerless dataset  providing both  3D hand
joints and 3D  object pose annotations for  real images, while the  hand and the
object are heavily occluded by each other.

%\begin{table}
%		\begin{center}
%			%\begin{tabular}{C{1.4cm}|C{1.3cm}|C{1.3cm}|C{1.3cm}|C{1.3cm}|C{1.5cm}|C{1.9cm}|C{2.3cm}|C{1.0cm}}
%			\resizebox{.47\textwidth}{!}{%
%				\begin{tabular}{lccccc} %{c|c|c|c|c|c|c|c|c}
%					\toprule
%					Dataset & \specialcell{No. of\\Frames} & \specialcell{3D Object\\Pose} & \specialcell{Marker-\\less} & \specialcell{Real\\Images}  & Labels \\
%					\midrule
%					PAN~\cite{JooSS18} & 675K  & - & + & + & automatic \\
%					GAN~\cite{Mueller2018} & 300K & - & + & - & synthetic\\
%					FPHA~\cite{garcia2018first} & 100K  & + & - & + & automatic\\
%					ObMan~\cite{hasson2019learning} & 150K  & + & + & -  & synthetic \\
%					Freihand~\cite{zimmermann2019freihand} & 37K  & - & + & + &  hybrid \\
%					\textbf{HO-3D (ours)} & 78K  & + & + & + &  automatic \\
%					\bottomrule
%			\end{tabular}}
%			\vspace{-2mm}
%			\caption{ Comparison of hand+object datasets.\shreyasrmk{add more cols}}
%			\vspace{-9mm}
%			\label{tab:hand_obj_dataset}
%		\end{center}
%\end{table}}

\begin{table}
  \begin{center}
    %\begin{tabular}{C{1.4cm}|C{1.3cm}|C{1.3cm}|C{1.3cm}|C{1.3cm}|C{1.5cm}|C{1.9cm}|C{2.3cm}|C{1.0cm}}
    \resizebox{.47\textwidth}{!}{%
      \begin{tabular}{lccccccc} %{c|c|c|c|c|c|c|c|c}
	\toprule
	Dataset & \specialcell{No. of\\Frames} & \specialcell{3D Object\\Pose} & \specialcell{Marker-\\less} & \specialcell{Real\\Images}  & Labels & \specialcell{No. of\\Objects} & \specialcell{No. of\\Subjects}\\
	\midrule
	PAN~\cite{JooSS18} & 675K  & - & + & + & automatic & - & 70\\
	GAN~\cite{Mueller2018} & 300K & - & + & - & synthetic & - & -\\
	FPHA~\cite{garcia2018first} & 100K  & + (23K frames) & - & + & automatic & 26 (4 models) & 6\\
	ObMan~\cite{hasson2019learning} & 150K  & + & + & -  & synthetic & 2.7K & 20\\
	Freihand~\cite{zimmermann2019freihand} & 37K  & - & + & + &  hybrid & 27 & 35\\
	\textbf{HO-3D (ours)} & 78K  & + & + & + &  automatic & 10 & 10\\
	\bottomrule
    \end{tabular}}
    \vspace{-2mm}
    \caption{ Comparison of hand+object datasets.}
    \vspace{-9mm}
    \label{tab:hand_obj_dataset}
  \end{center}
\end{table}

%------------------------------------------------------------------------

\section{3D Annotation Method}
\label{sec:method}

%% -*- mode: latex; mode: flyspell -*-

\newcommand{\calC}{\mathcal{C}}
\newcommand{\calD}{\mathcal{D}}
\newcommand{\calL}{\mathcal{L}}
\newcommand{\calM}{\mathcal{M}}
\newcommand{\calN}{\mathcal{N}}
\newcommand{\calP}{\mathcal{P}}
\newcommand{\calW}{\mathcal{W}}
\newcommand{\bj}{\mathbf{j}}
\newcommand{\bn}{\mathbf{n}}
\newcommand{\bv}{\mathbf{v}}
\newcommand{\bP}{\mathbf{P}}
\newcommand{\bS}{\mathbf{S}}
\newcommand{\bT}{\mathbf{T}}
\newcommand{\bbeta}{\boldsymbol{\beta}}
\newcommand{\IR}{\mathbb{R}}

\newcommand{\dpt}{\text{dpt}}
\newcommand{\depth}{\text{depth}}
\newcommand{\mask}{\text{mask}}
\newcommand{\joint}{\text{joint}}
\newcommand{\kps}{\text{j2D}}
\newcommand{\icp}{\text{3D}}
\newcommand{\phy}{\text{phy}}
\newcommand{\silh}{\text{silh}}
\newcommand{\tc}{\text{tc}}
\newcommand{\proj}{\text{proj}}

\newcommand{\hh}{\text{h}}
\newcommand{\hw}{\text{h}}
\newcommand{\ho}{\text{h:o}}
\newcommand{\oo}{\text{o}}

\newcommand{\De}{\text{D}}
\newcommand{\Img}{\text{I}}
\newcommand{\PCloud}{\text{P}}
\newcommand{\PCloudO}{\text{P}_\oo}
\newcommand{\PCloudH}{\text{P}_\hh}
\newcommand{\Ver}{\text{V}}
\newcommand{\pose}{\text{p}}
\newcommand{\tildePose}{\tilde{\text{p}}}
\newcommand{\hatPose}{\hat{\text{p}}}
\newcommand{\barPose}{\bar{\text{p}}}
\newcommand{\Ke}{\text{K}}
\newcommand{\Jo}{\text{J}}
\newcommand{\he}{\text{h}}

\newcommand{\ke}{\text{k}}
\newcommand{\di}{\text{d}}

\newcommand{\mvleft}{\!\!\!\!}

We    describe     below    our    method    for     annotating    a    sequence
$\mathcal{T}=\big\{\{(\Img^t_c,  \De^t_c)\}^{N_C}_{c=1}\big\}^{N_F}_{i=1}$   of
$N_C \times N_F$ of RGB-D frames, captured by  $N_C$ cameras.  The sequence  captures a
hand interacting  with an  object.  Each RGB-D  frame is made  of a  color image
$\Img^t_c$ and a depth map $\De^t_c$.

We  define the  3D hand  and object  poses in  Section~\ref{sec:poses}, and  our
general  cost  function  in  Section~\ref{sec:cost}.  We  initialize  the  poses
automatically and optimize the cost function  in multiple stages as described in
Sections~\ref{sec:multiCameraSetup} and \ref{sec:singleCameraSetup}.

\subsection{3D Hand and Object Poses}
\label{sec:poses}

We    aim   to    estimate    the   3D    poses    $\calP   =    \{(\pose_\hh^t,
\pose_\oo^t)\}^{N_F}_{t=1}$ for both  the hand and the object in  all the images
of the sequence.  We adopt the MANO hand model~\cite{romero2017embodied} and use
the objects from the YCB-Video dataset~\cite{posecnn2018} as their corresponding
3D models are available and of good quality. The MANO hand pose $\pose_\hh^t \in
\IR^{51}$ consists of 45 DoF~(3 DoF for each of the 15 finger joints) plus 6 DoF
for rotation  and translation of the  wrist joint.  The 15  joints together with
the wrist joint form  a kinematic tree with  the wrist joint node as  the first parent
node.  In  addition to  the pose  parameters $\pose_\hh^t$,  the hand  model has
shape parameters $\bbeta \in \IR^{10}$ that are  fixed for a given person and we
follow a  method similar to  \cite{Tan_2016_CVPR} to estimate  these parameters.
More  details  about  the  shape   parameter  estimation  are  provided  in  the
supplementary material. The object  pose $\pose_\oo^t \in \text{SE}(3)$ consists
of 6 DoF for global rotation and translation.

%% We call  the  hand  pose in  the  object  frame of reference as
%% \textit{grasp pose} in order to differentiate it from the hand pose in the world
%% coordinate system.

\subsection{Cost Function}
\label{sec:cost}

We  formulate   the  hand+object  pose   estimation  as  an   energy 
minimization problem:
\begin{equation}
\label{eq:global}
\hat{\calP} = \arg\min\limits_{\calP} \sum^{N_F}_{t=1} 
\big(E_\calD(\pose_\hh^t, \pose_\oo^t) + E_\mathcal{C}(\pose_\hh^t, \pose_\oo^t)\big)\> ,
\end{equation}
where $E_\mathcal{D}$ and  $E_\mathcal{C}$ represent the energy  from data terms
and constraints, respectively.  We define $E_\mathcal{D}$ as,
\begin{equation}
  \begin{array}{r}
    \mvleft\mvleft
    E_\calD(\pose_\hh^t, \pose_\oo^t) = \sum\limits^{N_C}_{c=1} \Big( \alpha E_\mask(\Img^t_c,\pose_\hh^t,\pose_\oo^t) \:+ \beta E_\dpt(\De^t_c, \pose_\hh^t,\pose_\oo^t)  + \\[0.5em]
   \mvleft \gamma E_\kps(\Img^t_c, \pose_\hh^t) \Big) \> + \delta E_\icp(\{\De^t_c\}_{c=1..N_C}, \pose_\hh^t,\pose_\oo^t) \> ,
\label{eq:dataTerm}
\end{array}
\end{equation}

  %% \begin{array}{l}
  %%   \mvleft\mvleft\mvleft \> \>
  %%   E_\calD(p_h^t, p_o^t) = \sum\limits^{N_C}_{c=1} \Big( \alpha E_\mask(I^t_c,p_h^t,p_o^t) \:+ \\[0.5em]
  %%     \> \> \> \> \> \> \> \> \> \> \> \> \> \> \> \> \> \>\beta E_\dpt(D^t_c, p_h^t,p_o^t)  +  \gamma E_\kps(I^t_c, p_h^t) \Big) \> +\\[0.5em]
  %%    \> \> \> \> \> \> \> \> \> \> \> \> \> \> \> \> \> \> \delta E_\icp(\{D^t_c\}_{c=1..N_C}, p_h^t,p_o^t) \> ,

%\begin{align}
%%\mvleft\mvleft
%E_\calD(p_h^t, p_o^t) =& \sum\limits^{N_C}_{c=1} \Big( \alpha E_\mask(I^t_c,p_h^t,p_o^t) \:+ \\ \nonumber
%\mvleft &\beta E_\dpt(D^t_c, p_h^t,p_o^t)  +  \gamma E_\kps(I^t_c, p_h^t) \Big) \> +\\ \nonumber
%\mvleft &\delta E_\icp(\{D^t_c\}_{c=1..N_C}, p_h^t,p_o^t) \> ,
%\label{eq:dataTerm}
%\end{align}
%
\noindent where $E_\mask(\cdot)$ is a  silhouette discrepancy term, $E_\dpt(\cdot)$ a
depth  residual  term,  $E_\kps(\cdot)$  a 2D  error  in  hand  joint
locations,  and $E_\icp(\cdot)$  a  3D  distance term.  This  last  term is  not
absolutely necessary,  however, we  observed  that it  significantly  speeds up  convergence.
$\alpha$, $\beta$,  $\gamma$, $\delta$ are weights.

The constraints energy $E_\mathcal{C}$ is defined as,
\begin{align}
E_\mathcal{C}(\pose_\hh^t, \pose_\oo^t)  = \; & \epsilon E_\joint(\pose_\hh^t) \; + \zeta E_\phy(\pose_\hh^t,\pose_\oo^t) \; + \nonumber\\
& \eta E_\tc(\pose_\hh^t,\pose_\oo^t,\pose_\hh^{t-1},\pose_\oo^{t-1},\pose_\hh^{t-2},\pose_\oo^{t-2}) \> ,
\end{align}
where $E_\joint(\cdot)$  denotes a prior on  the hand pose to  prevent unnatural
poses, $E_\phy(\cdot)$ is a physical plausibility term ensuring the hand and the
object  do not  interpenetrate,  and $E_\tc(\cdot)$  is  a temporal  consistency
term. The  terms are weighted by  parameters $\epsilon$, $\zeta$ and  $\eta$.

We detail  each of the terms  in $E_\mathcal{D}$ and $E_\mathcal{C}$  below. For
simplicity, we  omit the  frame index  $t$ from our  above notation  except when
necessary.

\vspace{-0.3cm}
\paragraph{Silhouette discrepancy term $E_\mask$. }
The $E_\mask(\cdot)$ term compares the silhouettes  of the hand and the object models
rendered  with the  current estimated  poses and  their segmentation  masks.  We
obtain a segmentation $S_c(\Img)$ of the hand and the object in the color image $\Img$ of camera $c$ using
DeepLabv3~\cite{deeplabv2017}  trained   on  images  created   by  synthetically
over-laying  and under-laying  images of  hands on  YCB objects.  More
  details about this step are given in the supplementary material. The hand and
object  models  are  rendered  on   the  camera  plane  using  a  differentiable
renderer~\cite{Henderson2019},  which  enables   computing  the  derivatives  of
$E_\mask$ with  respect to the pose  parameters. The silhouette of  the hand and
object rendered on  camera $c$ is denoted by $\text{RS}_c(\pose_\hh,\pose_\oo)$  and the silhouette
discrepancy is defined as,
\begin{align}
E_\mask(\Img_c, \pose_\hh, \pose_\oo) = \lVert \text{RS}_c(\pose_\hh, \pose_\oo) - S(\Img_c) \rVert^2 \> .
\end{align}

\paragraph{Depth residual term $E_\dpt$. }
The depth residual term is similar to the segmentation discrepancy term:
\begin{equation}
E_\dpt(\De_c, \pose_\hh, \pose_\oo) = \text{Tukey}(\lVert \text{RD}_c(\pose_\hh, \pose_\oo) - \De_c \rVert) \> ,
\end{equation}
where $\text{RD}_c(\pose_\hh, \pose_\oo)$ is the depth rendering of the hand and
the object under their current estimated poses $\pose_\hh$ and $\pose_\oo$.  The
Tukey function is a robust estimator that  is similar to the $\ell_2$ loss close
to  0, and  constant after  a threshold.   It is  useful to  be robust  to small
deviations in the scale  and shape of the hand and object  models and also noise
in the  captured depth maps.   $E_\dpt$ is differentiable as we  employ a
differentiable renderer~\cite{Henderson2019} for rendering the depth maps.

\vspace{-0.3cm}
\paragraph{2D Joint error term $E_\kps$. }

%% Using our initial dataset  of 15,000 frames, we trained a CNN  to predict the 2D
%% locations of the 21 hand joints to further bolster our optimization strategy for
%% the subsequent sequences.  The predicted 2D joint locations in each camera image
%% is  denoted by  $K_c$.  \shreyas{

The 2D joint error term is defined as,
\begin{equation}
E_\kps(\Img_c, \pose_\hh) = \sum_{i=1}^{21} \he[i] \Big\lVert
\proj_c(\Jo_{\pose_\hh}[i]) - \Ke_c[i] \Big\rVert^2 \> ,
\label{eq:j2dErr}
\end{equation}
where $\Jo_{\pose_\hh}[i]$ denotes the $i^\text{th}$ 3D hand joint location under pose $\pose_\hh$, the $\proj_c(\cdot)$ operator projects it onto camera $c$, $\Ke_c[i]$ is its predicted 2D location, and $\he[i]$ its confidence.  The 21 hand joints in $E_\kps(\cdot)$ consist of 15 finger joints, 5 finger tips, and the wrist joint.

In practice,  we take  the $\Ke_c[i]$ as  the locations of  the maximum  values of
heatmaps, and  the $\he[i]$  as the  maximum values  themselves. To  predict these
heatmaps, we trained a CNN based  on the architecture of \cite{Wei16}.  Training
data  come   from  an  initial   dataset~\cite{ho3d_old}  we  created   using  a
semi-automatic method. This  dataset is made of 15,000 frames  from 15 sequences
in a single camera setup. We manually initialized the grasp pose and object pose
for the first  frame of each sequence.  The manipulators  were asked to maintain
their grasp poses as rigid as possible to make the registration easier.  We then
ran the optimization  stages for the single camera case  described in Section~\ref{sec:singleCameraSetup}.  After
optimization, we  augmented the  resulting dataset by  scaling and  rotating the
images,     and      adding     images     from     the      Panoptic     Studio
dataset~\cite{xiang2019monocular}, which contain 3D annotations for hands.

\vspace{-0.3cm}
\paragraph{3D error term $E_\icp$. }
This term  is not  absolutely necessary as  the depth  information from  all the  cameras is
already exploited by $E_\dpt$, however it accelerates the convergence by guiding
the optimization towards the  minimum even from far away.  We  build a point cloud
$\PCloud$ by  merging the depth maps  from the RGB-D cameras  after transforming
them  to   a  common  reference  frame.    More  details  on  the   point  cloud
reconstruction can be found in the supplementary material.
  
We segment  $\PCloud$ into  an object  point cloud $\PCloudO$  and a  hand point
cloud $\PCloudH$ using  the segmentation mask $S_c(\Img)$ in  each camera image.
At  each iteration  of the  optimization, for  each point  $\PCloudO[j]$ of  the
object point cloud, we look for the closest vertex $\Ver_\oo[j^*]$ on the object
mesh, and for each point $\PCloudH[k]$ of  the hand point cloud, we look for the
closest vertex  $\Ver_\hh[k^*]$ on  the hand mesh.   $E_\icp(\PCloud, \pose_\hh,
\pose_\oo)$ is then defined as,
\begin{align}
\sum_j &\big\lVert \PCloudO[j] - \Ver_\oo[j^*]\big\rVert^2 + \sum_k \big\lVert \PCloudH[k] - \Ver_\hh[k^*]\big\rVert^2 \> .
\end{align}

%% \begin{align}
%% E_\icp(P, \pose_\hh, \pose_\oo) = \sum_i &\big\lVert \pose_\oo[i] - V_o[i^*]\big\rVert^2 + \nonumber \\
%% &\sum_j \big\lVert \pose_\hh[j] - V_h[j^*]\big\rVert^2 \> .
%% \end{align}

%% The ICP  and silhouette discrepancy  terms allow  coarse alignment of  the model
%% with the  observed RGB-D data.  In scenarios where  the point cloud  is obtained
%% from a  single camera, only the  point cloud corresponding to  the camera facing
%% side of the  object and hand is  visible. ICP alignment in these  cases leads to
%% inaccurate poses  as shown in Fig.~\ref{fig:depth_in_opti}.

%% segmentation mask $S_c$ in each camera image.   Thus, $P = (\pose_\oo, \pose_\hh)$.  In each
%% iteration of  the optimization for  each point in the  point cloud, we  find the
%% closest mesh vertex in the corresponding model as
%% %
%% \begin{align}
%% i^* = \argmin_k \big\lVert \pose_\oo[i] - V_o[k] \big\rVert^2 \nonumber \\ j^* =
%% \argmin_k \big\Vert \pose_\hh[i] - V_h[k] \big\Vert^2 \> ,
%% \end{align}
%% %
%% where $V_o[i]$ and $V_h[i]$ denote the $i^{th}$  vertex in the set of object and
%% hand vertices, respectively. $E_\icp$ is then defined as
%% %
%% \begin{align}
%% E_\icp(P, \pose_\hh, \pose_\oo) = \sum_{i,j} &\big\lVert \pose_\oo[i] - V_o[i^*]\big\rVert^2 + \nonumber \\
%% &\big\lVert \pose_\hh[j] - V_h[j^*]\big\rVert^2 \> .
%% \end{align}

%% This term  helps  in aligning  the  models  of hand  and  object with  their
%% respective  point  cloud.  

\vspace{-0.5cm}
\paragraph{Joint angle constraint $E_\joint$. }
This  term imposes  restrictions on  the 15  joints of  the hand  to ensure  the
resulting  pose is  natural.   The  three-dimensional  rotation  of  a joint  is
parameterized using the axis-angle representation in MANO model, resulting in 45
joint angle  parameters.  A common  solution when using MANO model is to optimize in
the PCA space of 3D joint angles with an $\ell_2$ regularizer~\cite{zimmermann2019freihand, boukhayma20193d} for pose coefficients. However, we  observed  in practice  that
optimizing Eq.~\ref{eq:global} in PCA space had less expressibility: some of the complex grasp poses in our dataset could not be accurately expressed in the PCA space, which was constructed with relatively simpler grasp poses and free-hand gestures. %when  manipulating
%  objects,  the hand  joint  limits  tend to  be  larger  compared to  free-hand
%  gestures, which were used to construct the PCA space. 
  Instead, we optimize on joint angles directly and derive our
  own limits  for each  of the 45  joint parameters~(please  refer to supplementary
  material for these limits).  As in \cite{Zhou16c}, the joint angle constraint
term $E_\joint(\pose_\hh^t)$ is given by,
\begin{equation}
\sum_{i=1}^{45} \max(\underline{a_i} - a[i], 0) + \max(a[i] -
\overline{a_i}, 0) \> ,
\label{eq:ejoint}
\end{equation}
where $a[i]$  denotes the $i^{th}$ joint  angle parameter for pose  $\pose_\hh$, and
$\underline{a_i}$ and $\overline{a_i}$ correspond to its lower and upper limits.

\begin{figure*}[ht]
  \begin{center}
    \includegraphics[width=0.85\linewidth, trim=.2cm 3.5cm .0cm 2cm,clip]{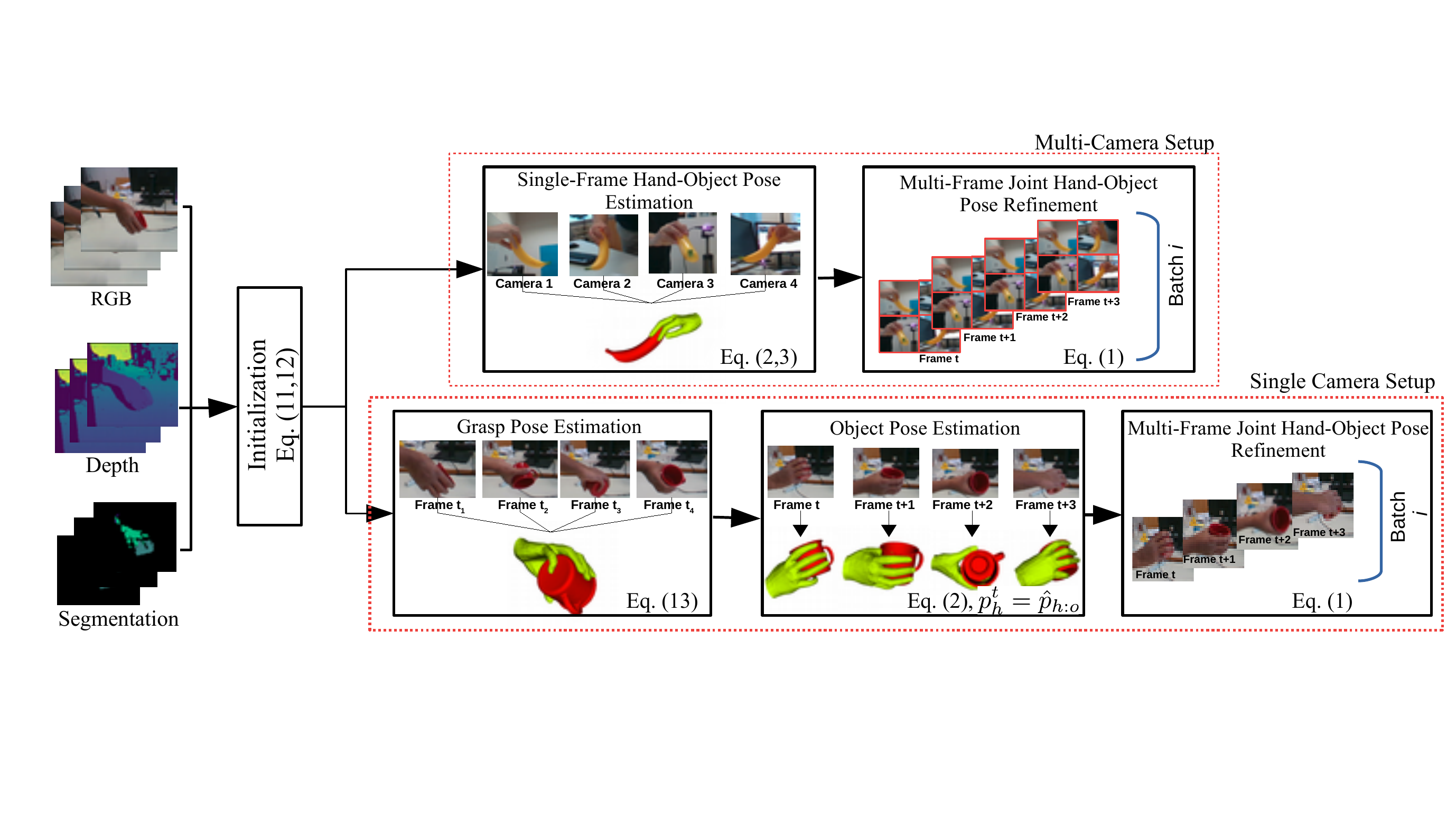}
  \end{center}
\vspace{-4mm}
\caption{The different stages of the multi-camera and single camera setups. See Section \ref{sec:fullOpt} for more details.}
\label{fig:methodOverview}
\vspace{-10px}
\end{figure*}	

%% \caption{The different stages of the multi-camera and single camera hand-object pose
%%   annotation methods. Frames $t_1,..,t_4$ are discontinuous frames in the sequence. See Section \ref{sec:fullOpt} for more details.}

\vspace{-0.3cm}
\paragraph{Physical plausibility term $E_\phy$. }
During optimization, the  hand model might penetrate the object  model, which is
physically not possible.  To avoid this, we add a repulsion term that pushes the
object and  the hand  apart if  they interpenetrate each  other.  For  each hand
vertex $V_\hh[m]$, the amount of penetration $\Gamma[m]$ is taken as, 
\begin{equation}
\Gamma[m] = \max(-\bn_\oo\big(\Ver_\oo[m^*]\big)^T\big(\Ver_\hh[m] -\Ver_\oo[m^*]\big), 0) \> , 
\end{equation}
where  $\Ver_\oo[m^*]$  is   the  vertex  on  object  closest   to  hand  vertex
$\Ver_\hh[m]$, and the $\bn_\oo(\cdot)$ operator  provides the normal vector for
a vertex.   In words, the amount  of penetration is estimated  by projecting the
vector  joining a  hand vertex  and its  nearest object  vertex onto  the normal
vector at  the object vertex location.   The physical plausibility term  is then
defined as,
\begin{equation}
E_\phy(\pose^t_\hh,\pose^t_\oo) = \sum_m \exp{\big(w~\Gamma[m]\big)} \> .
\end{equation}
We use  $w = 5$ in  practice, and only a subsampled  set of vertices of  the hand to
compute $E_\phy$ efficiently.

\paragraph{Temporal consistency term $E_\tc$.}

The previous  terms are all applied  to each frame independently.   The temporal
consistency term $E_\tc$  allows us to constrain together the  poses for all the
frames. We apply a 0-th and 1-st order  motion model on both the hand and object
poses:
\begin{equation}
\begin{array}{l}
E_\tc(\pose_\hh^t,\pose_\oo^t,\pose_\hh^{t-1},\pose_\oo^{t-1},\pose_\hh^{t-2},\pose_\oo^{t-2}) = \\[0.5em] \nonumber 
\lVert \Delta_h^t \rVert^2 + \lVert \Delta_o^t \rVert^2 + 
\lVert \Delta_h^t - \Delta_h^{t-1} \rVert^2 + \lVert \Delta_o^{t} -
\Delta_o^{t-1} \rVert^2 \> ,
\end{array}
\end{equation}
where $\Delta_h^t = \pose_\hh^t - \pose_\hh^{t-1}$ and $\Delta_o^t = \pose_\oo^t
- \pose_\oo^{t-1}$.  Since we  optimize a sum of these terms  over the sequence,
this effectively constrains all the poses together.

%% \begin{align}
%% E_\tc(\pose_\hh^t,\pose_\oo^t,\pose_\hh^{t-1},\pose_\oo^{t-1},\pose_\hh^{t-2},\pose_\oo^{t-2}) 
%% = \lVert \Delta_h^t \rVert^2 + \lVert \Delta_o^t \rVert^2 + \\ \nonumber 
%% \lVert \Delta_h^t - \Delta_h^{t-1} \rVert^2 + \lVert \Delta_o^{t} -
%% \Delta_o^{t-1} \rVert^2 \> ,
%% \end{align}

%% \begin{align}
%% E_\tc(\big\{\pose_\hh^{i+k},\pose_\oo^{i+k}\big\}_{k=\text{-}2}^0) 
%% = &\lVert \Delta_h^t \rVert^2 + \lVert \Delta_o^t \rVert^2 + \\ \nonumber 
%% &\lVert \Delta_h^{i} - \Delta_h^{i-1} \rVert^2 + \\ \nonumber 
%% &\lVert \Delta_o^{i} - \Delta_o^{i-1} \rVert^2,
%% \end{align}

%------------------------------------------------------------------------

\section{Optimization}
\label{sec:fullOpt}

Optimizing  Eq.~\eqref{eq:global} is  a  challenging  task, as  it  is a  highly
non-convex problem with  many parameters to estimate.  We  therefore perform the
optimization  in  multiple  stages as  shown  in  Fig.~\ref{fig:methodOverview}.
These stages are different for multi-camera  and single camera scenarios, and we
detail them below.

%% \subsection{2D Joint Detection}
%% \label{sec:autoInit}

%% To  perform  the initialization  automatically,  we  learned  to detect  the  2D
%% projections $\Ke_c[i]$ and their confidences $\he[i]$ of the hand joints required to
%% compute     $E_\kps$    of     Eq.~\eqref{eq:j2dErr}     using    an     initial
%% dataset~\cite{ho3d_old} we  created using a semi-automatic  method. This dataset
%% is made  of 15,000 frames  from 15  sequences in a  single camera setup,  and we
%% manually initialized the grasp pose and object  pose for the first frame of each
%% sequence.  The manipulators were asked to maintain their grasp poses as rigid as
%% possible to make  the registration easier.  We then ran  the optimization stages
%% for the  single camera case  described below.  After optimization,  we augmented
%% the resulting dataset by scaling and rotating the images, and adding images from
%% the   Panoptic  Studio   dataset~\cite{xiang2019monocular},  which   contain  3D
%% annotations  for hands.   We then  trained a  CNN based  on the  architecture of
%% \cite{Wei16} to predict the 2D projections  of the 21 hand
%% joints $\Ke$ in the image, in the  form of heatmaps trained with the $\ell_2$ loss
%% function.  Given a  new sequence, we use  the 2D joint predictions  made by this
%% CNN to initialize the hand poses as explained below.

\subsection{Multi-Camera Setup}
\label{sec:multiCameraSetup}

\paragraph{Initialization.}
In the multi-camera setup, we obtain a first estimate $\tildePose_\hw^0$ for the
hand pose in the first frame $(t=0)$ as,
\begin{equation}
\label{eq:multiViewInit}
\tildePose_\hw^0   =  \arg\min\limits_{\pose_\hw}   \sum_{c=1}^{N_C}
E_\kps(\Img_c^0, \pose_\hw) + \nu E_\joint(\pose_\hw) \> .
\end{equation}
We use the Dogleg  optimizer~\cite{ConnGoulToin00} to perform this optimization.
A  first estimate  $\tildePose_\oo^0$  for  the object  pose  in  this frame  is
obtained using  \cite{Rad17} trained by  synthetically over-laying hands  on YCB
objects as explained in Section~\ref{sec:cost}.

\paragraph{Single-frame joint pose optimization.}
We then obtain  estimates $\tildePose_\hw^t$ and $\tildePose_\oo^t$  for all the
other  frames ($t = 1..N_F$)  by   tracking.   We   minimize  $\big(
E_\calD(\pose_\hw^t,  \pose_\oo^t)  + E_\calC(\pose_\hw^t,  \pose_\oo^t)  \big)$
w.r.t.   $\pose_\hw^t$  and   $\pose_\oo^t$,  using  $\tildePose_\hw^{t-1}$  and
$\tildePose_\oo^{t-1}$ for initialization.  

%% \shreyas{Using the  initialization for the  first frame, the hand  and object
%% poses for all  the frames in the sequence are  estimated using tracking}.  We
%% denote these new estimates  $\hatPose_\hw^t$ and $\hatPose_\oo^t$.  The poses
%% for the first  frame ($t=0$) are taken to be  equal to $\tildePose_\hw^0$ and
%% $\tildePose_\oo^0$.     The    pose     parameters    $\hatPose_\hw^t$    and
%% $\hatPose_\oo^t$ for the other frames are

\paragraph{Multi-frame joint pose optimization.}
We finally  perform a  full optimization of  Eq.~\eqref{eq:global} w.r.t. $\pose_\hh^t$ and 
$\pose_\oo^t$ for $t=0..N_F$ over  all the
frames simultaneously using  estimates $\tildePose_\hw^t$ and $\tildePose_\oo^t$
for    initialization.     Due    to    memory    constraints,    we    optimize
Eq.~\eqref{eq:global}  in  batches instead  of  considering  all the  frames  in
sequence.   We use  a  batch size  of 20  frames  with $\alpha=20$,  $\beta=20$,
$\gamma=5\times10^{-5}$,    $\delta=50$,    $\epsilon=100$,   $\zeta=50$,    and
$\eta=100$,  and  the  Adam  optimizer  with  learning  rate  of  0.01  for  100
iterations.

\subsection{Single-Camera Setup}
\label{sec:singleCameraSetup}

\paragraph{Initialization.}
In the single camera  setup, as we assume that the  grasp pose varies marginally
across the sequence,  we initially make the assumption that  it remains constant
throughout the sequence.  In order to account for the minor changes which occur
in practice, we relax this assumption  in the latter stages of the optimization.
We thus obtain initial estimates $\tildePose_\hw^t$ for the hand poses as,
\begin{align}
\label{eq:singleViewPoseEst}
\{ \tildePose_\hw^t\}_t = \arg\min_{\{\pose_\hw^t\}_t} \sum_t E_\kps(\Img^t, \pose_\hw^t) + \nu E_\joint(\pose_\hw^t) \> ,
\end{align}
where the  joint angle parameters  are constrained to be  the same over  all the
frames, and only the rotation and translation parameters for the wrist joint can
be different.   In practice,  we perform  this optimization  only over  a random
subset $\Omega$ of the frames to save time.  We set $\nu = 50$, size of $\Omega$
to  20  and use  the  Dogleg  optimizer~\cite{ConnGoulToin00}.  First  estimates
$\tildePose_\oo^t$ for  the object  poses in  $\Omega$ are  obtained as  for the
multi-camera setup.

From the $\tildePose_\hw^t$ and the $\tildePose_\oo^t$, we can compute the grasp
pose in the  \emph{object coordinate system} $\tildePose_\ho^t$,  which is assumed
to be constant at this stage. The initial estimate of the constant grasp pose $\tildePose_\ho$  is taken as the average of $\{\tildePose_\ho^t\}_{t\in\Omega}$.% Since the object pose estimates $\tildePose_\oo^t$
%are  not  perfect,  we  take  $\tildePose_\ho$ as  the  average  motion  between
%$\tildePose_\hw^t$ and  $\tildePose_\oo^t$.

%% \shreyas{In the  rest of this  section, we denote  the grasp pose  i.e., hand
%% pose in the object  frame of reference as $\pose_\ho^t$ and  the hand pose in
%% the world  coordinate system as  $\pose_\hw^t$.  The poses  $\pose_\ho^t$ and
%% $\pose_\hw^t$ are connected by $\pose_\oo^t$}.  For initialization,

%% The  final  grasp and  object  poses  are estimated  in  3  stages as  shown  in
%% Fig.~\ref{fig:methodOverview}.  We  first estimate the grasp  pose $\pose_\ho$
%% (assuming same across all frames) from multiple frames in the first stage, which
%% is then used to  estimate object pose in each frame in the  second stage. In the
%% final stage, we employ temporal constraints and jointly optimize hand and object
%% poses over multiple frames and allow  for variation in grasp pose across frames.
%% More details about these stages is provided below.

\vspace{-0.5px}
\paragraph{Grasp pose estimation.}
We obtain  a better estimate of  the grasp pose $\hatPose_\ho$  and object poses
$\hatPose_\oo^t$ under fixed grasp pose assumption as,
%
%\begin{align}
%\hatPose_\ho, \{\hatPose_\oo^t\}_{t \in \Omega} =& \arg\min_{\pose_\ho,\{\pose_\oo^t\}_{t \in \Omega}} \sum_{t \in \Omega} E_\calD(f_{\pose_\oo^t}(\pose_\ho), \pose_\oo^t) + \\ \nonumber
%&\zeta E_\phy(f_{\pose_\oo^t}(\pose_\ho), \pose_\oo^t) + \epsilon E_\joint(\pose_\ho) \> ,
%\end{align} 
\begin{equation}
\begin{array}{r}
\hatPose_\ho, \{\hatPose_\oo^t\}_{t \in \Omega} = \arg\min_{\pose_\ho,\{\pose_\oo^t\}_{t \in \Omega}} \sum_{t \in \Omega} E_\calD(f_{\pose_\oo^t}(\pose_\ho), \pose_\oo^t) + \\[0.5em]
\zeta E_\phy(f_{\pose_\oo^t}(\pose_\ho), \pose_\oo^t) + \epsilon E_\joint(\pose_\ho), \\ [0.5em]
\end{array}
\end{equation}
wrt $\pose_\ho$ and $\pose_\oo^t$ over the frames  in $\Omega$, using $\tildePose_\ho$  and $\tildePose_\oo^t$
for  initialization. $f_{\pose_\oo^t}(\cdot)$ converts the grasp pose to the hand pose in the world coordinate system given object pose $\pose_\oo^t$.  This  optimization  accounts for  the  mutual  occlusions between hand and  object.
\vspace{-0.4cm}
\paragraph{Object pose estimation.}
Having a good estimate $\hatPose_\ho$ for  the grasp pose, we obtain object pose
estimates    $\hatPose_\oo^t$    for    all    the    frames    by    minimizing
$E_\calD\big(f_{\pose_\oo^t}(\hatPose_\ho),\pose_\oo^t\big)$ wrt $\pose_\oo^t$   over  each   frame  independently. We  use
$\hatPose_\oo^{t-1}$  to  initialize  the  optimization  over  $\pose_\oo^t$ at frame $t$,
except for $\hatPose_\oo^0$ where $\tildePose_\oo^0$  is used.  Note that the hand
pose is not optimized in this stage.

%% \shreyas{Having estimated the approximate hand grasp pose for the sequence, the object pose in each frame is estimated by tracking.
%% More specifically, for the first frame, $t=0$ the object pose is initialized with $\tildePose_\oo^0$ and for all subsequent frames $\pose^{t}_\oo$ is initialized with $\hatPose_\oo^{t-1}$.  The object pose $\hatPose_\oo^{t}$ at each frame is obtained by minimizing the cost function $E_\mathcal{D}(\hatPose_\ho,\pose_\oo^t)$. Note that the grasp pose is not optimized in this stage but the hand mesh in the current grasp pose $\hatPose_\ho$ is used in all the cost terms of $E_\mathcal{D}(\cdot)$.}
\vspace{-0.3cm}
\paragraph{Multi-frame joint hand+object pose refinement.}
In this  final stage, we  allow variations in the  grasp pose across  frames and
introduce temporal constraints.  We thus optimize Eq.~\eqref{eq:global} w.r.t. $\{(\pose_\hh^t,
\pose_\oo^t)\}^{N_F}_{t=1}$ over  all the frames  simultaneously, using  pose parameters
  $\hatPose_\oo^t$, $\hatPose_\hh^t=f_{\hatPose_\oo^t}(\hatPose_\ho)$ estimated  in  the  previous   stages  as initialization for $\pose_\oo^t$ and $\pose_\hh^t$.

%------------------------------------------------------------------------

\section{Monocular RGB based 3D Hand Pose}% \\ Estimation Method}
\label{sec:handposeSingleFrameSec}

%% -*- mode: latex; mode: flyspell -*-

For establishing a  baseline on our proposed dataset for  single RGB image based
hand pose  prediction, we use a  CNN architecture based on  a Convolutional Pose
Machine~(CPM)~\cite{Wei16}   to   predict   the    2D   hand   joint   locations
$\{{\ke_i}\}_{i=1..21}$.  In addition,  we also predict the root-relative hand  joint directions  $\{\di_{i}\}_{i=1..20}$, by adding  an additional
stage at the  end of the CPM  and replacing the last layer  with a fully connected
layer.  More details on the architecture are provided in the supplementary material.
The 3D  joint locations and  shape parameters of the  hand are then  obtained by
fitting a MANO  model to these predictions.  The loss  function for this fitting
procedure is:
\begin{align}
\label{eq:manoFitEst}
\sum_{i=1}^{21} \| \hat{\ke}_i - \ke_i\|^2
 + \rho \sum_{i=1}^{20} \big(1 - \hat{\di}_i \cdot \di_i\big)   +
\sigma E_\joint(\pose_\hh) + \tau \| \bbeta \|^2 \> ,
\end{align}
where  $\hat{\di}_i  =  \frac{\pose_\hh[i+1]-  \pose_\hh[1]}{\|  \pose_\hh[i+1]  -
  \pose_\hh[1]\|}$,   $\hat{k}_i    =   \proj\big(\Jo_{\pose_\hh}[i]\big)$   and
$E_\joint$ is  defined in  Eq.~\eqref{eq:ejoint}. We use  $\rho=10$, $\sigma=5$,
and $\tau=1$.

%% The weights  $\rho$, $\sigma$,  and $\tau$  are set to  10, 5,  and 1,
%% respectively.

%------------------------------------------------------------------------

%\section{Exploring the \datasetname Dataset}
%
%\input{dataset.tex}

%------------------------------------------------------------------------

\section{Benchmarking \datasetname}

%% -*- mode: latex; mode: flyspell -*-

\begin{table*}
  \begin{center}
    \resizebox{\textwidth}{!}{%
      \begin{tabular}{ccccccccc}%{|c|c|c|c|c|c|c|c|c|}
	  \hline
    	\multirowcell{2}{\textbf{Terms}} & \multirowcell{2}{Initialization} & \multicolumn{6}{c}{Single-frame Optimization} & \multirowcell{2}{Multi-frame\\Opt. (Eq. 1)} \\
	  \Xcline{3-8}{0.05em}
        %			qq & \multicolumn{7}{c}{Multi-camera Single-frame Optimization} & Opt
        %			\midrule
	&  & $E_\silh$ & $E_\dpt$ & $E_\silh+E_\dpt$  & $E_\silh+E_\dpt+E_\icp$ & $E_\silh+E_\dpt+E_\icp+E_\phy$ & $E_\silh+E_\dpt+E_\icp+E_\phy+E_\tc$ & \\
	\hline
	\textbf{mean (std)} & 4.20 ($\pm$3.32) & 1.17 ($\pm$1.12) & 2.22 ($\pm$1.22) & 1.04 ($\pm$0.43) & 0.98 ($\pm$0.40) & 0.99 ($\pm$0.40) & 0.92 ($\pm$0.34) & 0.77 ($\pm$0.29) \\
	\hline
    \end{tabular}}
	\vspace{-2mm}
    \caption{Evaluation of  the accuracy for  the multi-camera setup.   We report
      the average hand-joint errors (in cm) for different combinations of the terms
      in  Eq.~\eqref{eq:global}.  The  final error  is  comparable to  the
      recent FreiHAND dataset~\cite{zimmermann2019freihand}.}
    \vspace{-7mm}
    \label{tab:table2}
  \end{center}
\end{table*}

\begin{table}
  \begin{center}
    \resizebox{0.47\textwidth}{!}{%
      \begin{tabular}{lcccc}
        \toprule
        Stages & Init. & Grasp Pose Est. & Object Pose Est.  & Refinement\\
        \midrule
        Hand & 5.40 & 3.60 & 0.91 & 0.77 \\
        Object & 4.02 & 4.02 & 0.52 & 0.45 \\
        \bottomrule
    \end{tabular}}
    \vspace{-2mm}
    \caption{Evaluation  of  the  accuracy  for  the  single-camera  setup.  The
      accuracy  (average  mesh  error  in  cm) is  measured  at  each  stage  of
      optimization  by   comparing  with   the  annotations   from  multi-camera
      setup. The results  show that the annotation quality of  our single camera
      method is similar to that of the multi-camera setup.}
    \vspace{-6mm}
    \label{tab:table5}
  \end{center}
\end{table}

In this  section, we evaluate  both our annotation  method and our  baseline for
hand  pose prediction  from  a  single color  image  in hand+object  interaction
scenarios.  We used  our 3D pose annotation method to  annotate 68 sequences, totalling 77,558 frames of  10 different users manipulating one among
10 different objects  from the YCB dataset.  The image  sizes are $640\times480$
pixels for both the color and depth  cameras, and we used 5 synchronized cameras
in our multi-camera setup. The cameras were synchronized with an
accuracy of 5ms.

\subsection{Evaluation of the Annotation Method}
\label{sec:evalAnno}

For validating the accuracy of our  annotation method, we manually annotated the
3D locations  of the  3D joints in  randomly selected frames  of a  sequence, by
relying on  the consolidated point cloud  from the 5 cameras.   We then compared
these locations to the ones predicted with our method (explained in Section~\ref{sec:multiCameraSetup}) using the multi-camera setup.

As shown in the last column of Table~\ref{tab:table2}, our method achieves an
average joint error accuracy of lower than 8mm on average, with an Area Under the Curve
metric~(AUC) of 0.79. This metric is comparable with the results reported
for the recent FreiHAND dataset~\cite{zimmermann2019freihand}~(AUC=0.791). Note that the
occlusions in our dataset are higher due to larger objects and that we do not use green screens.

To analyze the influence of the different terms in Eq.~\eqref{eq:global}, we run
the optimization  of Eq.~\eqref{eq:global}  by enabling only  a subset  of these
terms, and  report the results  in Table~\ref{tab:table2}.  While  $E_\silh$ and
$E_\dpt$ terms alone  cannot provide good pose estimates,  together they provide
better estimates  as it leads  to a loss function  with less local  minima.  The
$E_\icp$  term provides  a  minor improvement  in estimates  but  speeds up  the
convergence.  Though  the physical plausibility  term $E_\phy$ does not  help in
improving the pose  estimates, it results in more natural  grasps.  The last two
columns  show the  effect of  multi-frame based  joint optimization  compared to
single-frame  based  optimization  when  all   the  terms  are  considered.  The
multi-frame  multi-camera based  optimization over  all the  terms improves  the
accuracy by about 15\%.

The  accuracy of  the single  camera based  annotation method  is calculated  by
considering the annotations from the multi-camera method as ground truth for a given
sequence. More specifically, for a sequence of
1000  frames, we compute the average difference   between  hand+object  mesh   vertices  obtained  from  single  and
multi-camera setups. Further, we calculate the  accuracy after each stage of the
single-camera  setup.  The  results  are given  in  Table~\ref{tab:table5}.  The
estimated poses  with these two  methods are consistent  with each other  with an
average mesh error  of 0.77cm and 0.45cm for hand  and object, respectively. The
final refinement stage yields a 15\% improvement in accuracy.

\begin{table}[]
  \begin{center}
    \resizebox{0.47\textwidth}{!}{%
      \begin{tabular}{lcccc}
	\toprule
	Method & \specialcell{Mesh Error$\downarrow$} & F@5mm$\uparrow$ & F@15mm$\uparrow$  & \specialcell{Joint Error$\downarrow$}\\
	\midrule
	Joints2D & 1.14 & 0.49 & 0.93 & 3.14 \\
	Joints2D + Dir. Vec. & \textbf{1.06} & \textbf{0.51} & \textbf{0.94} & \textbf{3.04} \\
	%		Hasson \textit{et al.}\cite{hasson2019learning} & 1.28 & 0.43 & 0.90 & 7.75 \\
	\cite{hasson2019learning}  & 1.10 & 0.46 & 0.93 & 3.18 \\
	\bottomrule
    \end{tabular}}
    \vspace{-2mm}
    \caption{Evaluation  of  different  methods   for  single  frame  hand  pose
      prediction. The Mesh Error (in cm) and \textit{F}-score are obtained after
      aligning the  predicted meshes  with ground truth  meshes. The  Mean joint
      error (in  cm) is obtained after  aligning the position of  the root joint
      and overall scale with the ground  truth. Hand pose prediction using joint
      direction  predictions along  with  2D joint  predictions provides  better
      accuracy    than   directly    predicting the   MANO    parameters   as    in
      \cite{hasson2019learning}.}
    \vspace{-6mm}
    \label{tab:tablePoseEst}
  \end{center}
  
\end{table}

%%%%%%%%%%%%%%%%%%%%%%%%%%%%%%%%%%%%%%%%%%%%%%%%%%%%%%%%%%%%%%%%%%%%%%%%%%%%%%%%

\subsection{Evaluation of Hand Pose Prediction Method}
%% \subsection{Evaluation of the Single Frame Hand Pose Prediction Method}

We  trained our  single frame  hand  pose prediction  method explained  in
Section~\ref{sec:handposeSingleFrameSec} on 66,034  frames from our \datasetname
dataset.  We evaluated it  on a test set of 13  sequences captured from different
viewpoints  and  totaling  11,524   frames.  The  test  set  sequences also  contain
subjects and objects  not present  in  the training  set.

We  report three  different metrics  from previous  works: Mean  joint position
error after aligning the position of the root joint and global scale with ground
truth~\cite{Zimmermann2017}; Mesh error measuring the average Euclidean distance
between predicted  and ground truth mesh vertices~\cite{zimmermann2019freihand};  and the
$F$-score~\cite{zimmermann2019freihand},  defined as  the harmonic  mean between
recall and  precision between two meshes  given a distance threshold.   The mesh
error  and \textit{F}-score  are obtained  after aligning  the predicted  meshes
using  Procrustes alignment  with the  ground truth  meshes and  hence does  not
measure the accuracy of wrist joint rotation.  The mean joint error on the other
hand considers wrist joint location  as the 3D points  are not rotated
before evaluation.

To understand the effect of joint direction predictions on the overall accuracy,
we evaluate  the results  of the  MANO fitting  by dropping  the second  term in
Eq.~\eqref{eq:manoFitEst}.  We also compare our  results with the hand branch of
\cite{hasson2019learning}, a  very recent work  that predicts the MANO  pose and
shape parameters directly from a single RGB image, retrained on our dataset.  As
shown in Table~\ref{tab:tablePoseEst}, predicting joint directions along with 2D
joint locations  significantly improves the  hand pose estimation  accuracy.  It
can also be inferred that predicting 2D hand joint locations and fitting MANO model to them is  more   accurate  than   direct   MANO  parameter   predictions  as   in
\cite{hasson2019learning}.      Qualitative     results     are     shown     in
Fig.~\ref{fig:baseline_qualitative}. The last row shows that our method robustly
predicts hand poses even when interacting with unknown objects.

\comment{
  \begin{figure}
    \begin{center}

      \includegraphics[page=17, width=0.3\linewidth, trim=6cm 1.35cm 6cm 1.9cm,clip]{figures/singleFramePose/handposes}
      \includegraphics[page=9, width=0.3\linewidth, trim=6cm 1.35cm 6cm 1.9cm,clip]{figures/singleFramePose/handposes}
      \includegraphics[page=6, width=0.3\linewidth, trim=6cm 1.35cm 6cm 1.9cm,clip]{figures/singleFramePose/handposes} \\
      \includegraphics[page=10, width=0.3\linewidth, trim=6cm 1.35cm 6cm 1.9cm,clip]{figures/singleFramePose/handposes}
      \includegraphics[page=19, width=0.3\linewidth, trim=6cm 1.35cm 6cm 1.9cm,clip]{figures/singleFramePose/handposes}
      \includegraphics[page=7, width=0.3\linewidth, trim=6cm 1.35cm 6cm 1.9cm,clip]{figures/singleFramePose/handposes} \\
      \includegraphics[page=18, width=0.3\linewidth, trim=6cm 1.35cm 6cm 1.9cm,clip]{figures/singleFramePose/handposes}
      \includegraphics[page=20, width=0.3\linewidth, trim=6cm 1.35cm 6cm 1.9cm,clip]{figures/singleFramePose/handposes}
      \includegraphics[page=12, width=0.3\linewidth, trim=6cm 1.35cm 6cm 1.9cm,clip]{figures/singleFramePose/handposes}
    \end{center}
    \vspace{-2mm}
    \caption{Qualitative results for single RGB frame based hand pose estimation method. We recover hand poses when it is heavily occluded by objects and in cluttered scenes.}
    \label{fig:baseline_qualitative}
    \vspace{-5mm}
  \end{figure}
}

\begin{figure}
  \begin{center}
    
    \includegraphics[page=17, width=0.24\linewidth, trim=6cm 1.35cm 6cm 1.9cm,clip]{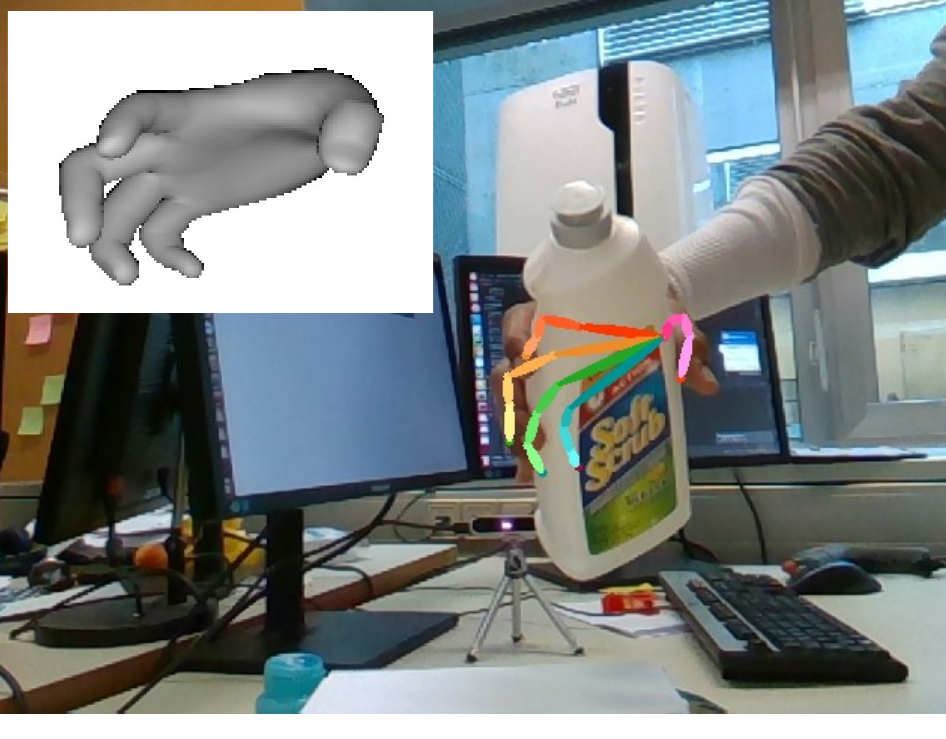}
    \includegraphics[page=9, width=0.24\linewidth, trim=6cm 1.35cm 6cm 1.9cm,clip]{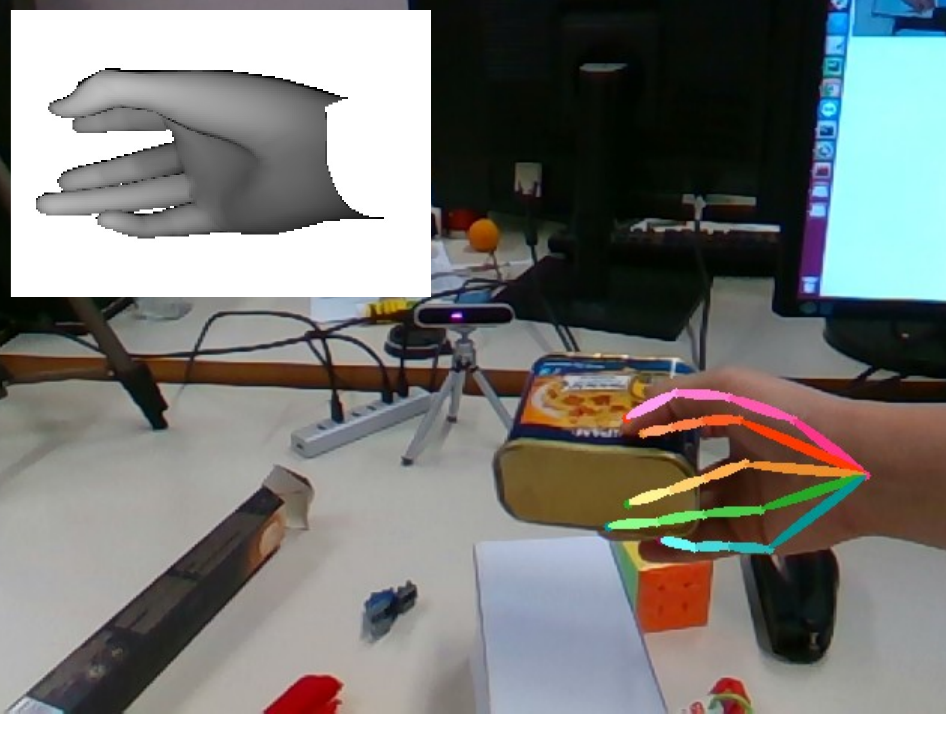}
    \includegraphics[page=6, width=0.24\linewidth, trim=6cm 1.35cm 6cm 1.9cm,clip]{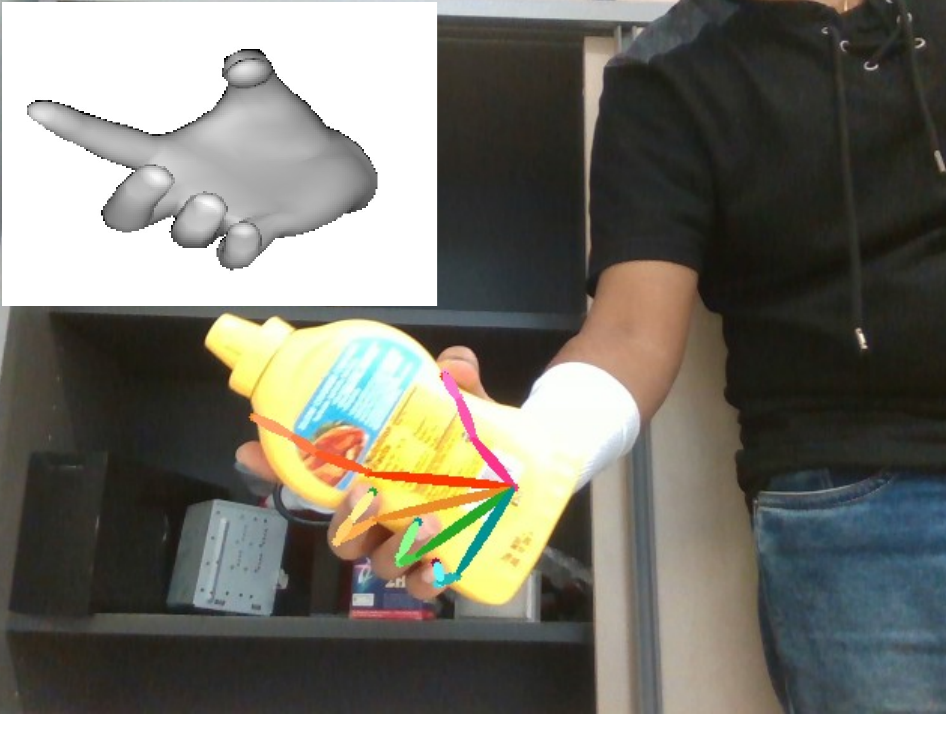} 
    \includegraphics[page=10, width=0.24\linewidth, trim=6cm 1.35cm 6cm 1.9cm,clip]{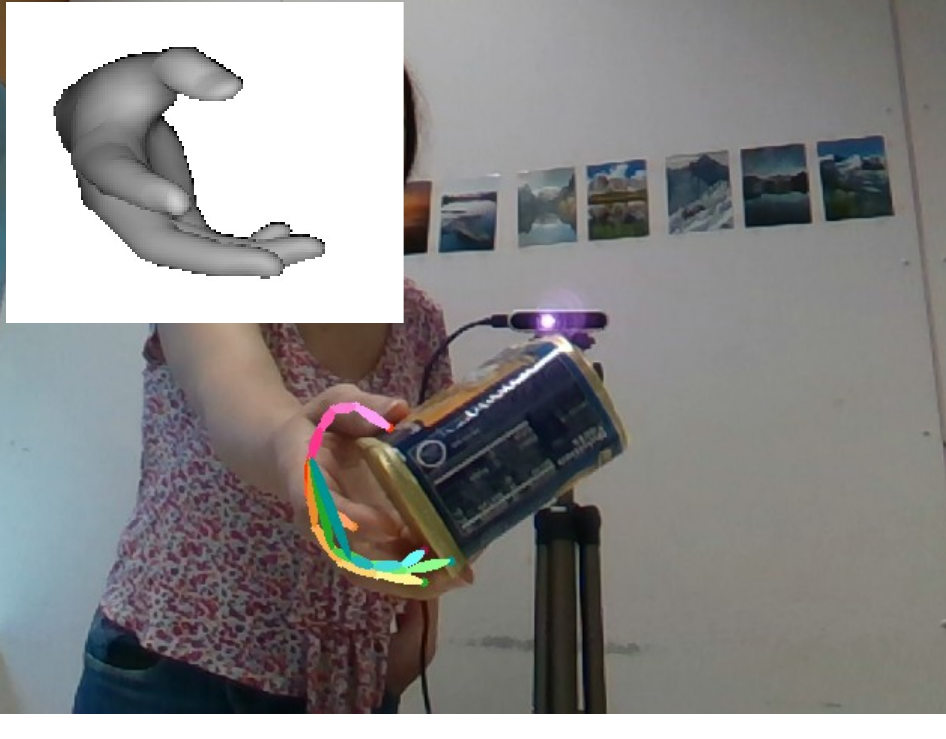}\\
    \includegraphics[page=19, width=0.24\linewidth, trim=6cm 1.35cm 6cm 1.9cm,clip]{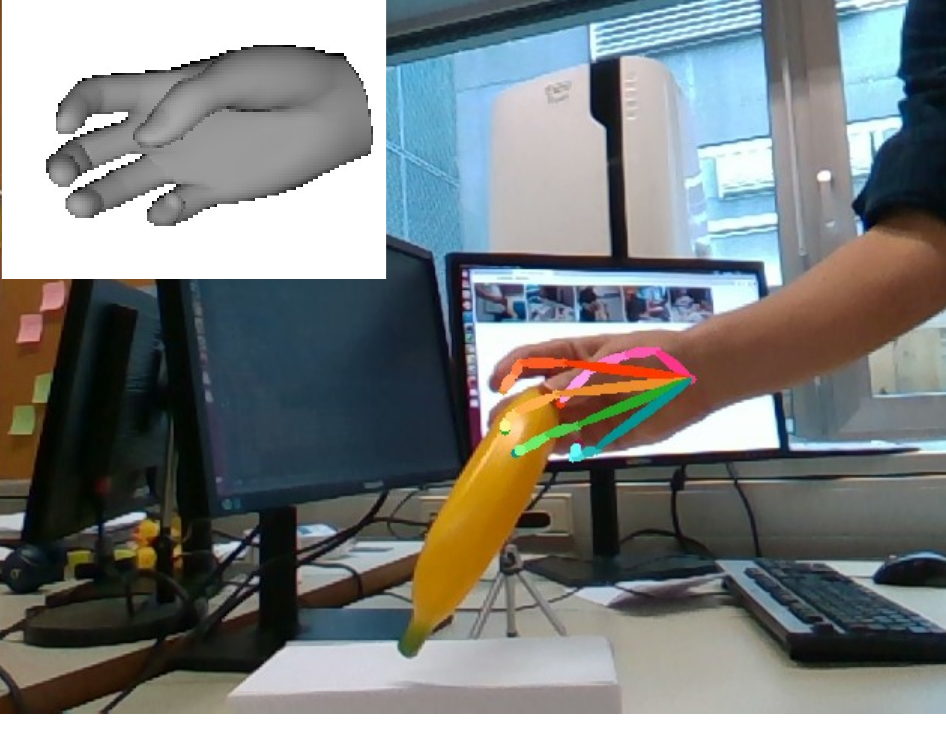}
    \includegraphics[page=7, width=0.24\linewidth, trim=6cm 1.35cm 6cm 1.9cm,clip]{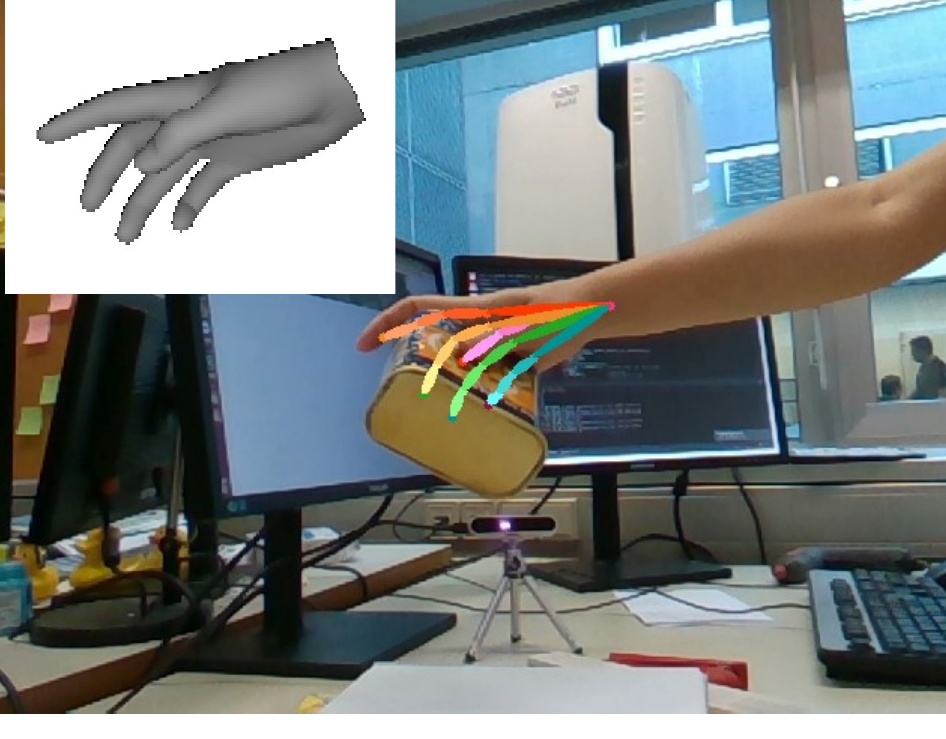} 
    \includegraphics[page=18, width=0.24\linewidth, trim=6cm 1.35cm 6cm 1.9cm,clip]{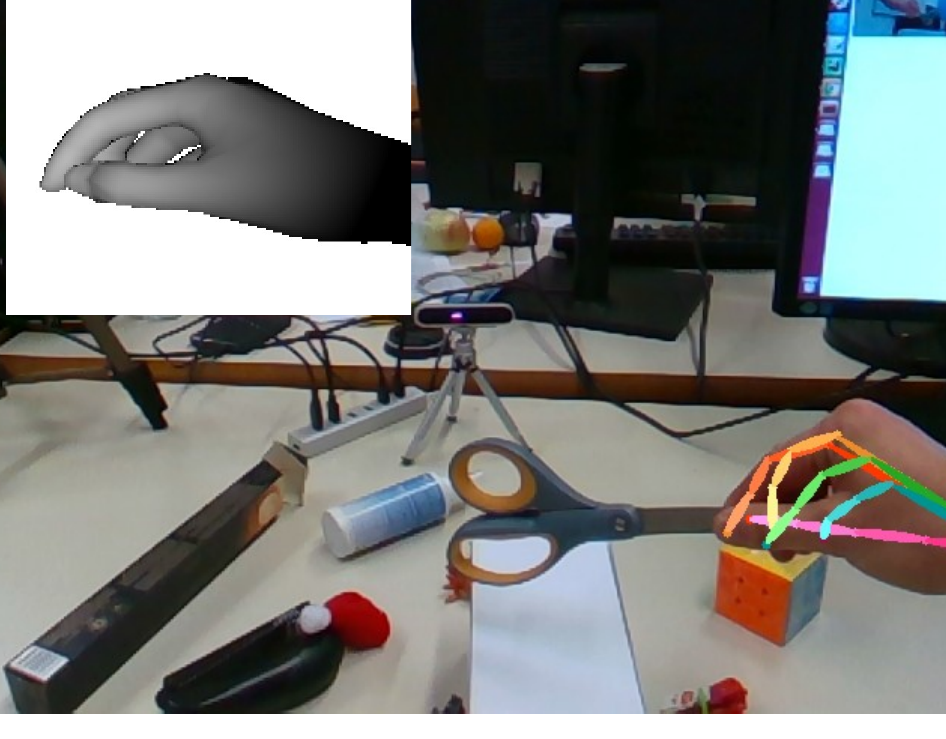}
    \includegraphics[page=20, width=0.24\linewidth, trim=6cm 1.35cm 6cm 1.9cm,clip]{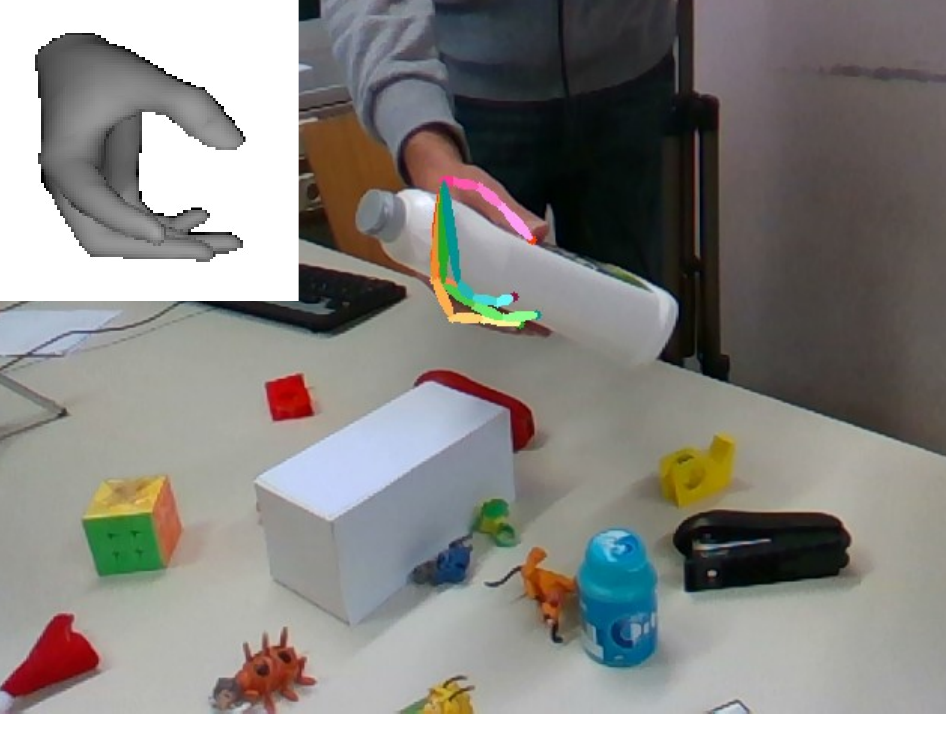}\\
    \includegraphics[page=22, width=0.24\linewidth, trim=6cm 1.35cm 6cm 1.9cm,clip]{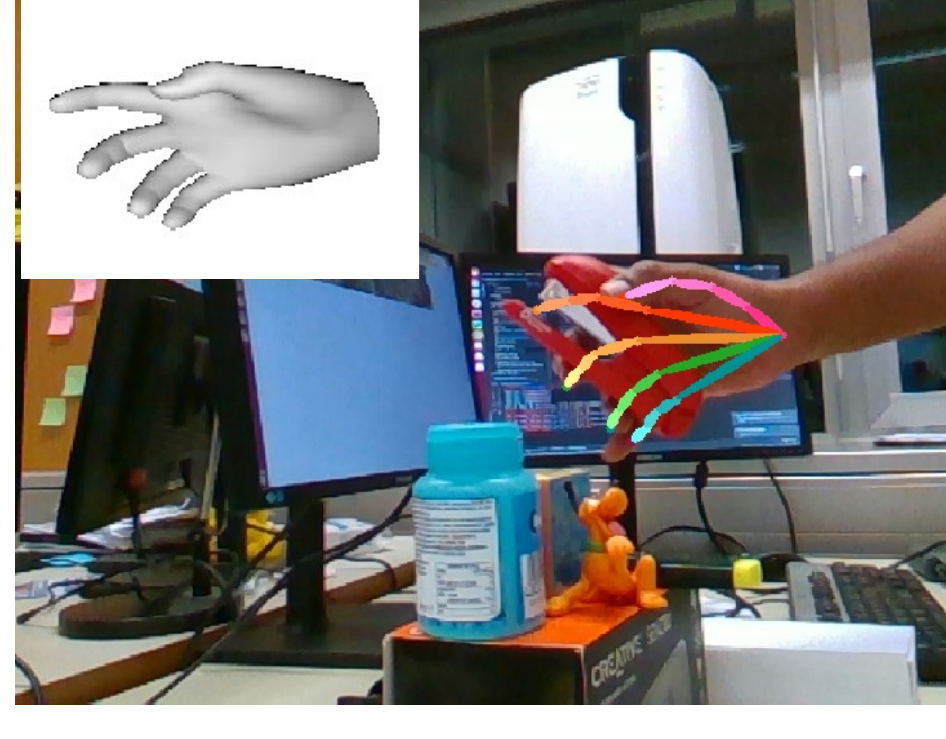}
    \includegraphics[page=24, width=0.24\linewidth, trim=6cm 1.35cm 6cm 1.9cm,clip]{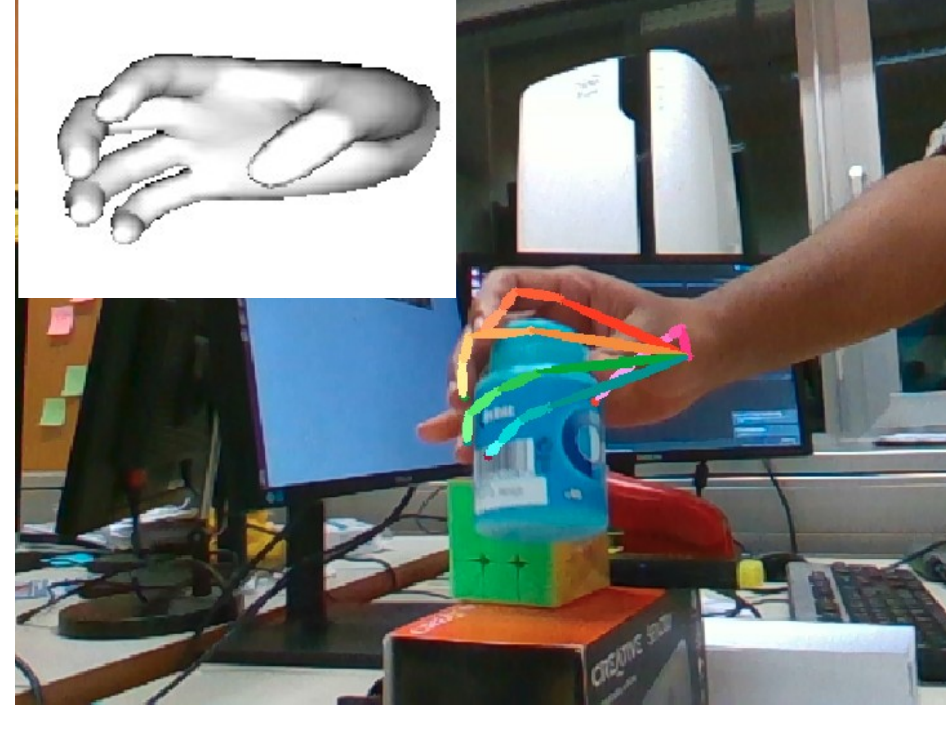}
    \includegraphics[page=25, width=0.24\linewidth, trim=6cm 1.35cm 6cm 1.9cm,clip]{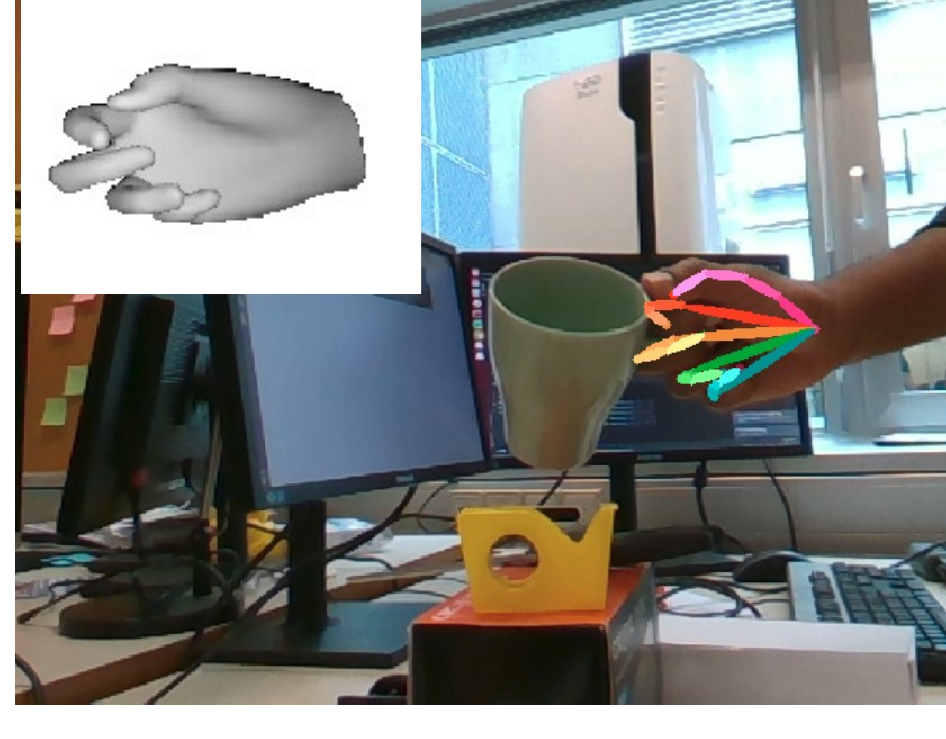}
    \includegraphics[page=28, width=0.24\linewidth, trim=6cm 1.35cm 6cm 1.9cm,clip]{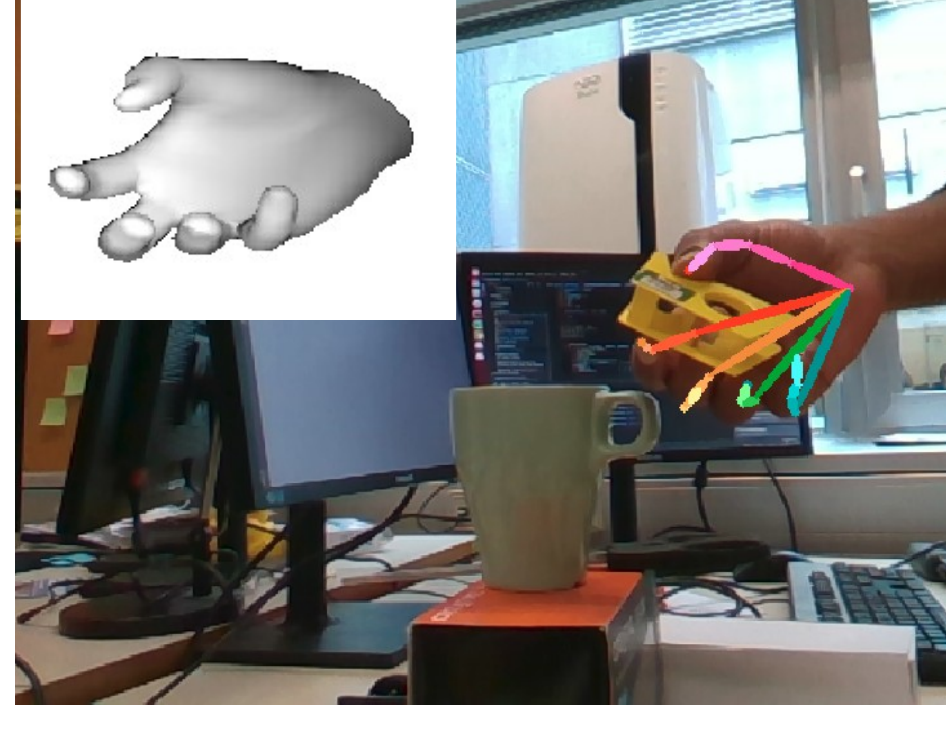}
  \end{center}
  \vspace{-2mm}
  \caption{Qualitative results of  our single color image  hand pose estimation
    method. It can recover hand poses even when the hand is heavily occluded by objects and in
    cluttered scenes. The last row shows it can handle unseen objects.}
  \label{fig:baseline_qualitative}
  \vspace{-5mm}
\end{figure}

%%%%%%%%%%%%%%%%%%%%%%%%%%%%%%%%%%%%%%%%%%%%%%%%%%%%%%%%%%%%%%%%%%%%%%%%%%%%%%%%

\comment{
  \paragraph{PCK metric: Percentage of Correct Keypoints in a Side View.}
  We use  a second  RGB camera registered  with respect  to the  RGB-D camera  used to
  capture  the sequences.   This camera  has a  viewpoint that  is approximately
  orthogonal to  the RGB-D  camera.  We use  the first camera  to obtain  the pose
  annotations and the second camera for  validation only. We manually annotate the
  hand pose  with the 2D  joint locations of visible  joints in the  second camera
  view for 10 randomly chosen frames  of 15 different sequences, totaling over 150
  frames.   Further, we  annotated  the object  pose by  providing  the 2D  corner
  locations of visible selected points in the second camera view totaling over 100
  frames.

  To evaluate the PCK metric, we project  the estimated 3D pose annotations on the
  second camera, and  compute the 2D distances between the  manual annotations and
  the projected automatic  annotations. Fig.~\ref{fig:anno_frameswithin} shows the
  PCK metric  for varying  distance thresholds, \ie, the  percentage of  3D points
  that have an error smaller than  a threshold. The final stage of the global optimization discussed in Section \ref{GO} increases the accuracy of annotations by about 7\% by refining the poses. Qualitative results before and after the final stage of global optimization are shown in Fig.~\ref{fig:global_qualitative}. More qualitative examples on different objects, persons and poses are shown in Fig.~\ref{fig:anno_qualitative}.

  \begin{figure}
    \begin{center}
      \includegraphics[width=0.45\linewidth]{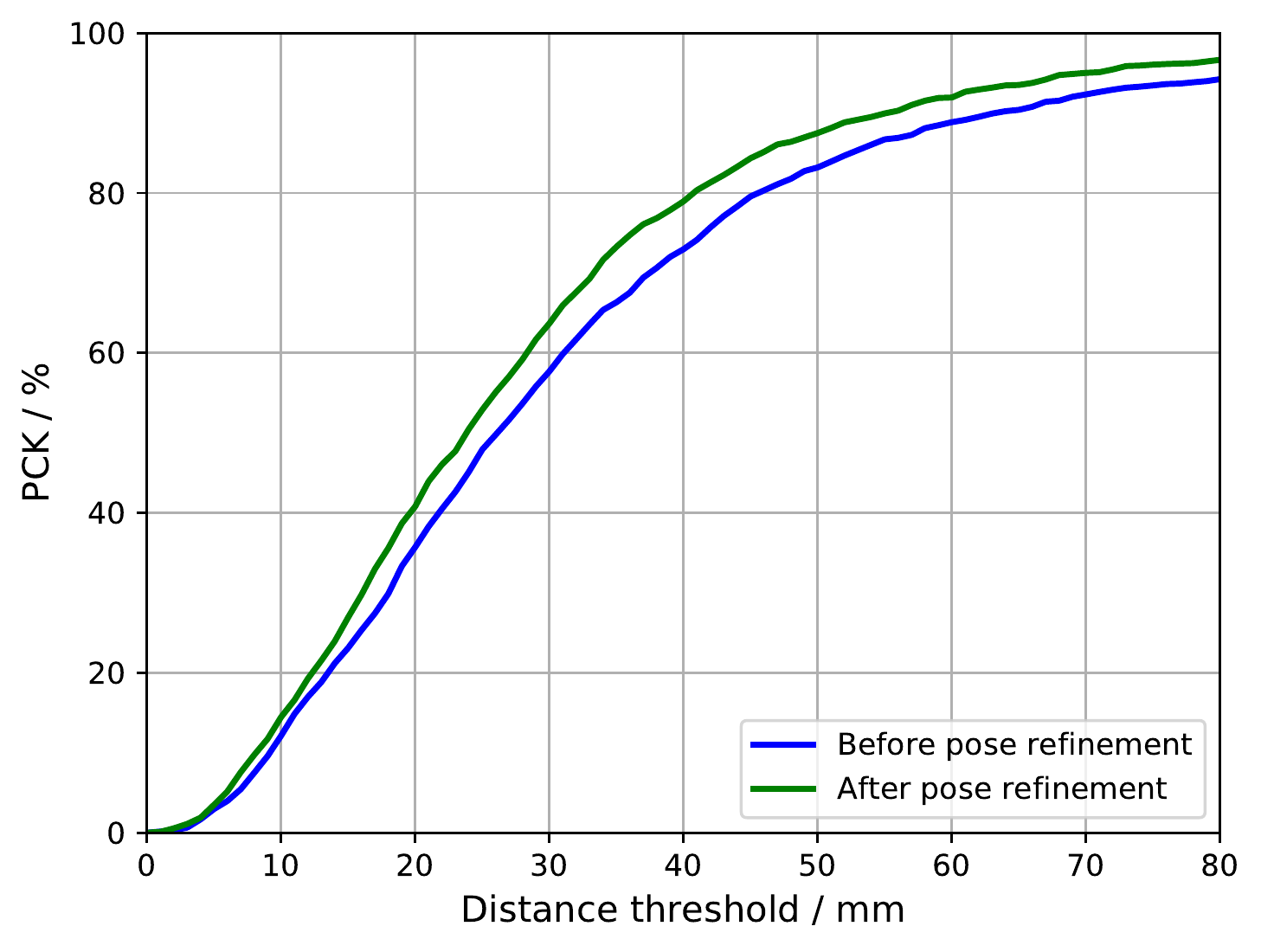}
      \includegraphics[width=0.45\linewidth]{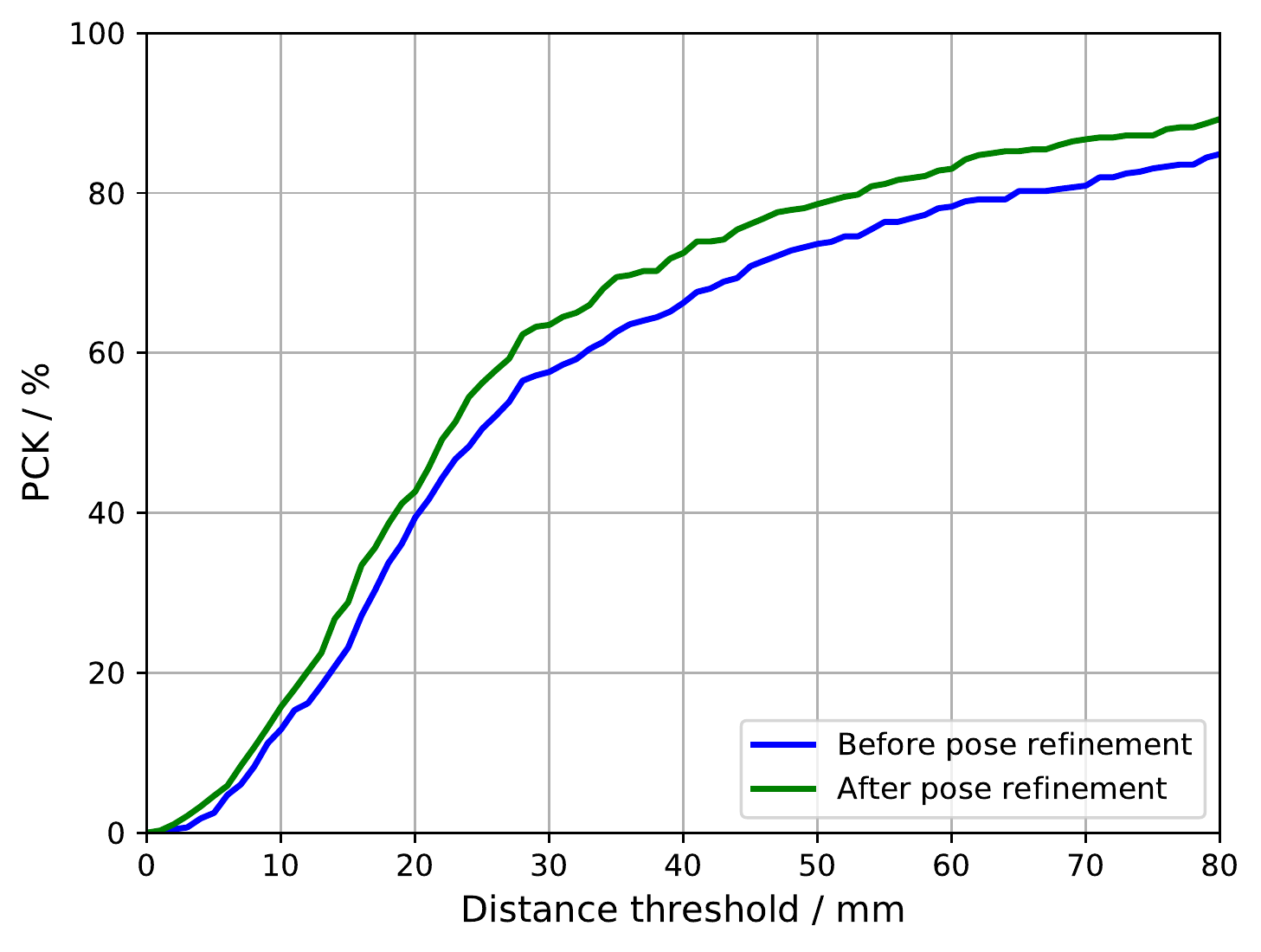}
    \end{center}
    \vspace{-5mm}
    \caption{PCK metric (before and after final refinement stage) for evaluating the proposed pose annotation method on 15 sequences with  varying distance thresholds.  Left: Hand.  Right: Object.}
    \label{fig:anno_frameswithin}
  \end{figure}

  \begin{figure}
    \begin{center}
      \includegraphics[width=0.28\linewidth]{figures/mask_0045pngimg2_SM1_b.png}
      \includegraphics[width=0.28\linewidth]{figures/mask_0304pngimg2_MC1_b.png}
      \includegraphics[width=0.28\linewidth]{figures/mask_0652jpgimg2_SM2_b.jpg} \\
      \includegraphics[width=0.28\linewidth]{figures/mask_0045jpgimg2_SM1_a.jpg}
      \includegraphics[width=0.28\linewidth]{figures/mask_0304pngimg2_MC1_a.jpg}
      \includegraphics[width=0.28\linewidth]{figures/mask_0652jpgimg2_SM2_a.jpg}
    \end{center}
    \vspace{-2mm}
    \caption{Qualitative results of the proposed annotation
      method. Each column shows the 3D models of the hand and
      object superimposed on the images from the side view camera
      before and after refinement stage, respectively. The final
      refinement still improves the estimated poses.}
    \vspace{-6mm}
    \label{fig:global_qualitative}
  \end{figure}

  \begin{figure}
    \hspace{-2mm}
    \begin{center}
      \includegraphics[width=0.6\linewidth]{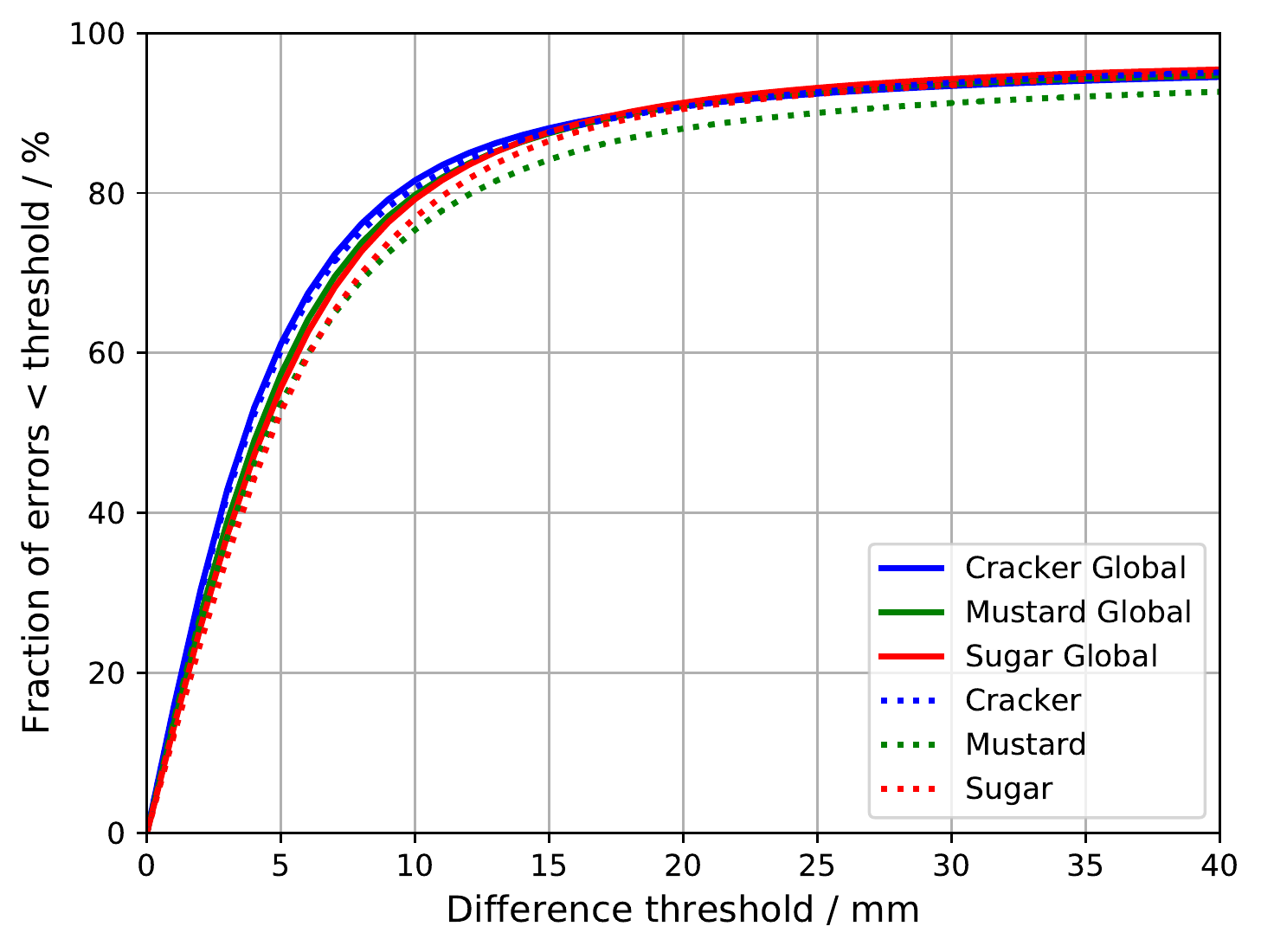}
    \end{center}
    \vspace{-4mm}
    \caption{Reconstruction error. We plot the fraction of depth residuals where the depth residual is smaller than a threshold. A large fraction of errors are well below 10~mm, which shows that the articulated models can explain the captured data accurately.}
    \label{fig:depth_errors}
  \end{figure}

  \begin{figure}
    \begin{center}
      \includegraphics[width=0.65\linewidth]{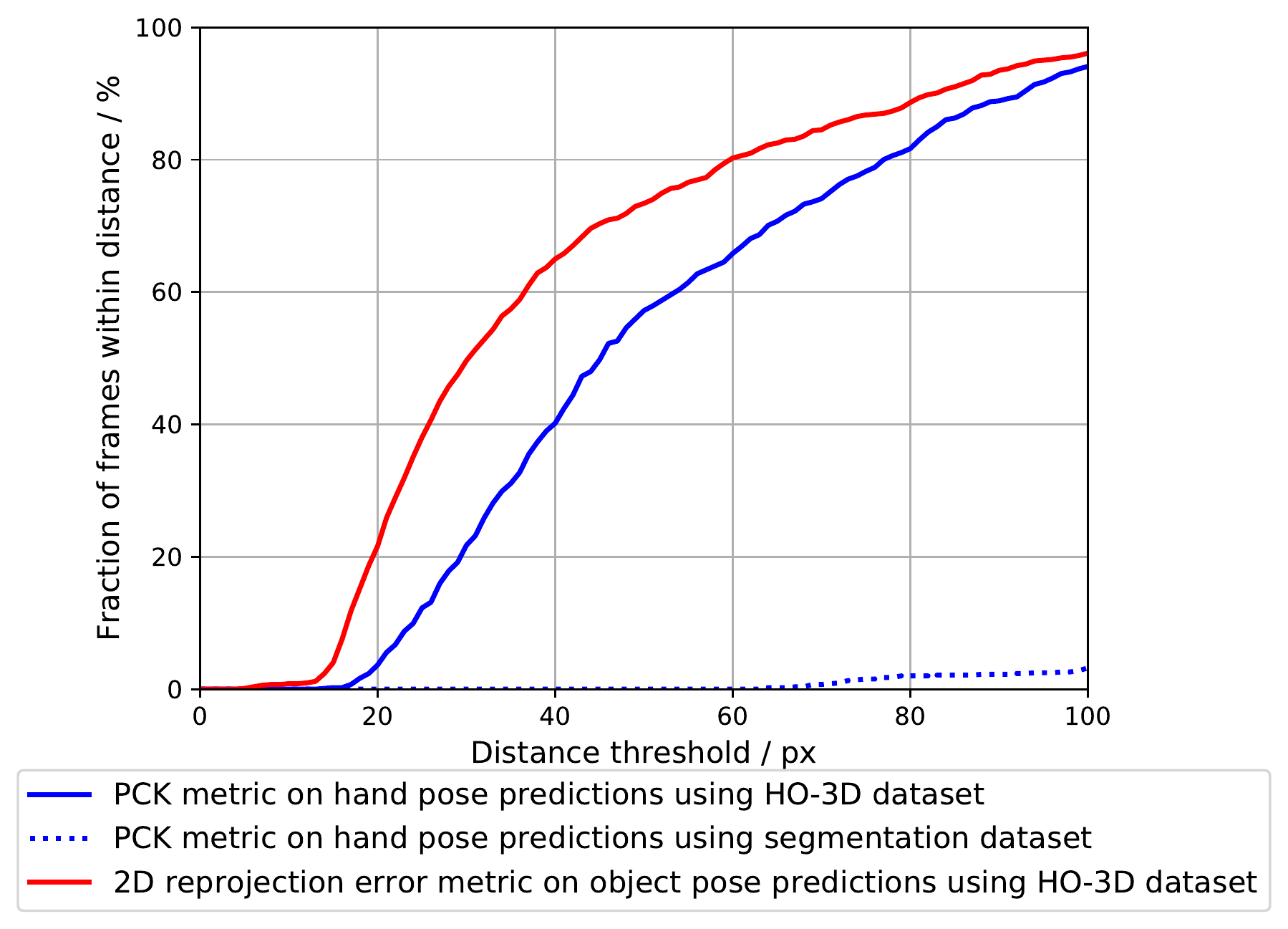}
    \end{center}
    \vspace{-5mm}
    \caption{Evaluation of our baseline for hand pose estimation. We plot the percentage of correct keypoints~(PCK) over distance threshold. The dashed curve shows that the synthetic images used to train the image segmentation are clearly not sufficient to train this baseline, even if they work well for the segmentation task. }
    \label{fig:baseline_hand}
    \vspace{-5mm}
  \end{figure}
  %\vspace{-20mm}
  %\begin{figure}
  %\begin{center}
  %\includegraphics[width=0.32\linewidth]{figures/annotated_Shreyas_Mustard_1_61}
  %\includegraphics[width=0.32\linewidth]{figures/annotated_Sinisa_Sugar_1_351_hand}
  %\includegraphics[width=0.32\linewidth]{figures/annotated_Markus_cracker_2_81_hand}
  %\includegraphics[width=0.32\linewidth]{figures/annotated_Sinisa_Sugar_1_161_obj}
  %\includegraphics[width=0.32\linewidth]{figures/annotated_Markus_cracker_1_751_obj}
  %\includegraphics[width=0.32\linewidth]{figures/annotated_Shreyas_Sugar_1_401_obj}
  %%\includegraphics[width=0.32\linewidth]{figures/annotated_Sinisa_Sugar_1_841_obj}
  %%\includegraphics[width=0.32\linewidth]{figures/annotated_Sinisa_Sugar_1_841_obj}
  %%\includegraphics[width=0.32\linewidth]{figures/annotated_Sinisa_Sugar_1_841_obj}
  %\end{center}
  %\caption{Qualitative comparison between manual annotations and our annotations. Manual annotations are in grayscale, our automatic annotations in color. Top: Hand comparison. Bottom: Object comparison.}
  %\label{fig:anno_qualitative}
  %\end{figure}

  \vspace{-6mm}

  \paragraph{Reconstruction error.}
  As an additional  evaluation, we rely on  a reconstruction error.  To  do so, we
  compute the  difference between the  captured depth maps  and the the  depth maps
  generated  using the  rendered  models of  the  hand and  the  object under  the
  estimated  poses.    Fig.~\ref{fig:depth_errors}  shows  the   evaluation for three different objects \textit{Mustard bottle, Cracker box} and \textit{Sugar box}.   The
  majority of all  errors, around 90\%, are below 10~mm.  The remaining errors can
  be attributed to  depth noise from the  sensor and the wrist  and forearm, which
  are not present in our model.
}

%------------------------------------------------------------------------

\section{Conclusion}

We introduced a fully automatic method to annotate images of a hand manipulating
an  object with  their  3D poses,  even under  large  occlusions, by  exploiting
temporal  consistency. We also introduced the first  markerless  dataset of  color images  for
benchmarking 3D  hand+object pose estimation.  To demonstrate the  usefulness of
our dataset, we  proposed a method  for predicting the 3D pose  of the hand
from  a single  color  image. Another  future  application is  the  joint estimation  of
hand+object poses from a single RGB frame.

The lack  of high quality  segmentations (we had to  use a synthetic  dataset as
explained in  Section~\ref{sec:method}) of the hand  sometimes affects accuracy.
Improving  these segmentations  and/or introducing  a attraction  term and  physics
constraints as  in~\cite{RealtimeHO_ECCV2016, Tzionas2016} would  further
improve our annotations.

%------------------------------------------------------------------------

\section{Acknowledgement}
This work  was supported  by the  Christian Doppler  Laboratory for  Semantic 3D
Computer Vision, funded in part by Qualcomm Inc.

\title{HOnnotate: A method for 3D Annotation of Hand and Object Poses \\ Supplementary Material}
\maketitle
\setcounter{section}{0}
%% -*- mode: latex; mode: flyspell -*-
\renewcommand{\thesection}{S.\arabic{section}}

\section{Hand Pose Estimation from Single Color Image}

\begin{figure*}
	\begin{center}
		\includegraphics[width=0.95\linewidth]{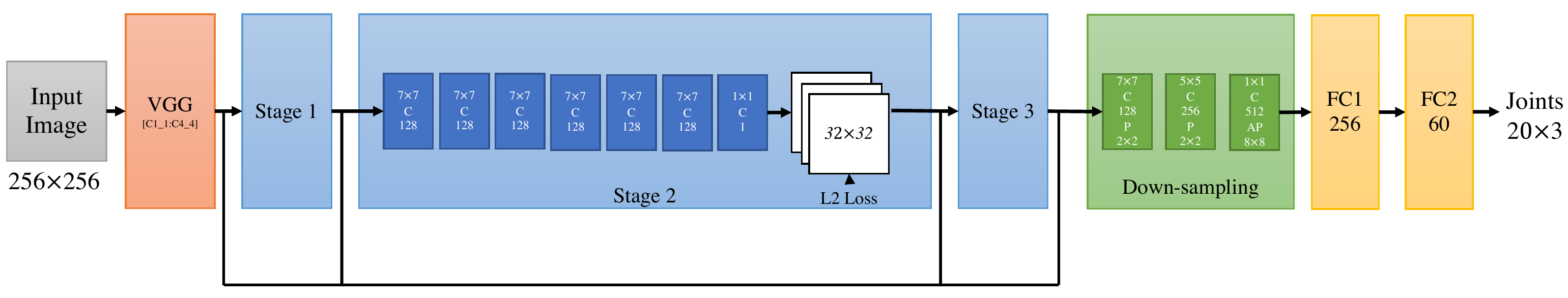}
	\end{center}
	\vspace{-2mm}
	\caption{Architecture of our hand pose estimator from a single color
          image. Given an input image of hand centered in the image, we extract
          the features using the convolutional layers of VGG~\cite{Simonyan14c}
          (\texttt{Conv1\_1} to \texttt{Conv4\_4}). Similarly
          to~\cite{Cao2017realtime}, we then predict heatmaps for the joint
          locations in multi-stages. The architecture for the different stages are all the same.  \texttt{C} denotes a convolutional layer with the number of filters and the filter size inscribed; \texttt{FC}, a fully-connected layer with the number of neurons; \texttt{P} and \texttt{AP} denote max-pooling and average pooling with their sizes, respectively.}
	\label{fig:net_arch}
\end{figure*}

Fig.~\ref{fig:net_arch} shows the architecture of our hand pose estimator from a
single frame. Given an image of the hand centered in the image window, we first
extract features using the convolutional layers of VGG~\cite{Simonyan14c},  and
then similar to \cite{Cao2017realtime} using a multi-stage CNN, we predict
heatmaps for the 2D hand joint locations and finally joint direction vectors with respect to wrist joint. The hand detection can be done using segmentation as described in Section~\ref{sec:hand_obj_seg}. %In the video sequence scenario, based on the 2D predictions of joints in the previous frame, the 2D bounding box can be calculated in the current frame. 

%% Fig.~\ref{fig:net_arch} illustrates the architecture of our hand pose estimator from a single frame. Given an image of the hand centered in the image window, we first extract features using convolutional layers of VGG~\cite{Simonyan14c},  and then similar to \cite{Cao2017realtime} using multi-stage CNN, predict heatmaps of 2D hand joint locations and finally joint direction vectors with respect to wrist joint. We trained the network using Momentum optimizer with a learning rate of 0.01, a batch size of 64 for 100 epochs. 
%% The hand detection can be done using segmentation as described in Section~\ref{sec:hand_obj_seg}. In the video sequence scenario, based on the 2D predictions of joints in the previous frame, the 2D bounding box can be calculated in the current frame. 

\section{Hand Pose Estimation for Hand Interaction with Unseen Objects}
Knowing the objects in advance can help to improve the performances of the
estimated 3D hand pose while hand interacts with objects, however, in practice,
the hand can manipulate any arbitrary objects. We have tested our hand pose
estimator trained on our annotations, and tested on sequences where a hand is
manipulating objects not present in the annotated images. As shown in Fig.~\ref{fig:unseen_objects}, our pose estimator performs well on these sequences.

\section{Hand Shape Estimation}
The MANO hand shape parameters $\beta \in \IR^{10}$ were estimated for each human manipulator in our \datasetname dataset. The shape parameters are estimated from a sequence $\Phi$ of hand only poses using a method similar to \cite{Tan_2016_CVPR} in two steps. More exactly, the pose of hand $p_h^t$ in the sequence is first estimated for each frame $t$ using a mean pose $\beta_{mean}$ as $\hat{p}_h^{t} = \argmin_{p_h} E_{H}(p_h, \beta_{mean})$, where, 
\begin{align}
E_{H}(p_h, \beta_{mean}) = &E_\mathcal{D}(p_h, \beta_{mean}) + \epsilon E_\joint(p_h) + \\ \nonumber 
&\eta E_\tc(p_h, p_h^{t-1}, p_h^{t-2}).
\end{align}
$E_\mathcal{D}(p_h, \beta_{mean})$ represents the data term defined in Eq. 2 of the paper where hand is rendered with pose parameters $p_h$ and shape parameters $\beta_{mean}$. $E_\joint$ and $E_\tc$ are explained in Section 3.2 of the paper. At each frame, the pose parameters are initialized with $p_h^{t-1}$. 
The personalized hand shape parameters are then obtained as,
\begin{equation}
\beta^* = \argmin_{\beta} \sum_{t\in \Phi} \min_{p_h^t} E_{H}(p_h^t, \beta),
\end{equation}
where the pose parameters are initialized with the values obtain in the first step ($\hat{p}_h^{t}$).

\section{Joint Angle Constraints}
The maximum and minimum limits on the joint angle parameters used in Eq.~(8) of the paper are provided in  Table~\ref{tab:table1}.
\begin{table}
	\begin{center}
		\resizebox{0.47\textwidth}{!}{%
			\begin{tabular}{lccccc}
				\toprule
				Joint & Index & Middle & Pinky  & Ring & Thumb\\
				\midrule
				MCP & \begin{tabular}{@{}c@{}}
					(0.00, 0.45)\\
					(-0.15, 0.20)\\
					(0.10, 1.80)\\
				\end{tabular} & \begin{tabular}{@{}c@{}}
					(0.00, 0.00)\\
					(-0.15, 0.15)\\
					(0.10, 2.00)\\
				\end{tabular} & \begin{tabular}{@{}c@{}}
					(-1.50, -0.20)\\
					(-0.15, 0.60)\\
					(-0.10, 1.60)\\
				\end{tabular} & \begin{tabular}{@{}c@{}}
					(-0.50, -0.40)\\
					(-0.25, 0.10)\\
					(0.10, 1.80)\\
				\end{tabular} & \begin{tabular}{@{}c@{}}
					(0.00, 2.00)\\
					(-0.83, 0.66)\\
					(0.00, 0.50)\\
				\end{tabular} \\
				\hline
			
				PIP & \begin{tabular}{@{}c@{}}
					(-0.30, 0.20)\\
					(0.00, 0.00)\\
					(0.00, 0.20)\\
				\end{tabular} & \begin{tabular}{@{}c@{}}
					(-0.50, -0.20)\\
					(0.00, 0.00)\\
					(0.00, 2.00)\\
				\end{tabular} & \begin{tabular}{@{}c@{}}
					(0.00, 0.00)\\
					(-0.50, 0.60)\\
					(0.00, 2.00)\\
				\end{tabular} & \begin{tabular}{@{}c@{}}
					(-0.40, -0.20)\\
					(0.00, 0.00)\\
					(0.00, 2.00)\\
				\end{tabular} & \begin{tabular}{@{}c@{}}
					(-0.15, 1.60)\\
					(0.00, 0.00)\\
					(0.00, 0.50)\\
				\end{tabular} \\
				\hline	
				DIP & \begin{tabular}{@{}c@{}}
					(0.00, 0.00)\\
					(0.00, 0.00)\\
					(0.00, 1.25)\\
				\end{tabular} & \begin{tabular}{@{}c@{}}
					(0.00, 0.00)\\
					(0.00, 0.00)\\
					(0.00, 1.25)\\
				\end{tabular} & \begin{tabular}{@{}c@{}}
					(0.00, 0.00)\\
					(0.00, 0.00)\\
					(0.00, 1.25)\\
				\end{tabular} & \begin{tabular}{@{}c@{}}
					(0.00, 0.00)\\
					(0.00, 0.00)\\
					(0.00, 1.25)\\
				\end{tabular} & \begin{tabular}{@{}c@{}}
					(0.00, 0.00)\\
					(-0.50, 0.00)\\
					(-1.57, 1.08)\\
				\end{tabular} \\
				\bottomrule
		\end{tabular}}
		\caption{Empirically derived minimum and maximum values for the joint angle parameters used in our implementation.}
		\label{tab:table1}
	\end{center}
\end{table}

\section{Point Cloud from Multiple Cameras}
The $E_\icp$ term in Section~3.2 of the paper uses the combined point cloud $P$ from all the RGB-D cameras. Let $P_c$ denote the point cloud corresponding to camera $c$ and $M_{c_1,c_2}$ denote the relative pose between two cameras $c_1$ and $c_2$. The consolidated point cloud $P$ is then obtained as,
\begin{equation}
P = [P_0,~M_{c_1,c_0}\cdot P_1,~M_{c_2,c_0}\cdot P_2, ... ,~M_{c_{N},c_0}\cdot
  P_{N}] \> ,
\end{equation}
where $[\cdot, \cdot]$ represents concatenation of point clouds.

\section{Hand-Object Segmentation Network}
\label{sec:hand_obj_seg}
\begin{figure}
	\begin{center}
		\begin{tabular}{ccc}
			\includegraphics[width=0.32\linewidth]{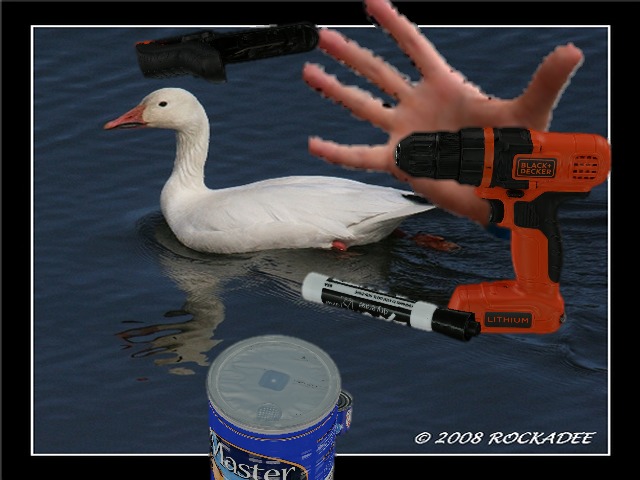} & \hspace{-4mm}
			\includegraphics[width=0.32\linewidth]{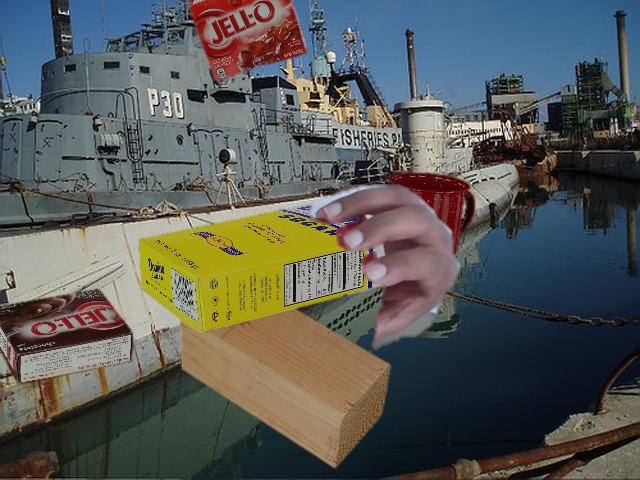} & \hspace{-4mm}
			\includegraphics[width=0.32\linewidth]{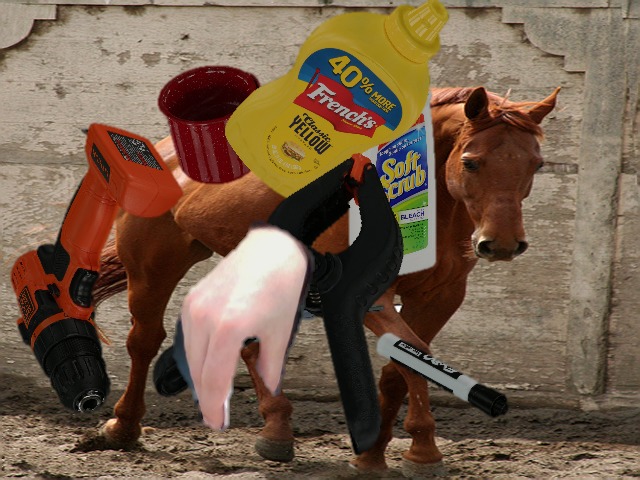}
		\end{tabular}
	\end{center}
	\vspace{-2mm}
\caption{Synthetic training images used for training the hand-object segmentation network.}
\label{fig:fig12}
\end{figure}

\begin{figure}
	\begin{center}
	\begin{tabular}{cc}
		\includegraphics[width=0.47\linewidth]{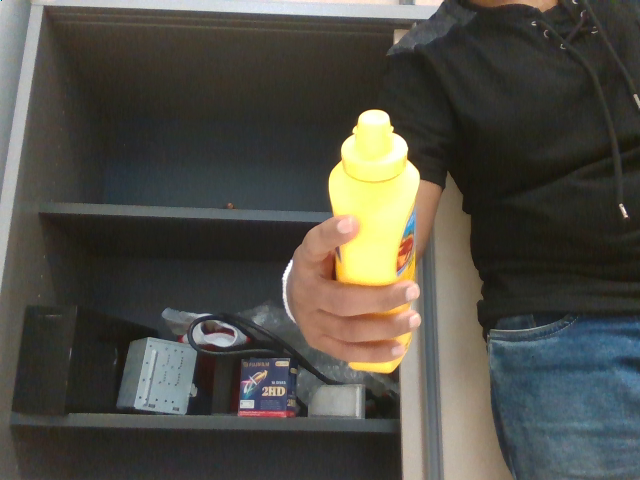} &  \hspace{-4mm}
		\includegraphics[width=0.47\linewidth]{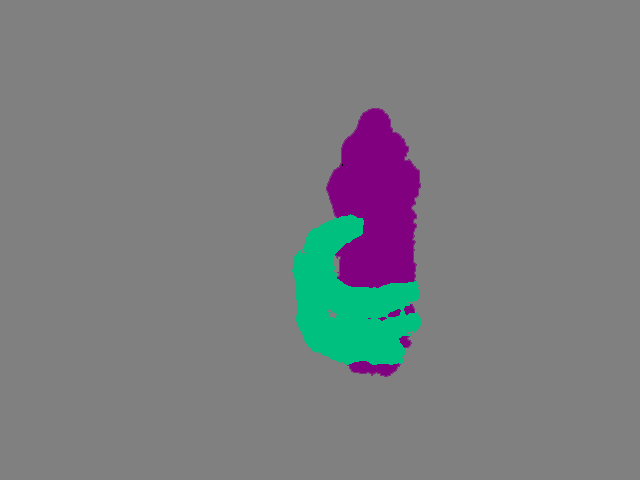} \\
		\end{tabular}
	\end{center}
	\vspace{-2mm}
	\caption{Example  of   hand  and   object  segmentation   obtained  with
		DeepLabV3.  Left: input image; Right: hand (green) and object (purple) 
		segmentation.}
	\vspace{-5mm}
	\label{fig:samples_seg}
\end{figure}

The segmentation maps for the hand and object are obtained from a DeepLabV3~\cite{deeplabv2017} network trained on synthetic images of hand and objects. The synthetic images are obtained by
over-laying and under-laying images of hands on images of objects at random locations and scales. We use the object masks provided by~\cite{ycbonline}.
The segmented hands were obtained using an RGB-D camera by applying simple depth
thresholding. We also use additional synthetic hand images from the RHD
dataset~\cite{Zimmermann2017}. A few example images from the training data are
shown in Fig.~\ref{fig:fig12}. We use 100K training images with
augmentations. Fig.~\ref{fig:samples_seg} shows segmentation of hand and object
using the trained DeepLabV3 network.

\section{Automatic Initialization}
\begin{figure}[ht]
	\begin{center}
		\includegraphics[width=0.8\linewidth]{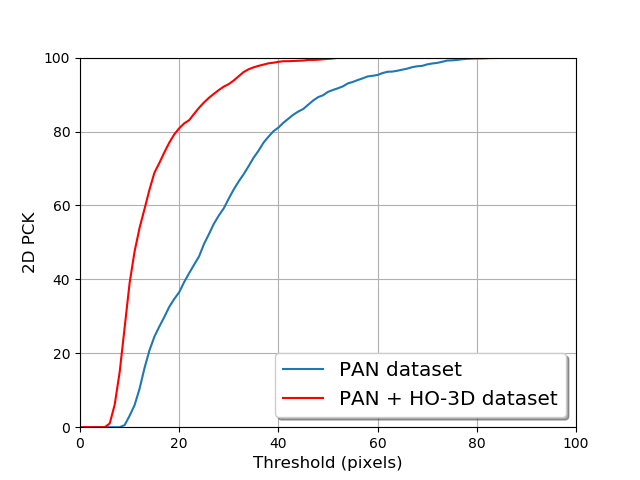}
	\end{center}
	\caption{Accuracy of keypoint prediction, described in~Section 3.2 of
          the paper when trained with PAN~\cite{JooSS18} dataset alone and PAN +
          our annotations. The accuracy is measured in percentage of correct 2D
          keypoints given a threshold. Only 15,000 images from our \datasetname dataset are used in training. Due to the presence of object occlusions, a network trained on hands-only dataset is less accurate in predicting keypoints when compared with a network trained with hand+object data.}
	\label{fig:pckCurve}
\end{figure}

As explained in Section~4.1 of the paper, a keypoint prediction network based on convolutional pose machine~\cite{Wei16} is used to obtain initialization for hand poses. Such a network is trained with our initial hand+object dataset of 15,000 images together with images from hand-only PAN~\cite{JooSS18} dataset. Fig. \ref{fig:pckCurve} compares the accuracy of network in predicting keypoints in hand-object interaction scenarios when trained with hands-only dataset and hands+object dataset. Our initial \datasetname dataset helps in obtaining a more accurate network for predicting keypoints and hence results in better initialization.

\section{Dataset Details}
We annotated 77,558 frames of 68 sequences hand-object interaction of 10 persons with different hand shape. On average there are 1200 frames per sequences. 16 sequences are captured and annotated in a single camera, and 52 sequences for the multi-camera setup. 

\paragraph{Hand+Object.}
The participants are asked to perform actions with objects. 
The grasp poses vary between frames in a sequence in the multi-camera setup and remain almost rigid in the single camera setup.

\paragraph{Participants.}
The participants are between 20 and 40 years old, 7 of them are males and 3 are
females. In total, 10 hand shapes are considered for the annotations.

\paragraph{Objects.}
We aimed to choose 10 different objects from the YCB dataset~\cite{posecnn2018} that are used in daily life. As shown in Fig.~\ref{fig:ycb_objects},
we have a wide variety of sizes such as large objects (e.g. Bleach)  that cause large hand occlusion, or the objects that make grasping and manipulation difficult (e.g. Scissors), while these are not the case in the existing hand+object datasets.

\paragraph{Multi-Camera Setup.} We use 5 calibrated RGB-D cameras, in our multi-camera setup. The cameras are located at different angles and locations. Our cameras are synchronized with a precision of about 5 ms. The scenes are cluttered with objects, and the backgrounds vary between scenes.

Figs.~\ref{fig:HO_3D_1} and \ref{fig:HO_3D_2} show some examples of the 3D annotated frames for both hand and object from our proposed dataset, HO-3D.

\begin{figure*}
	\begin{center}
		\begin{tabular}{cccc}
			\includegraphics[width=0.24\linewidth,trim=0.8cm 3cm 0.8cm 3cm,clip]{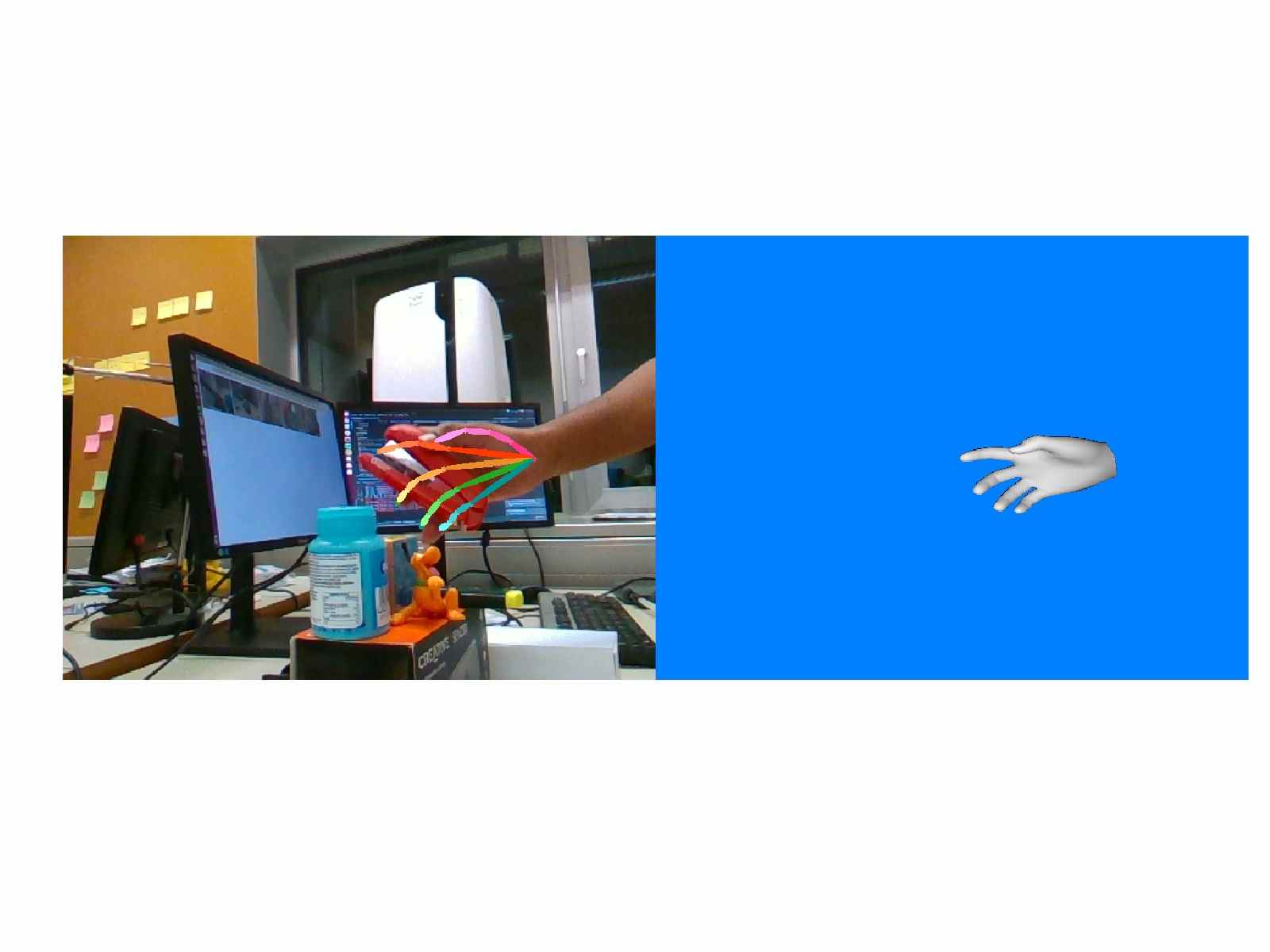} & \hspace{-3mm}\includegraphics[width=0.24\linewidth,trim=0.8cm 3cm 0.8cm 3cm,clip]{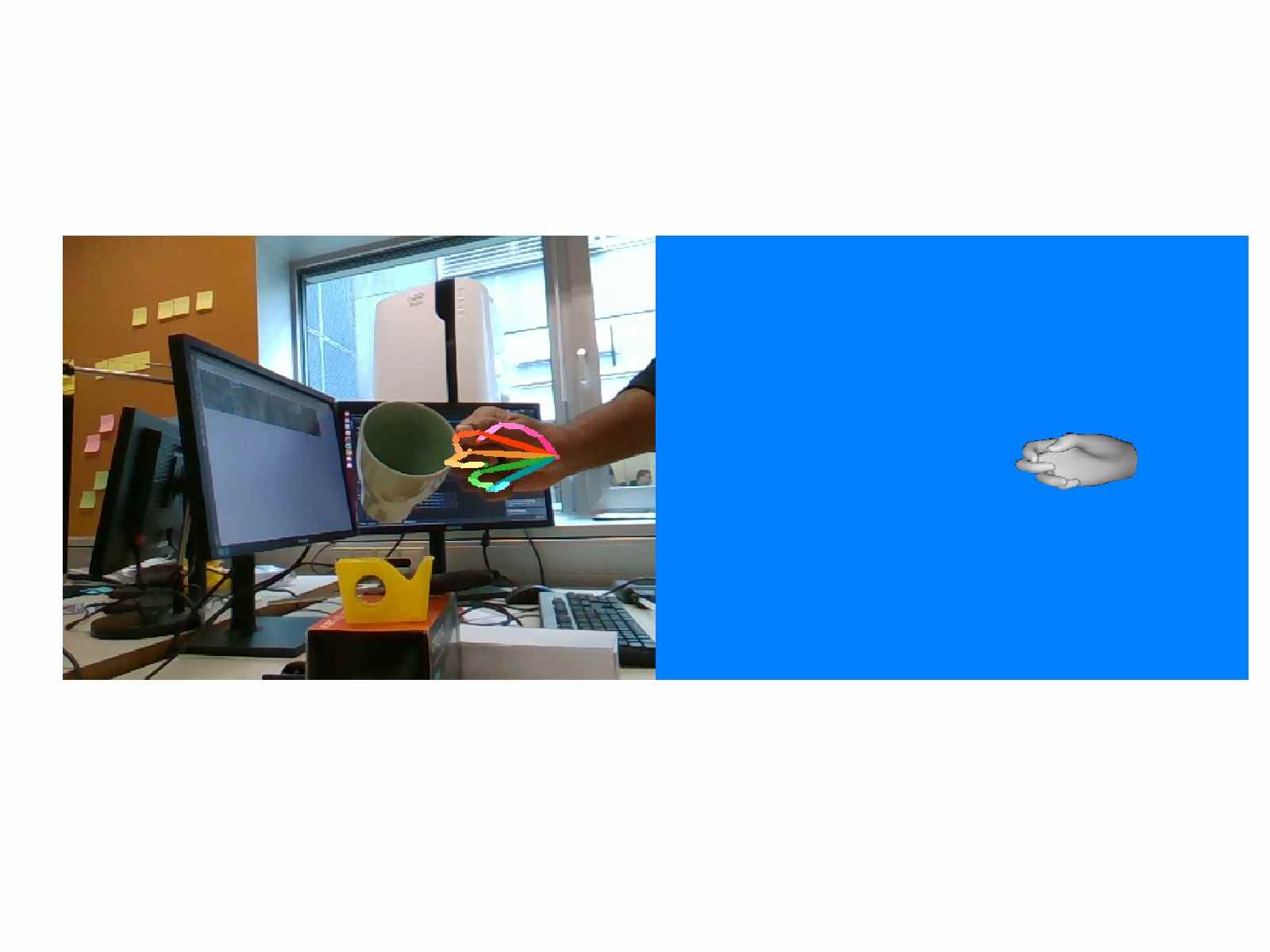} & \hspace{-3mm}\includegraphics[width=0.24\linewidth,trim=0.8cm 3cm 0.8cm 3cm,clip]{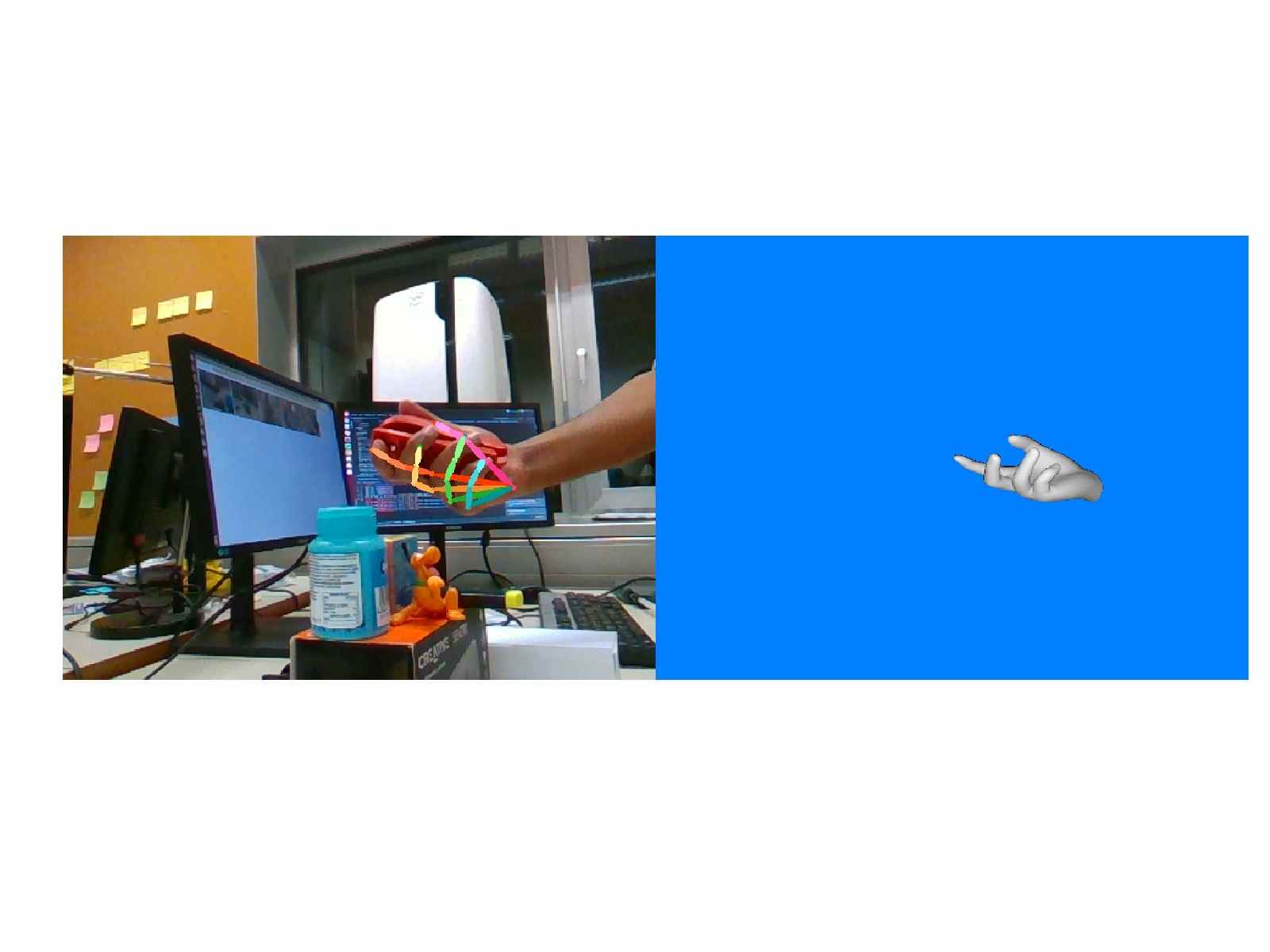} & \hspace{-3mm}\includegraphics[width=0.24\linewidth,trim=0.8cm 3cm 0.8cm 3cm,clip]{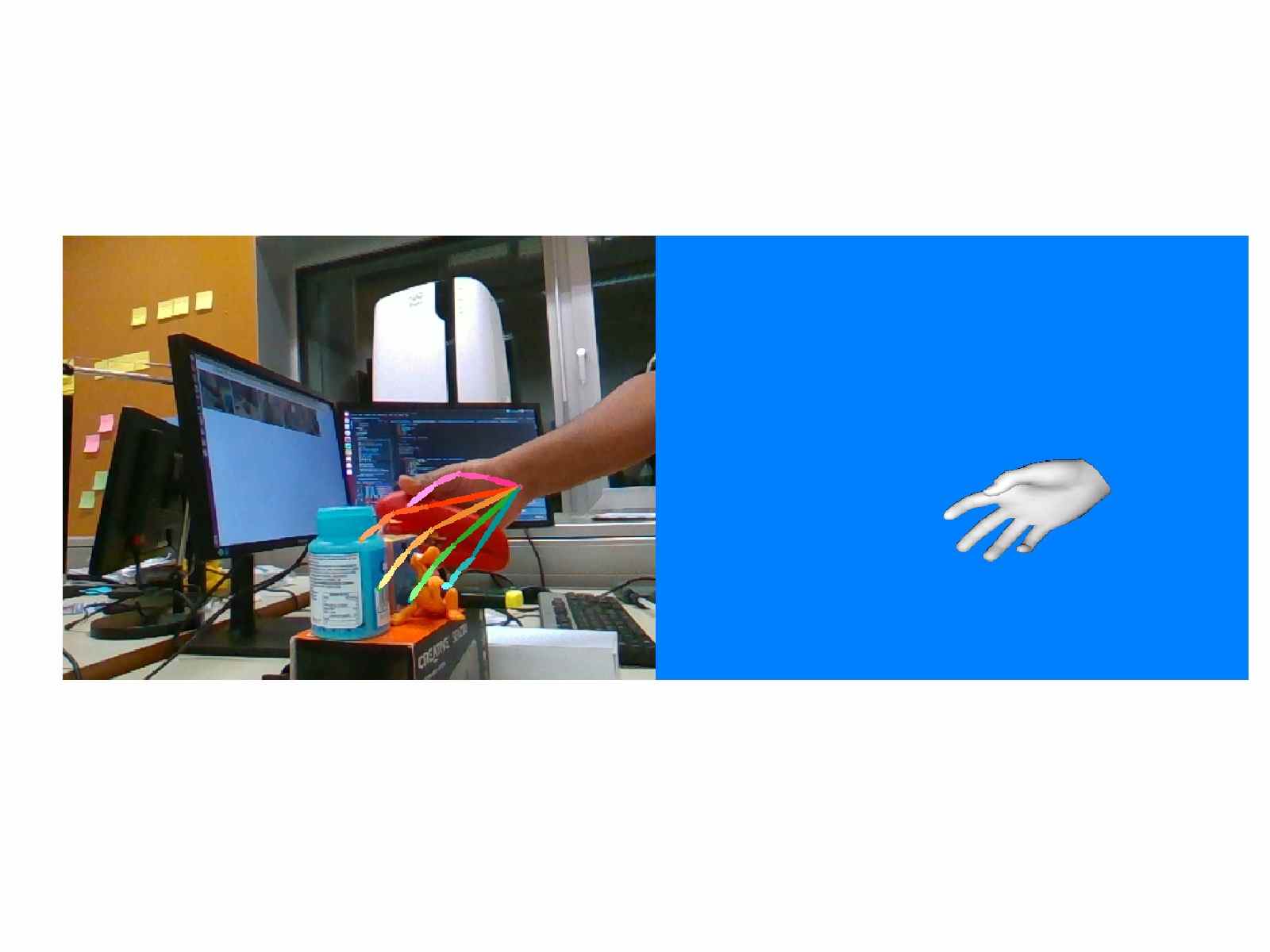} \\
			
			\includegraphics[width=0.24\linewidth,trim=0.8cm 3cm 0.8cm 3cm,clip]{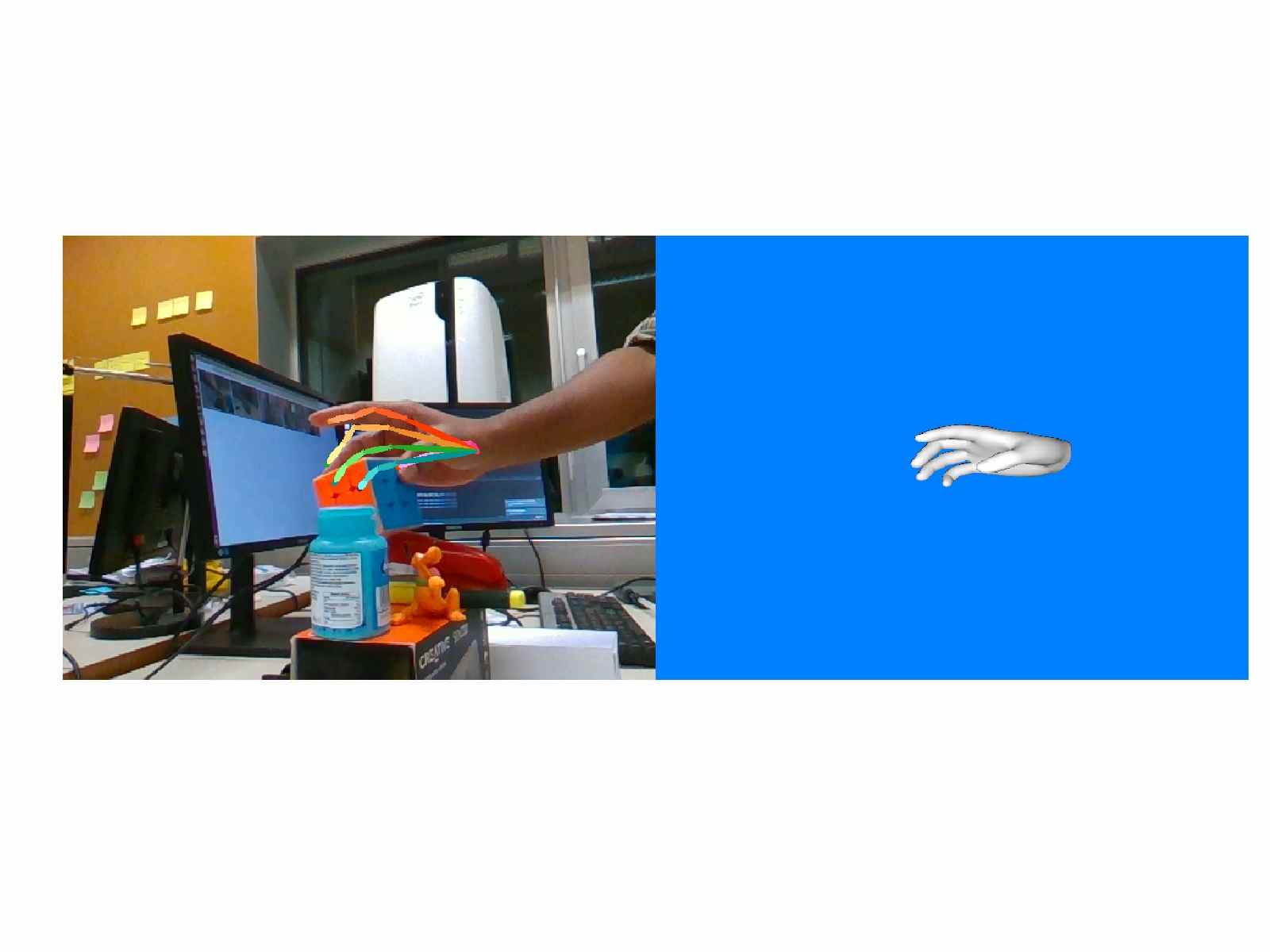} & \hspace{-3mm}\includegraphics[width=0.24\linewidth,trim=0.8cm 3cm 0.8cm 3cm,clip]{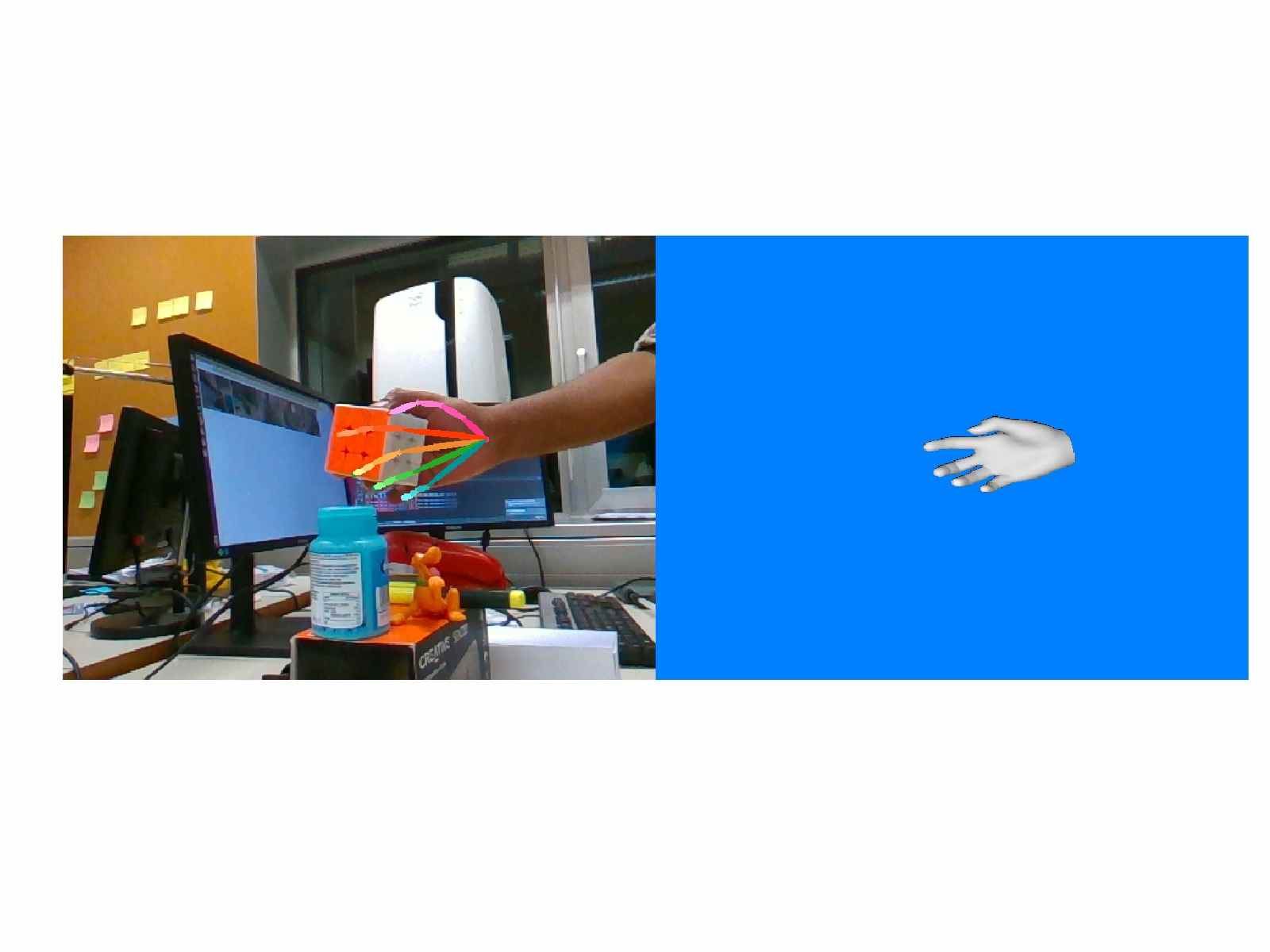} & \hspace{-3mm}\includegraphics[width=0.24\linewidth,trim=0.8cm 3cm 0.8cm 3cm,clip]{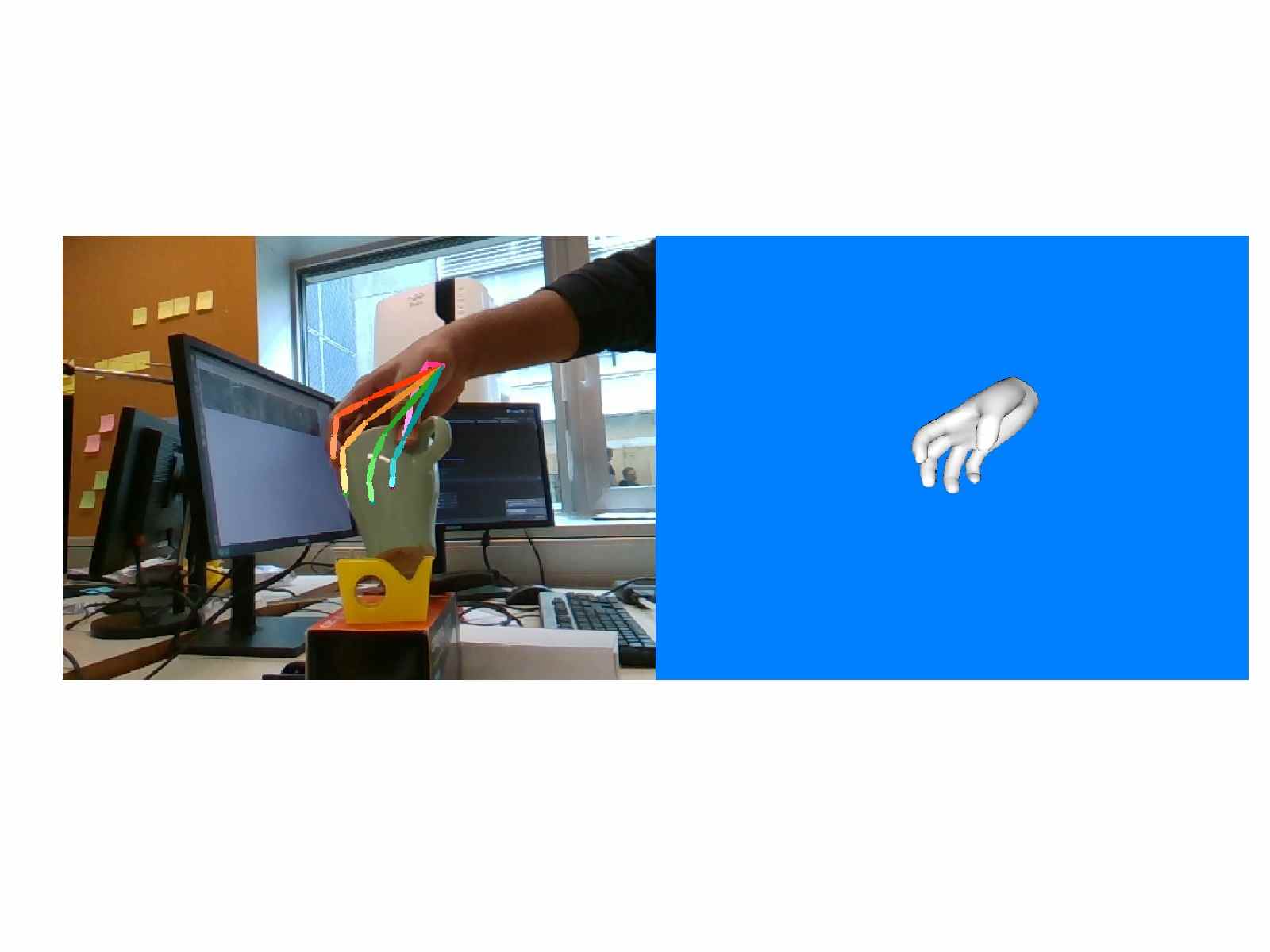} & \hspace{-3mm}\includegraphics[width=0.24\linewidth,trim=0.8cm 3cm 0.8cm 3cm,clip]{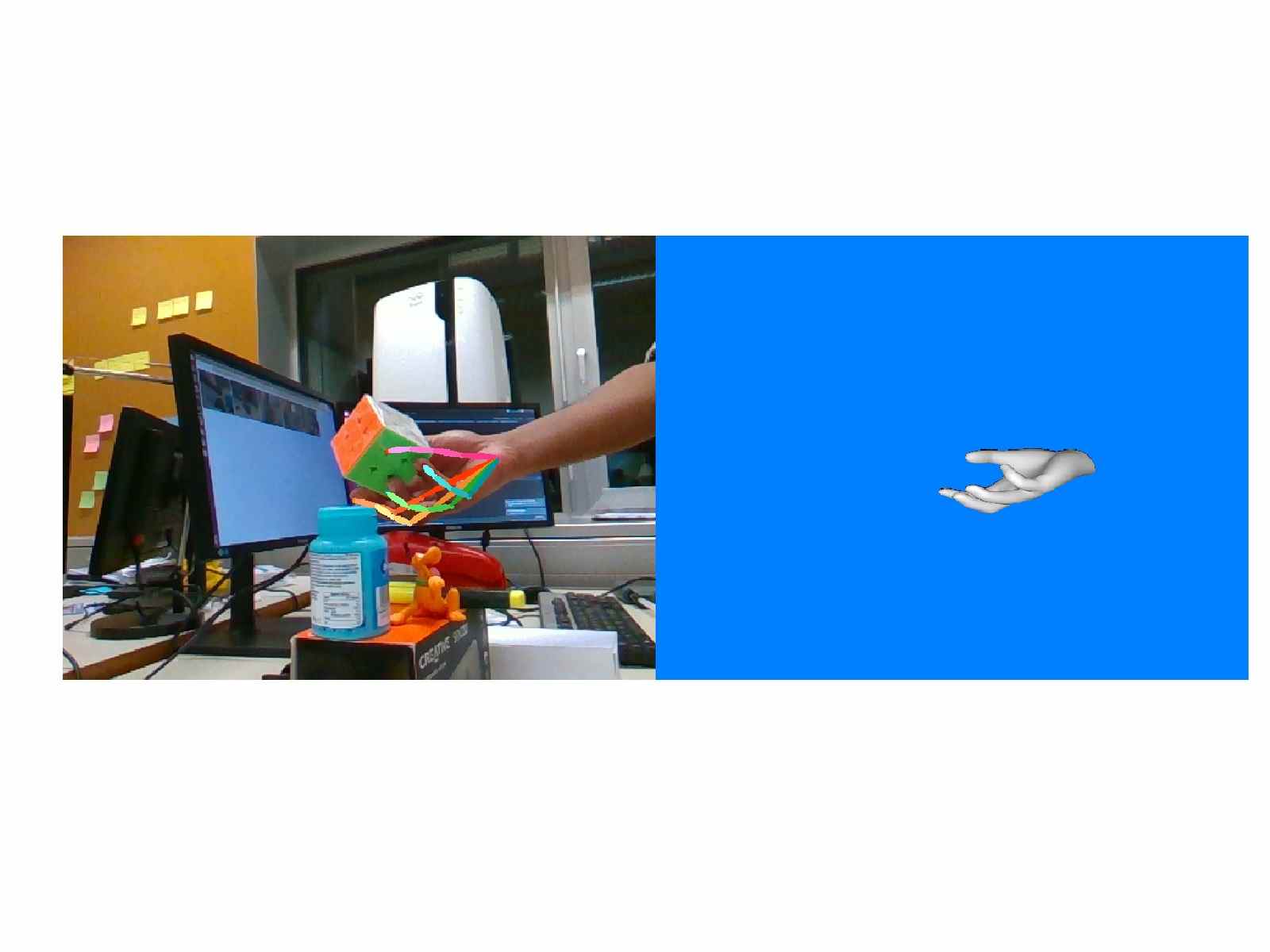} \\
			
			\includegraphics[width=0.24\linewidth,trim=0.8cm 3cm 0.8cm 3cm,clip]{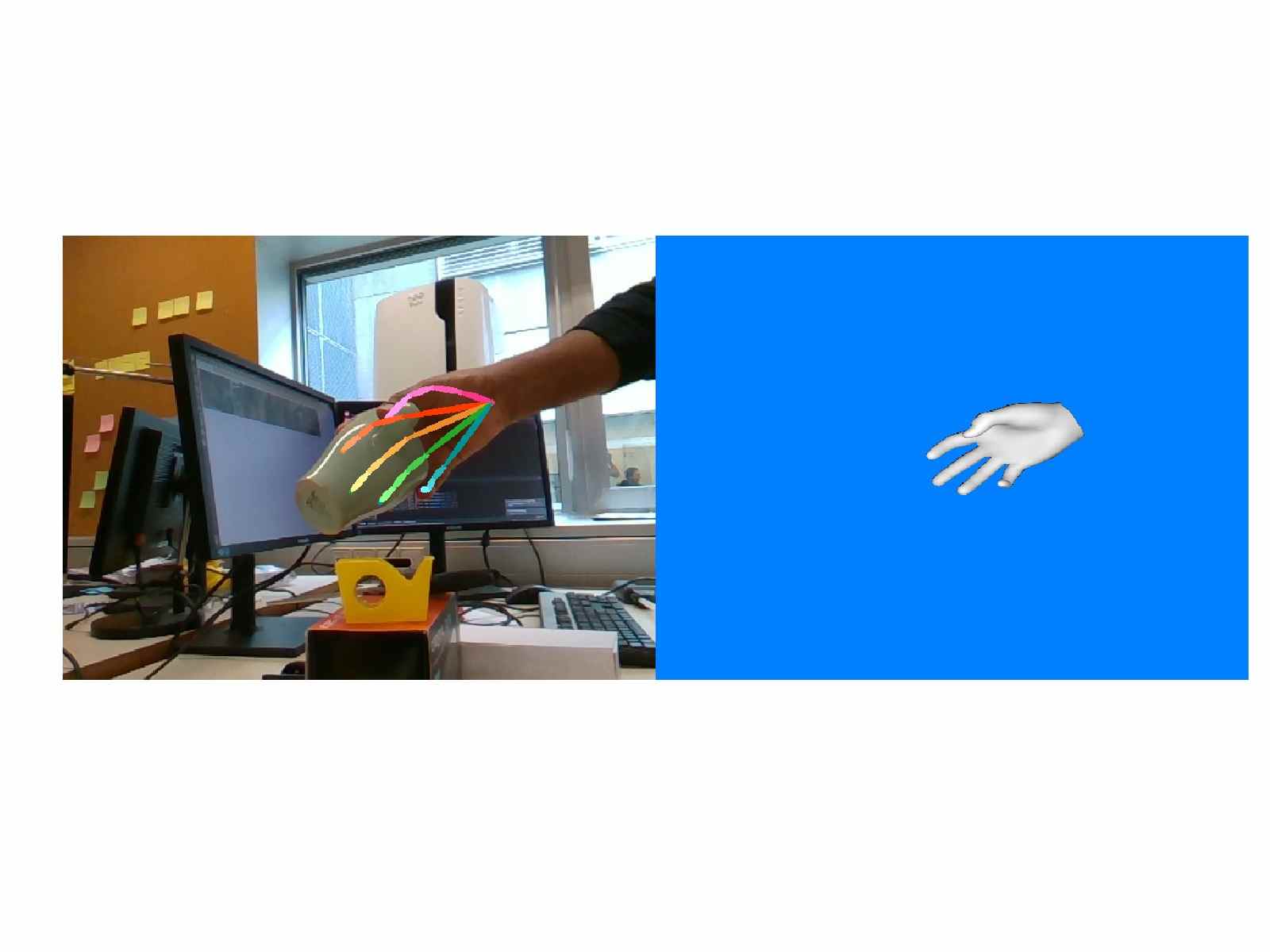} & \hspace{-3mm}\includegraphics[width=0.24\linewidth,trim=0.8cm 3cm 0.8cm 3cm,clip]{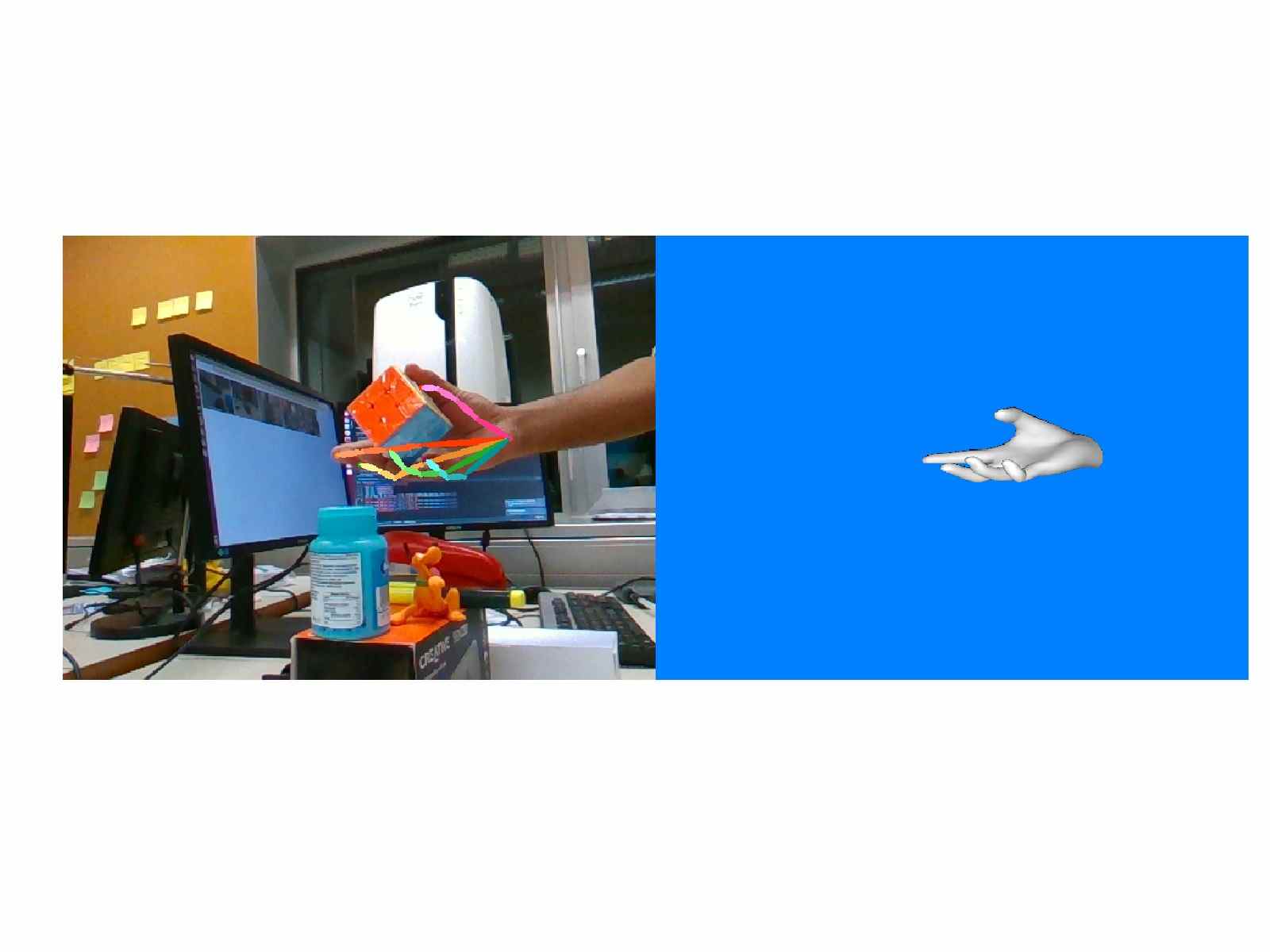} & \hspace{-3mm}\includegraphics[width=0.24\linewidth,trim=0.8cm 3cm 0.8cm 3cm,clip]{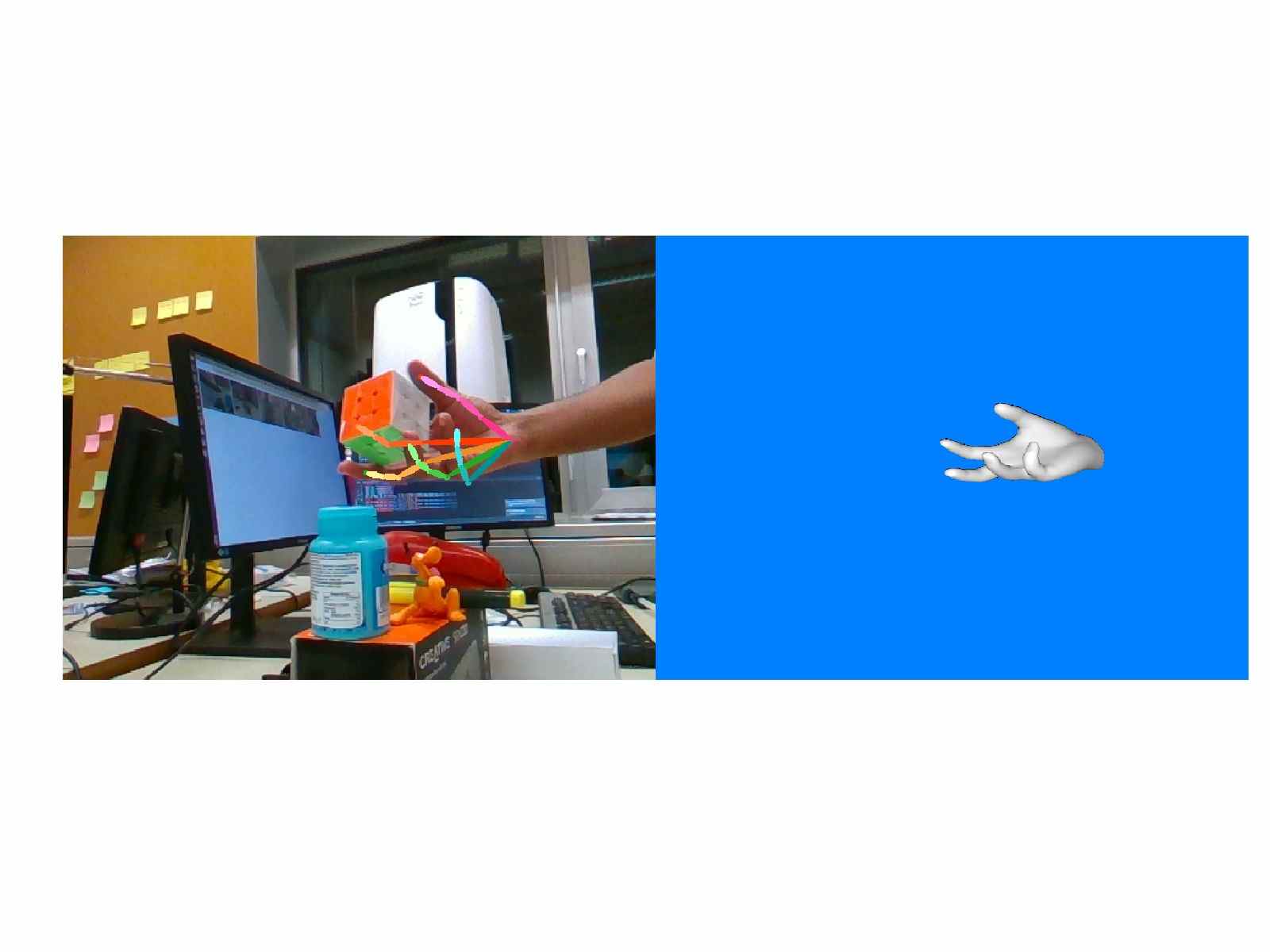} & \hspace{-3mm}\includegraphics[width=0.24\linewidth,trim=0.8cm 3cm 0.8cm 3cm,clip]{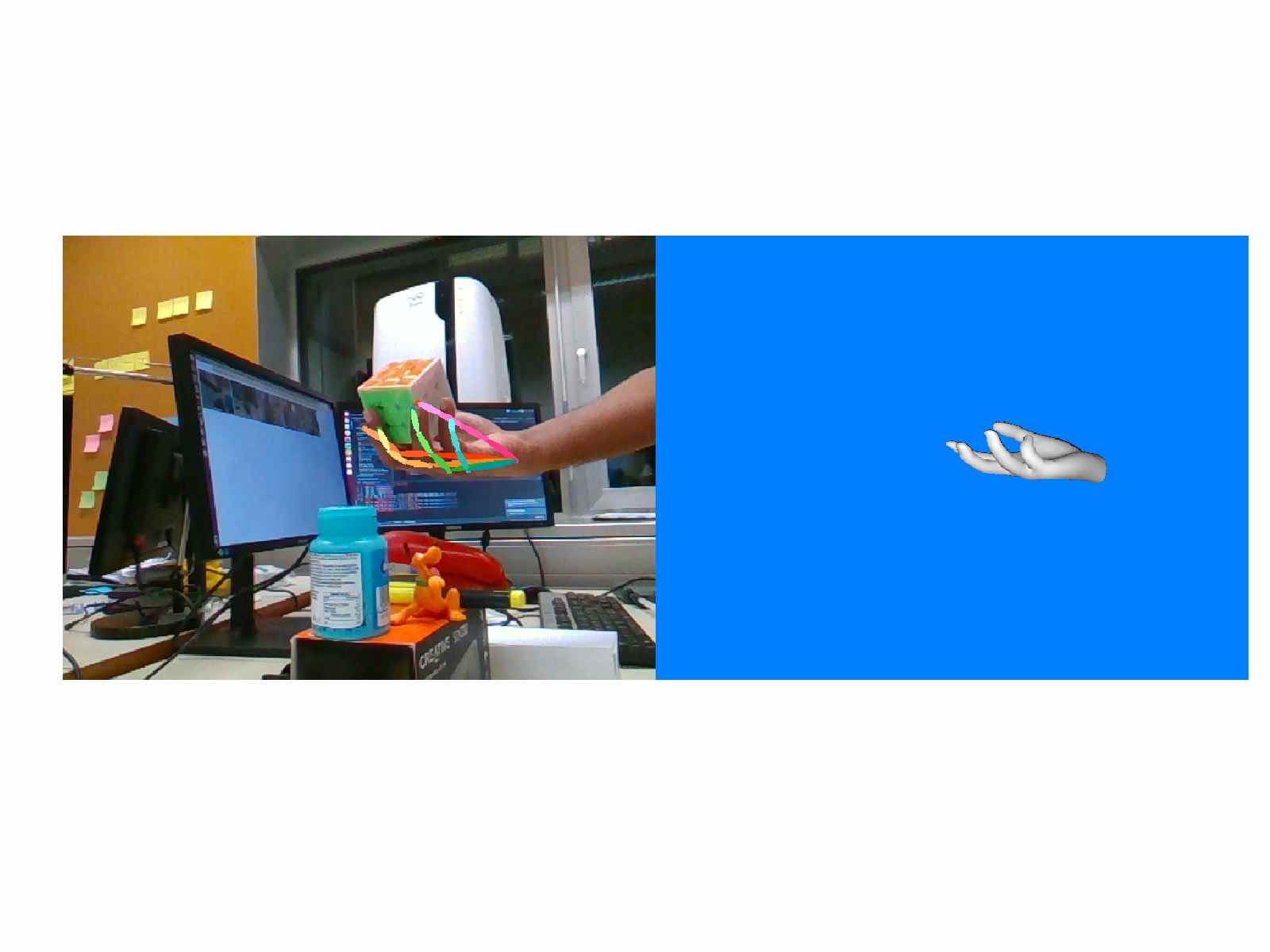} \\
			
			\includegraphics[width=0.24\linewidth,trim=0.8cm 3cm 0.8cm 3cm,clip]{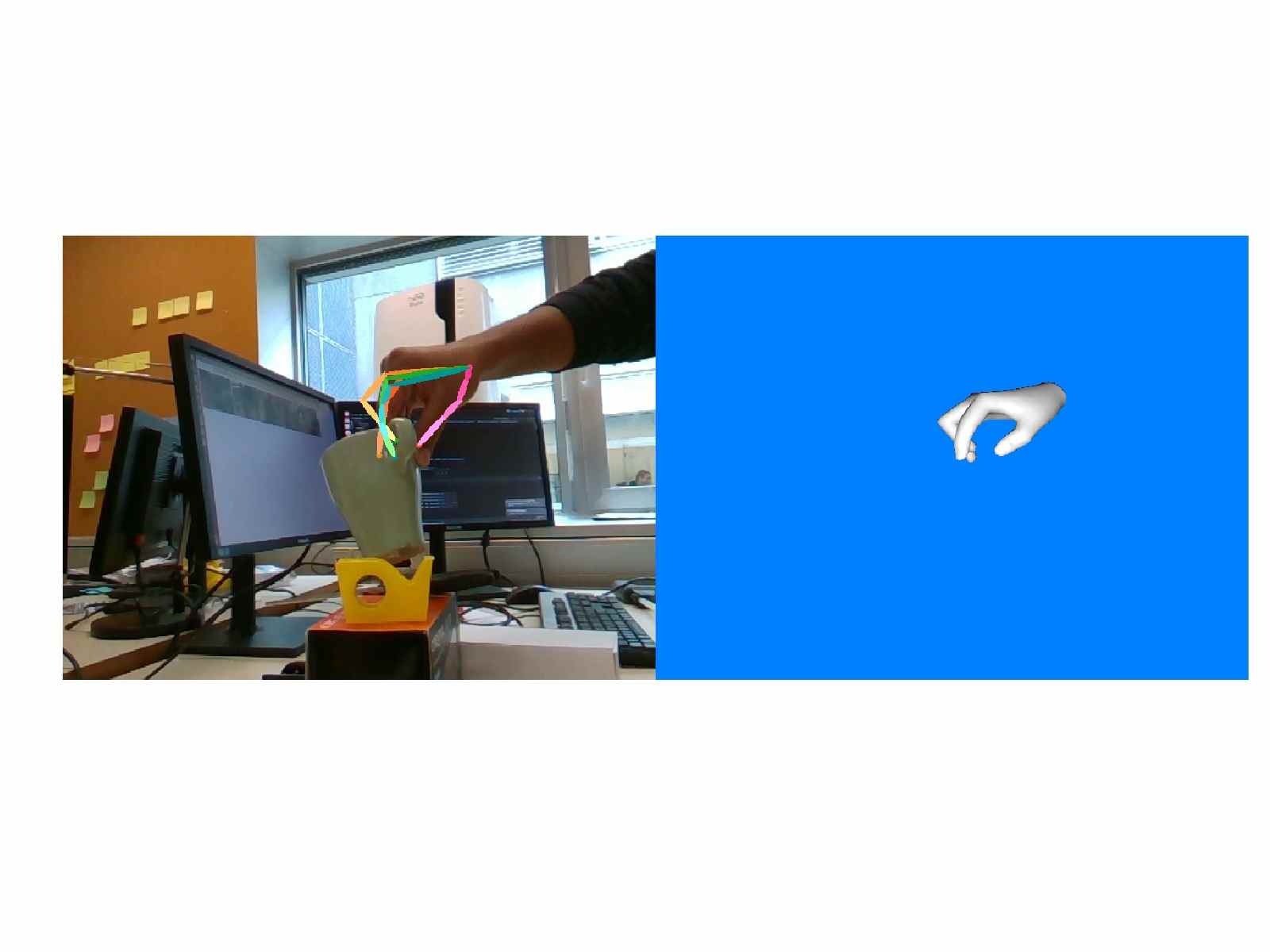} & \hspace{-3mm}\includegraphics[width=0.24\linewidth,trim=0.8cm 3cm 0.8cm 3cm,clip]{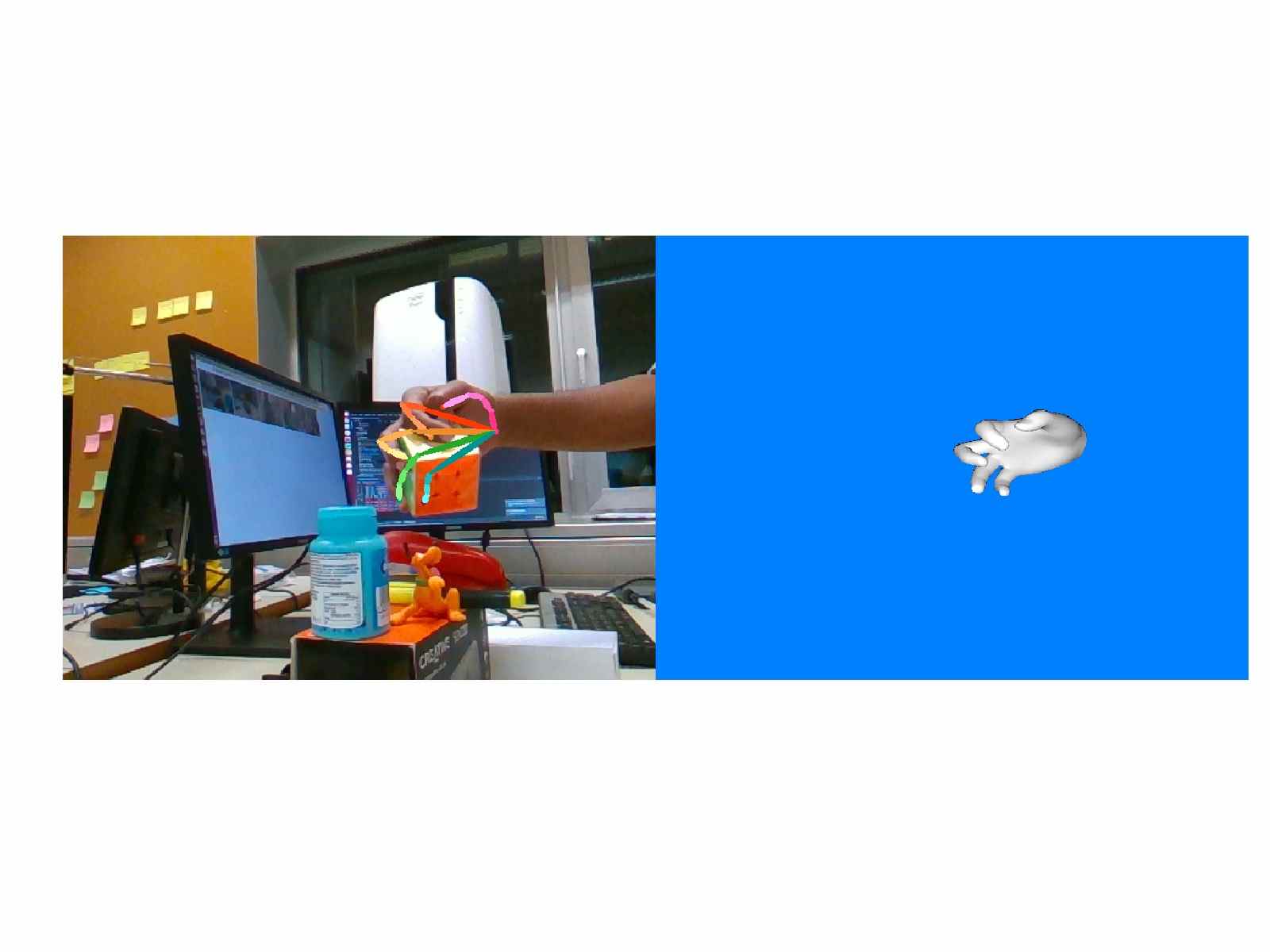} & \hspace{-3mm}\includegraphics[width=0.24\linewidth,trim=0.8cm 3cm 0.8cm 3cm,clip]{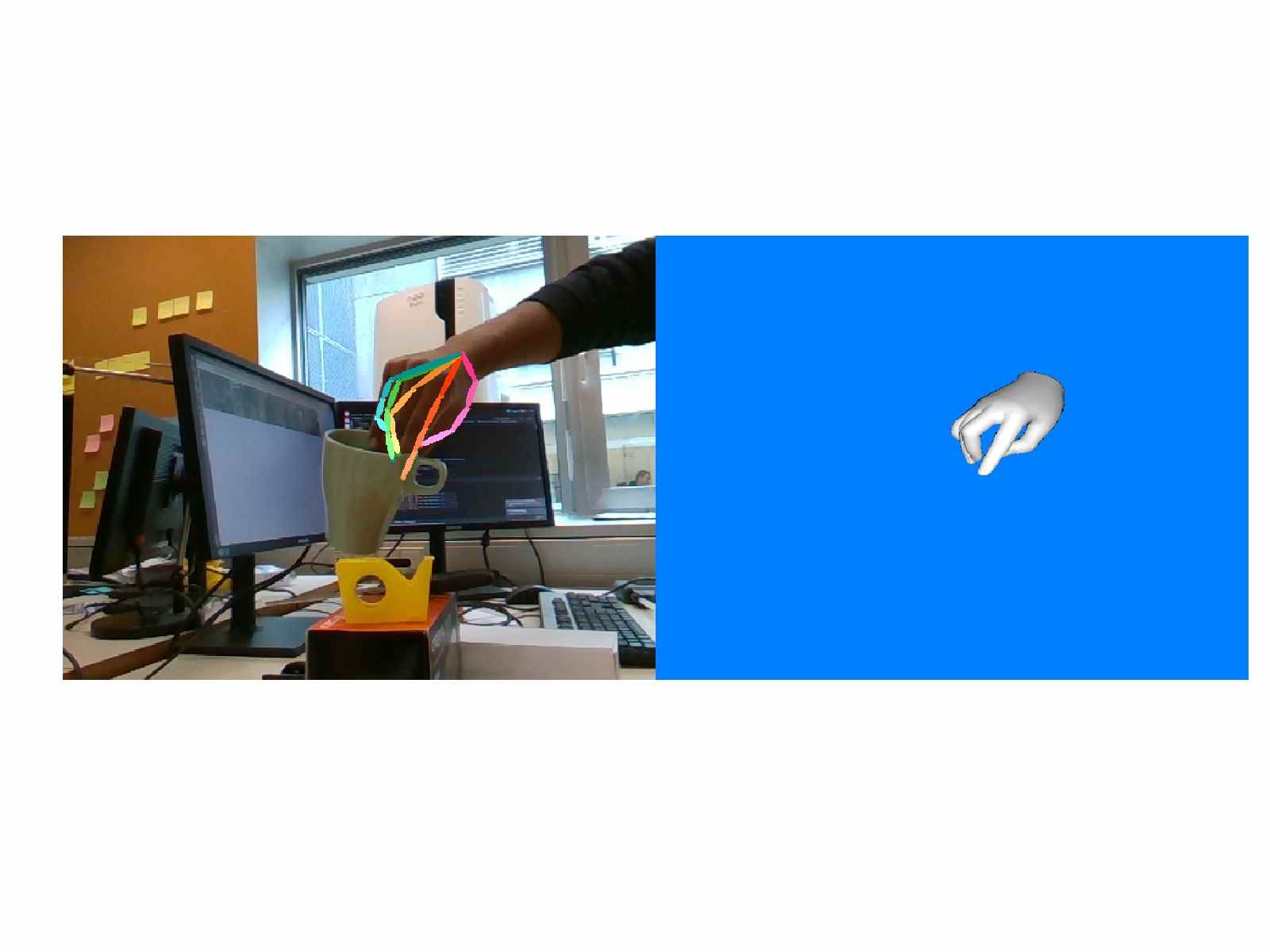} & \hspace{-3mm}\includegraphics[width=0.24\linewidth,trim=0.8cm 3cm 0.8cm 3cm,clip]{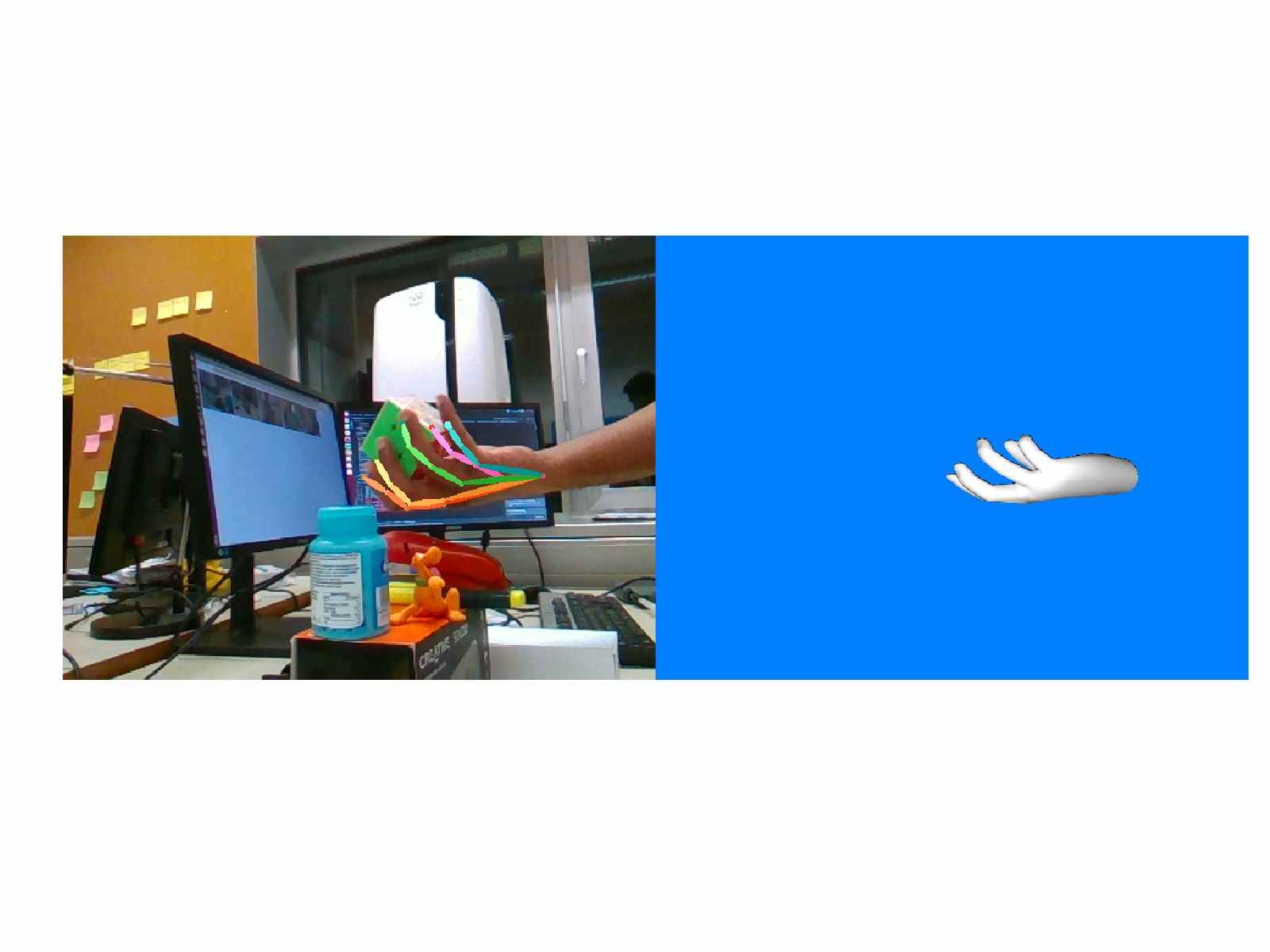} \\
			
			\includegraphics[width=0.24\linewidth,trim=0.8cm 3cm 0.8cm 3cm,clip]{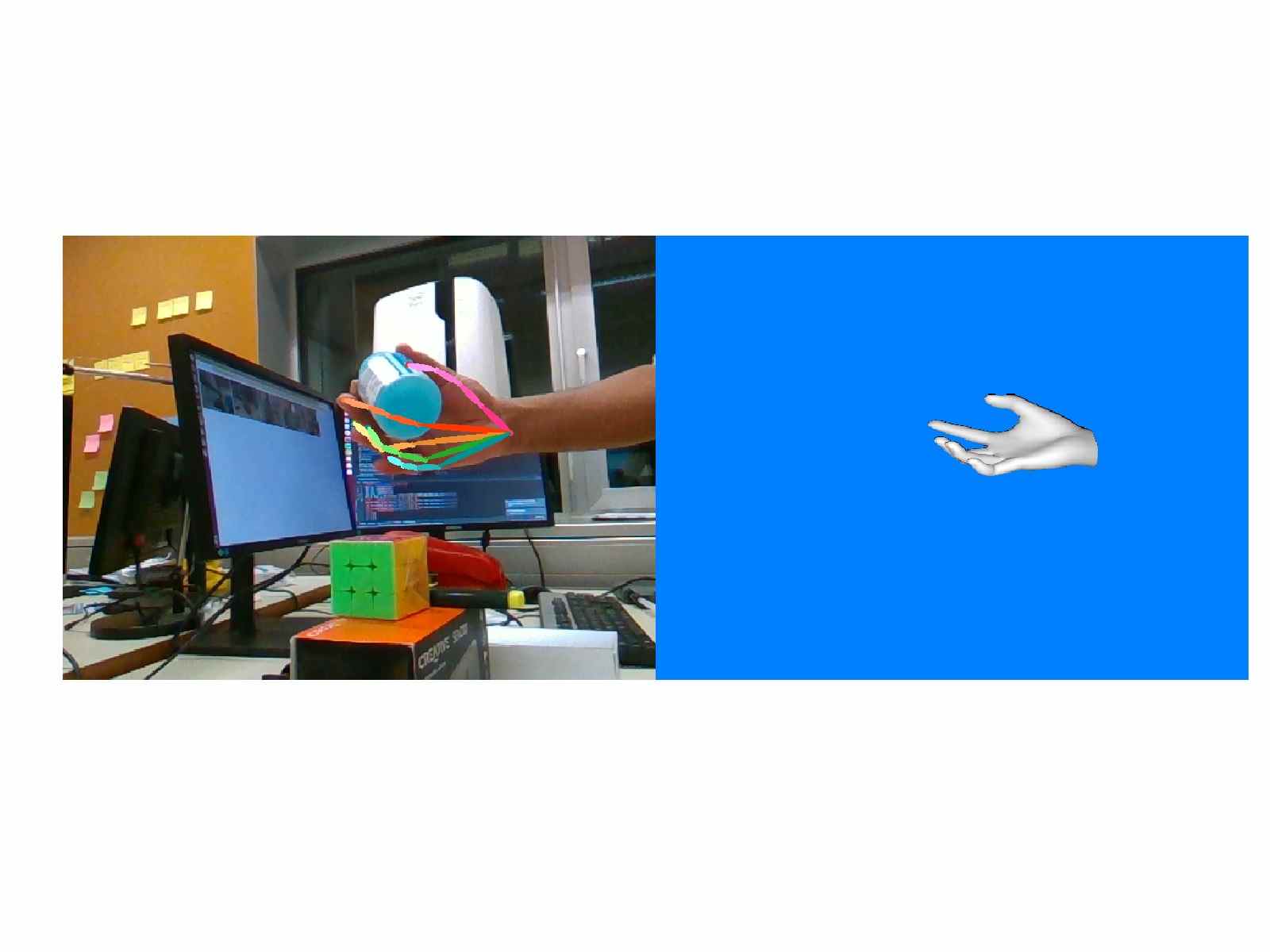} & \hspace{-3mm}\includegraphics[width=0.24\linewidth,trim=0.8cm 3cm 0.8cm 3cm,clip]{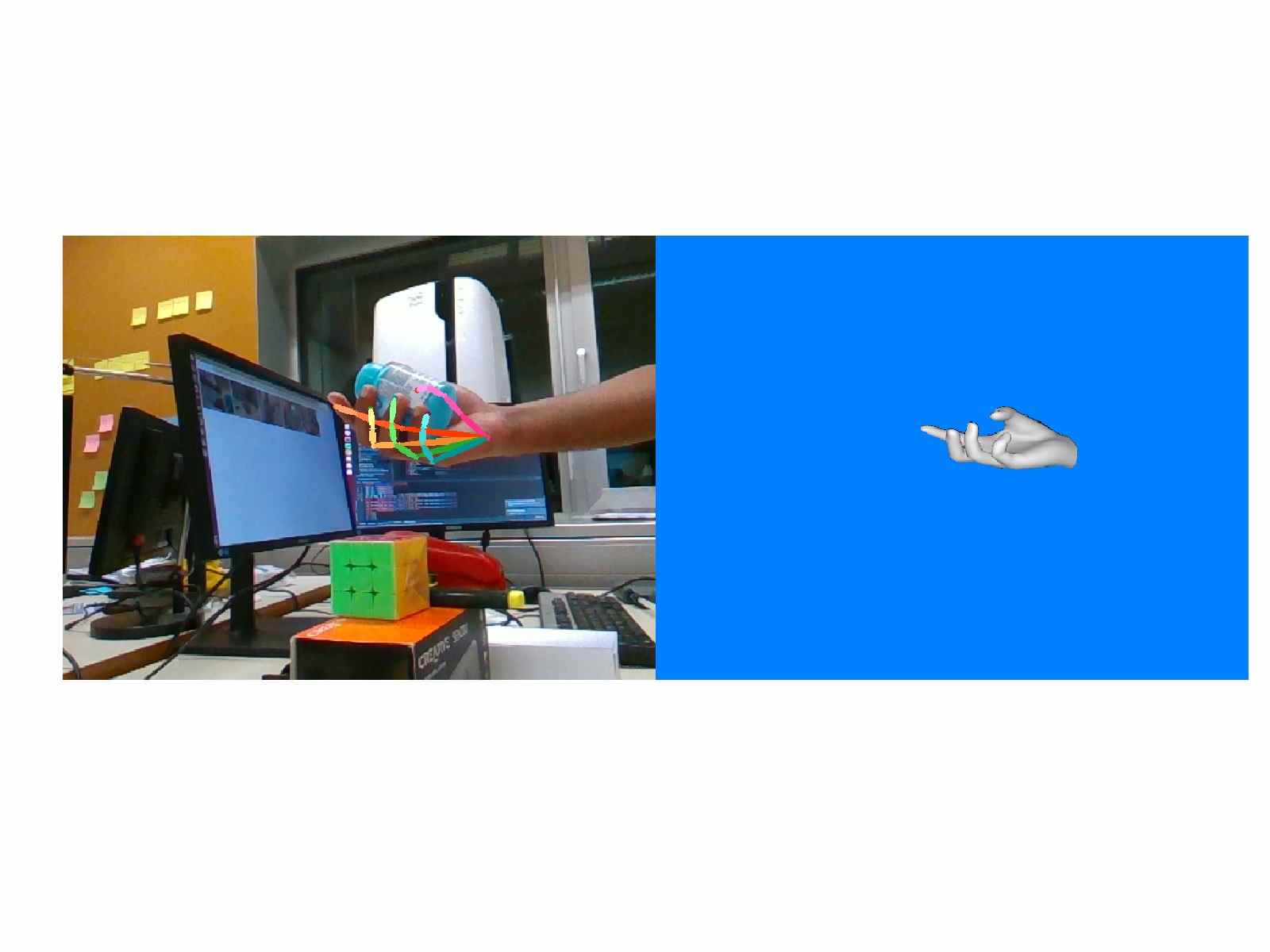} & \hspace{-3mm}\includegraphics[width=0.24\linewidth,trim=0.8cm 3cm 0.8cm 3cm,clip]{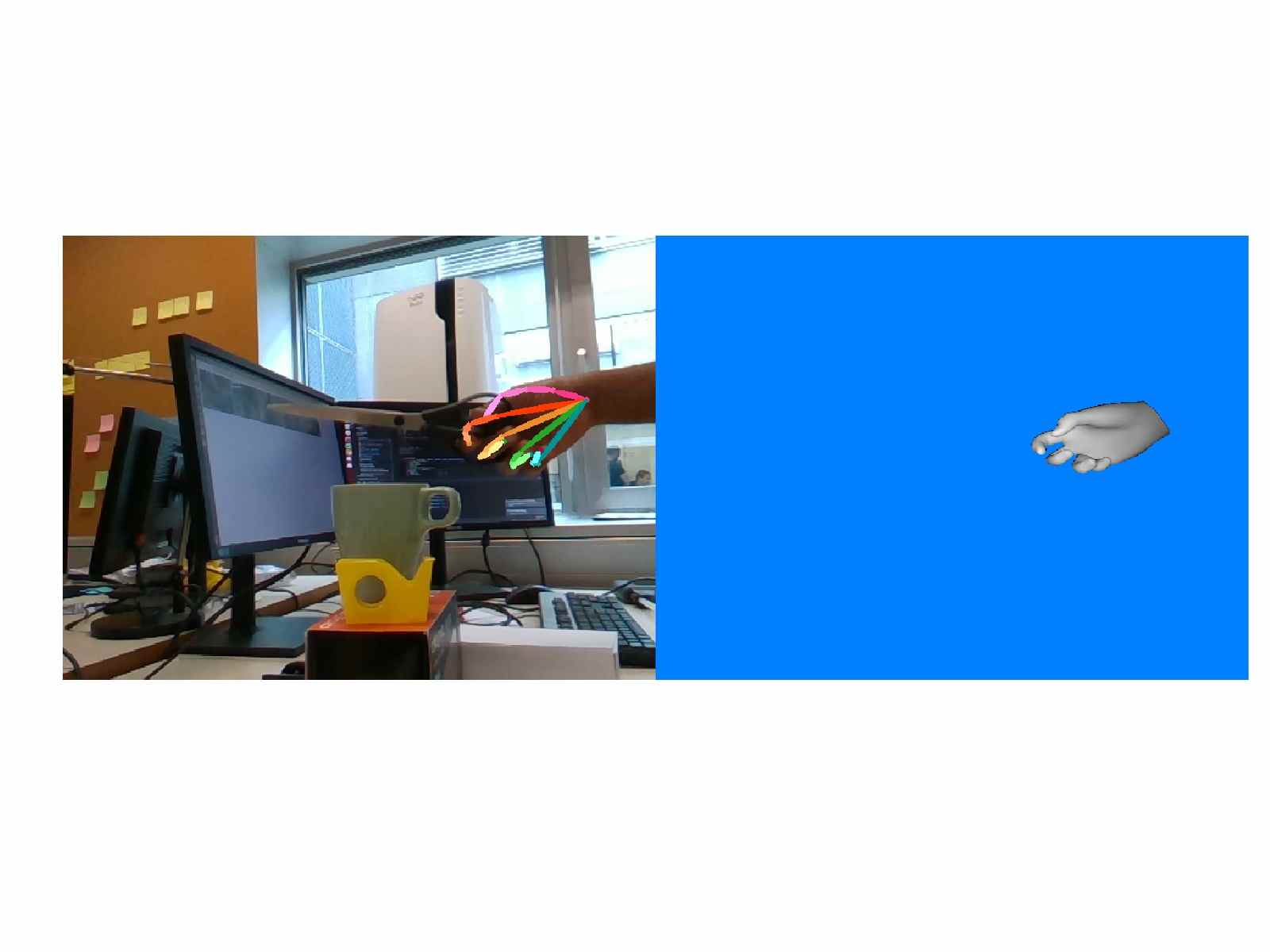} & \hspace{-3mm}\includegraphics[width=0.24\linewidth,trim=0.8cm 3cm 0.8cm 3cm,clip]{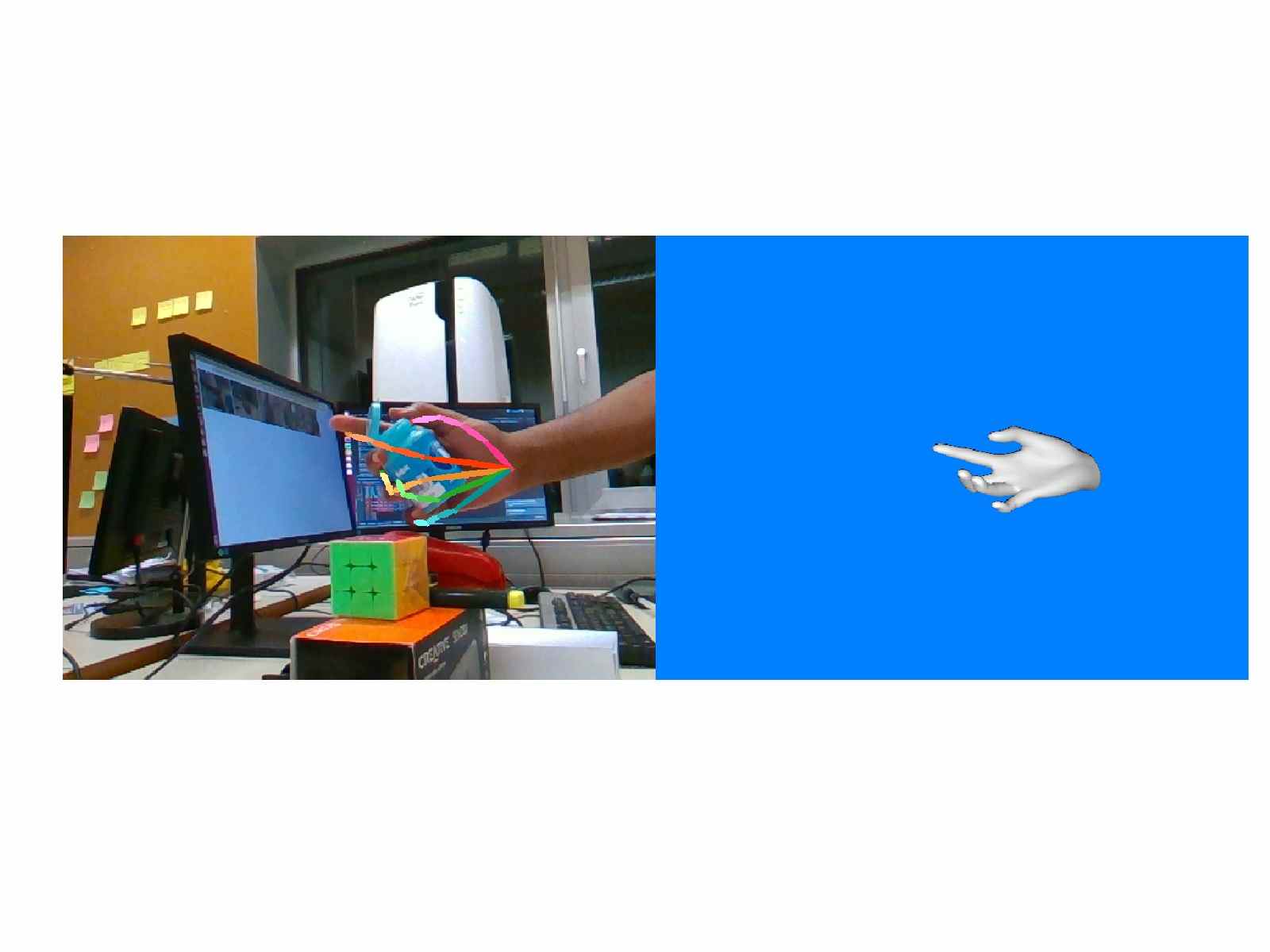} \\
			
			\includegraphics[width=0.24\linewidth,trim=0.8cm 3cm 0.8cm 3cm,clip]{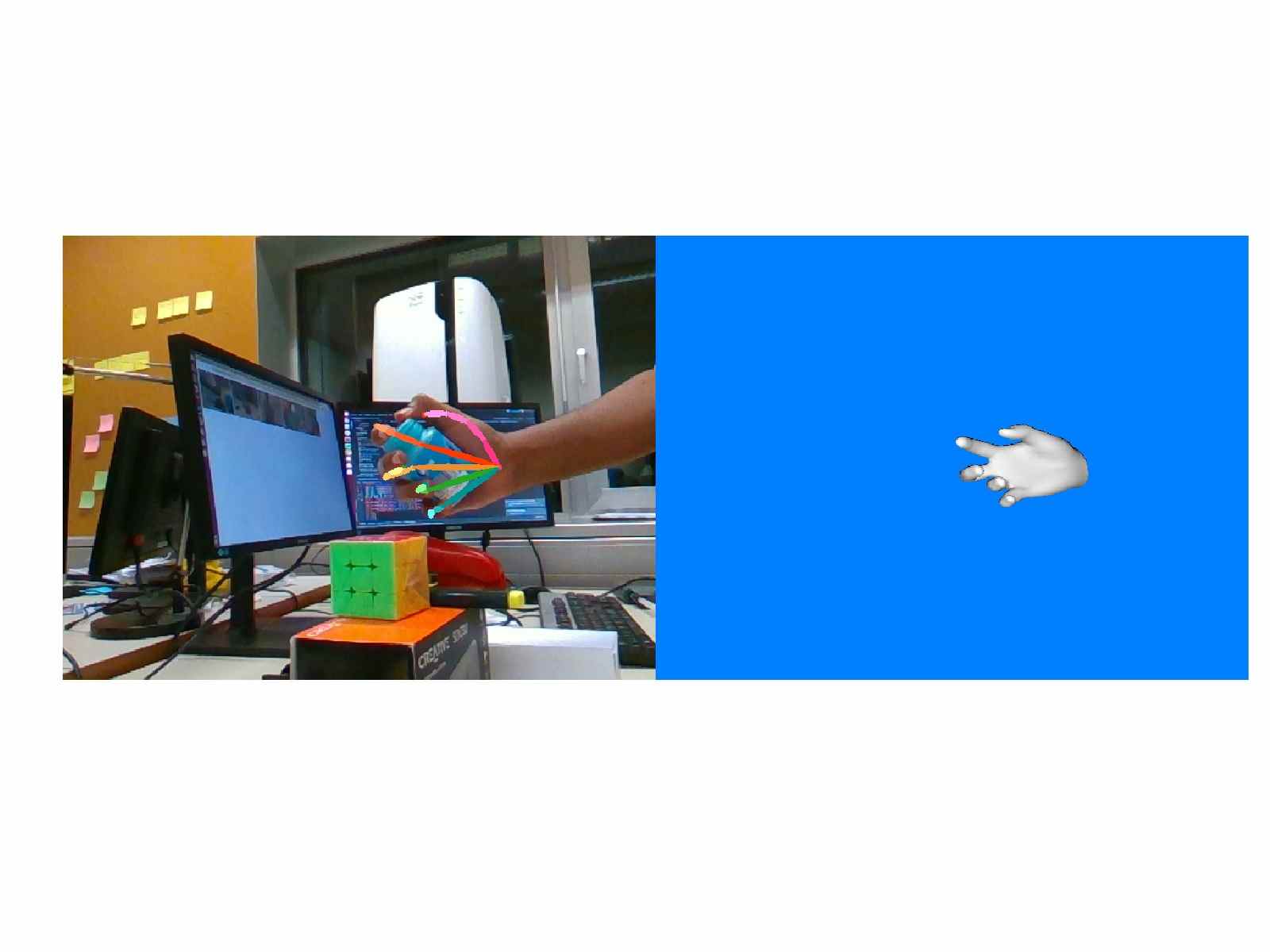} & \hspace{-3mm}\includegraphics[width=0.24\linewidth,trim=0.8cm 3cm 0.8cm 3cm,clip]{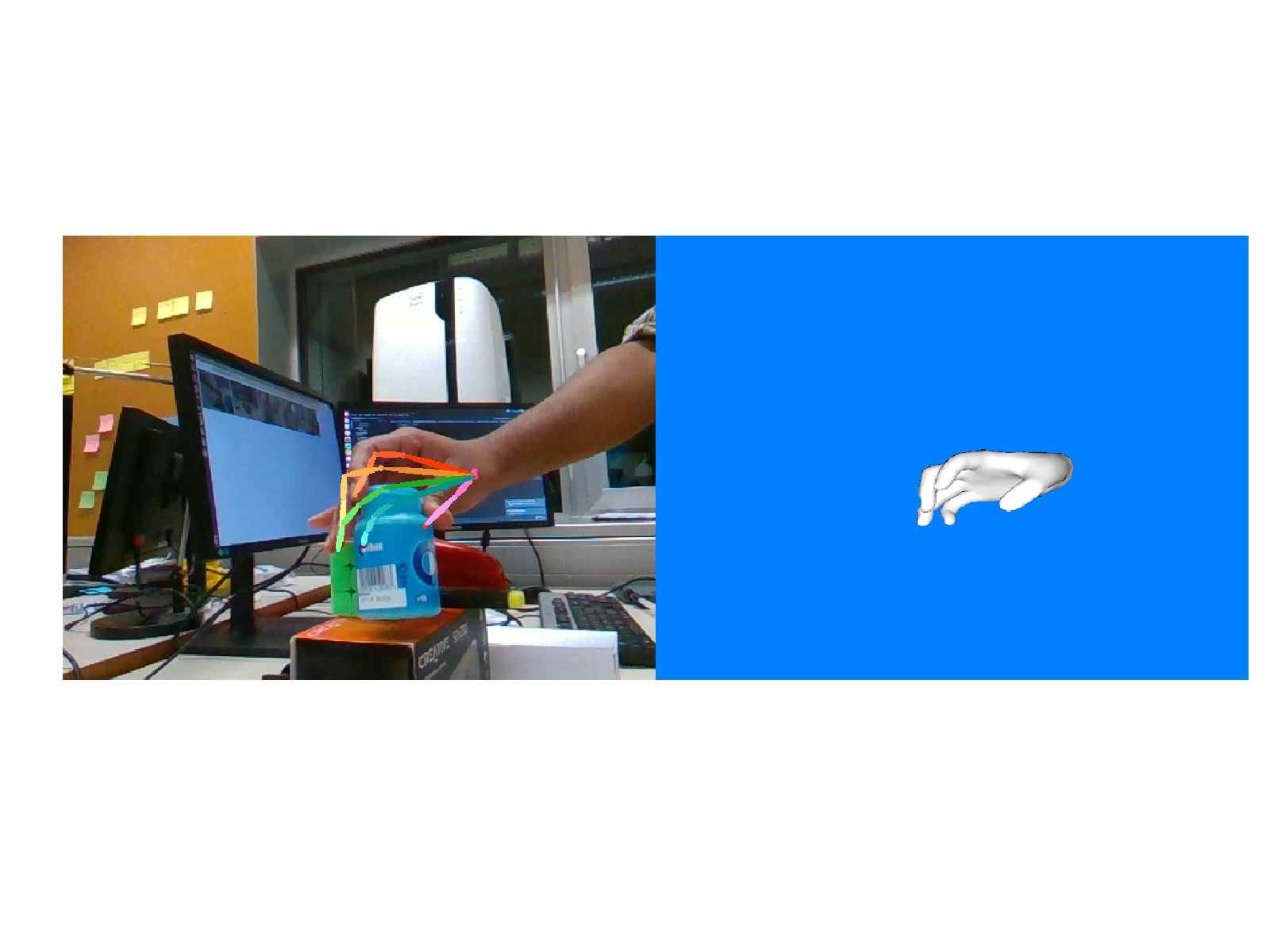} & \hspace{-3mm}\includegraphics[width=0.24\linewidth,trim=0.8cm 3cm 0.8cm 3cm,clip]{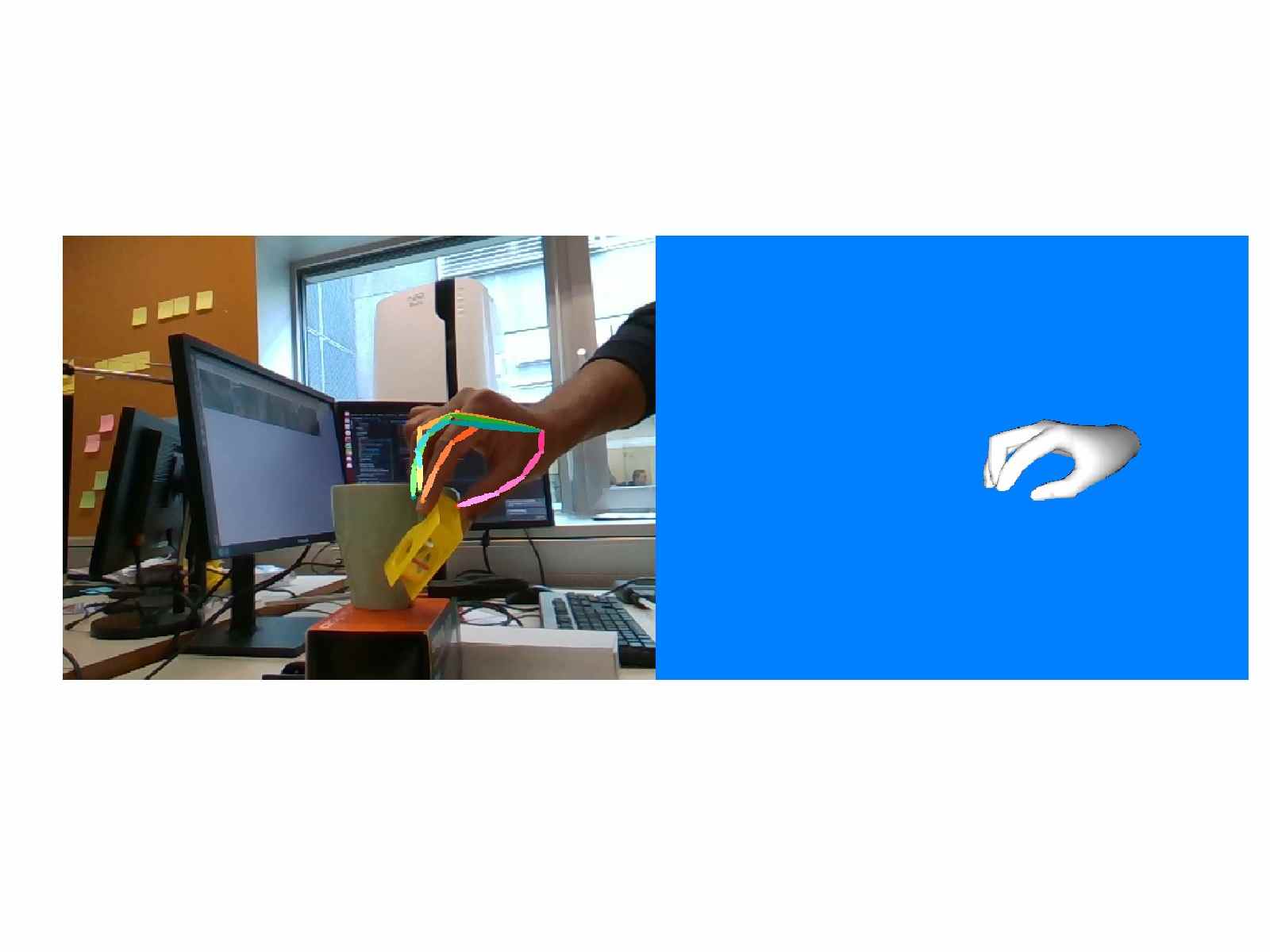} & \hspace{-3mm}\includegraphics[width=0.24\linewidth,trim=0.8cm 3cm 0.8cm 3cm,clip]{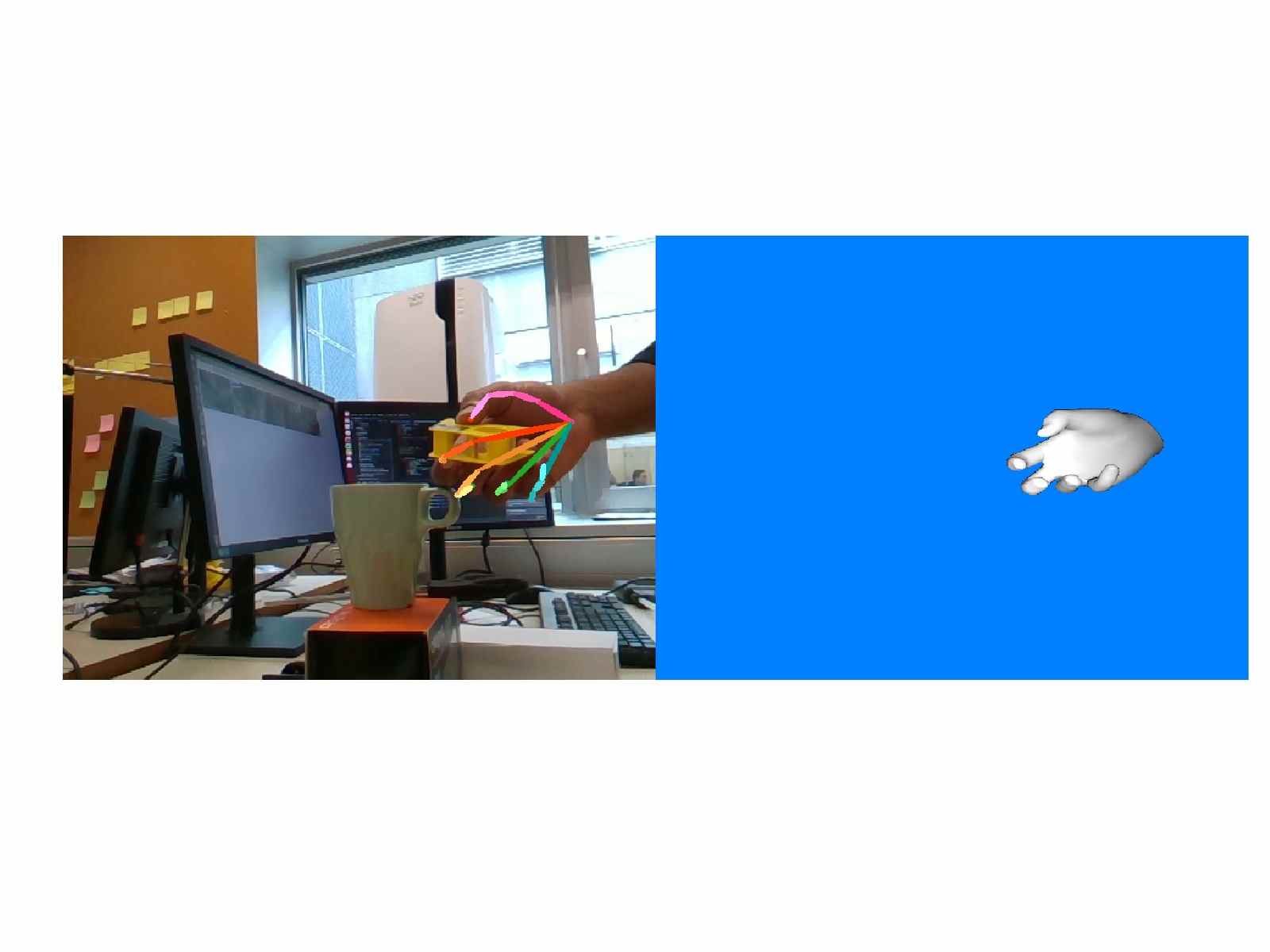} \\
			
			\includegraphics[width=0.24\linewidth,trim=0.8cm 3cm 0.8cm 3cm,clip]{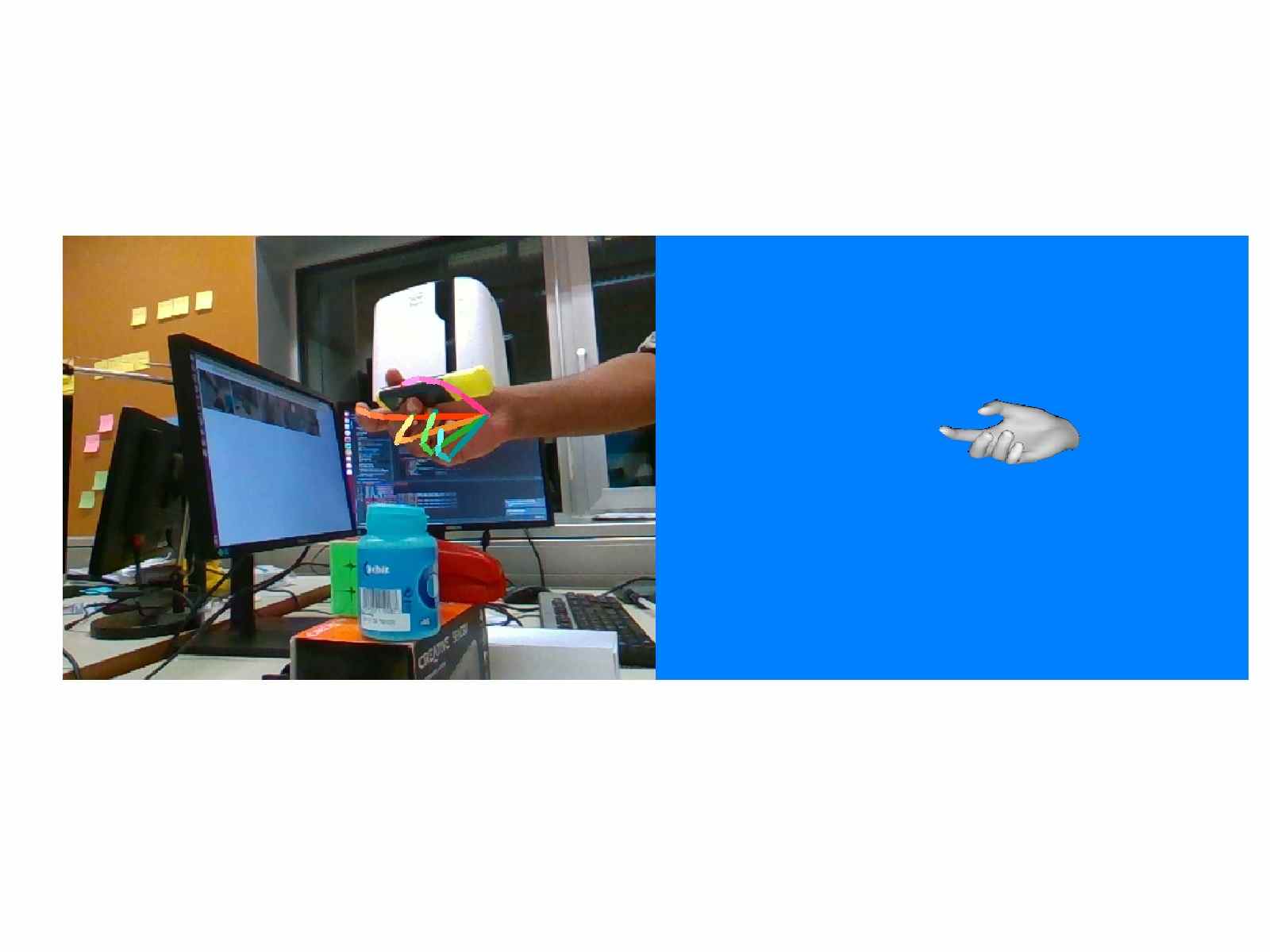} & \hspace{-3mm}\includegraphics[width=0.24\linewidth,trim=0.8cm 3cm 0.8cm 3cm,clip]{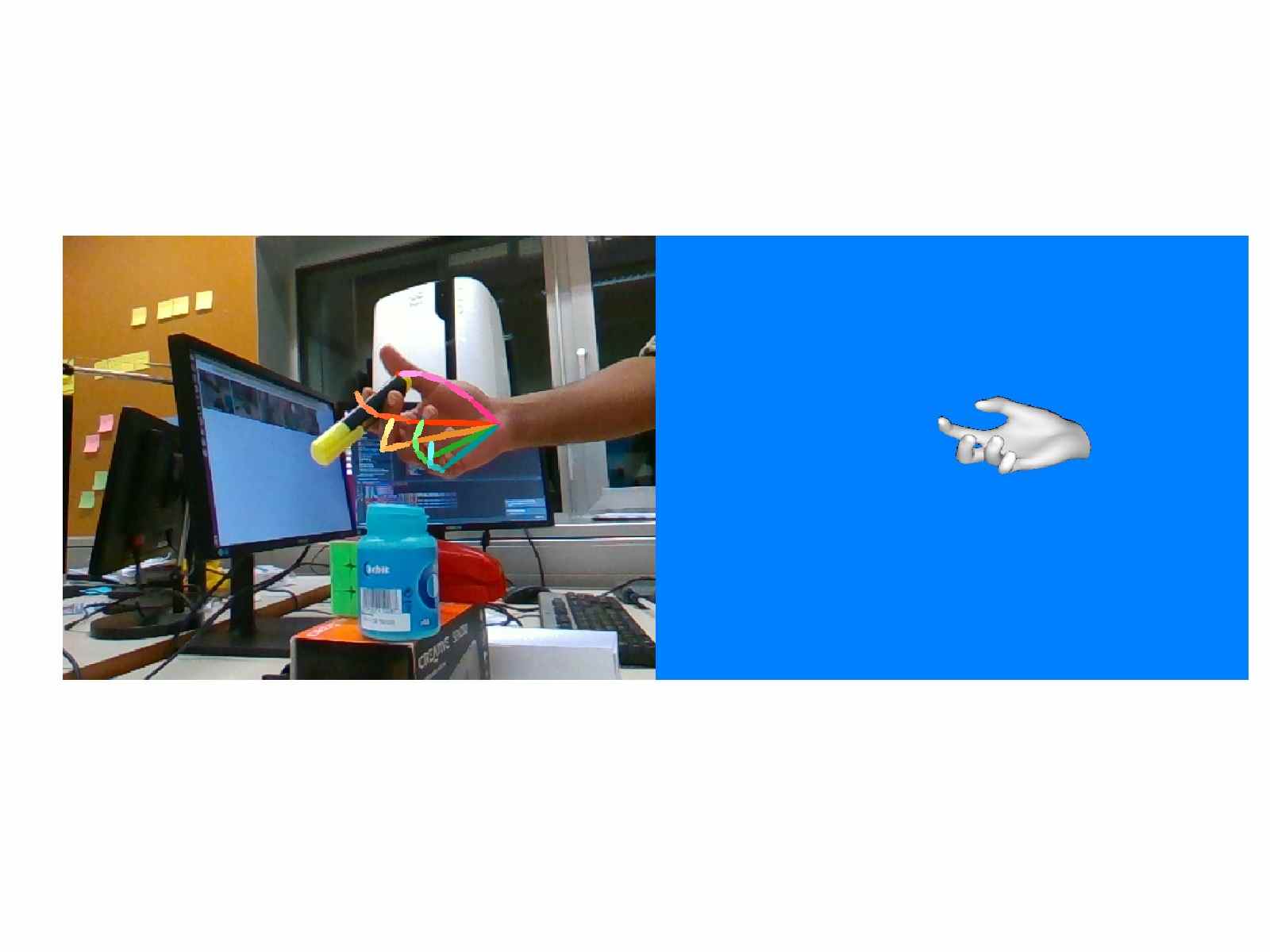} & \hspace{-3mm}\includegraphics[width=0.24\linewidth,trim=0.8cm 3cm 0.8cm 3cm,clip]{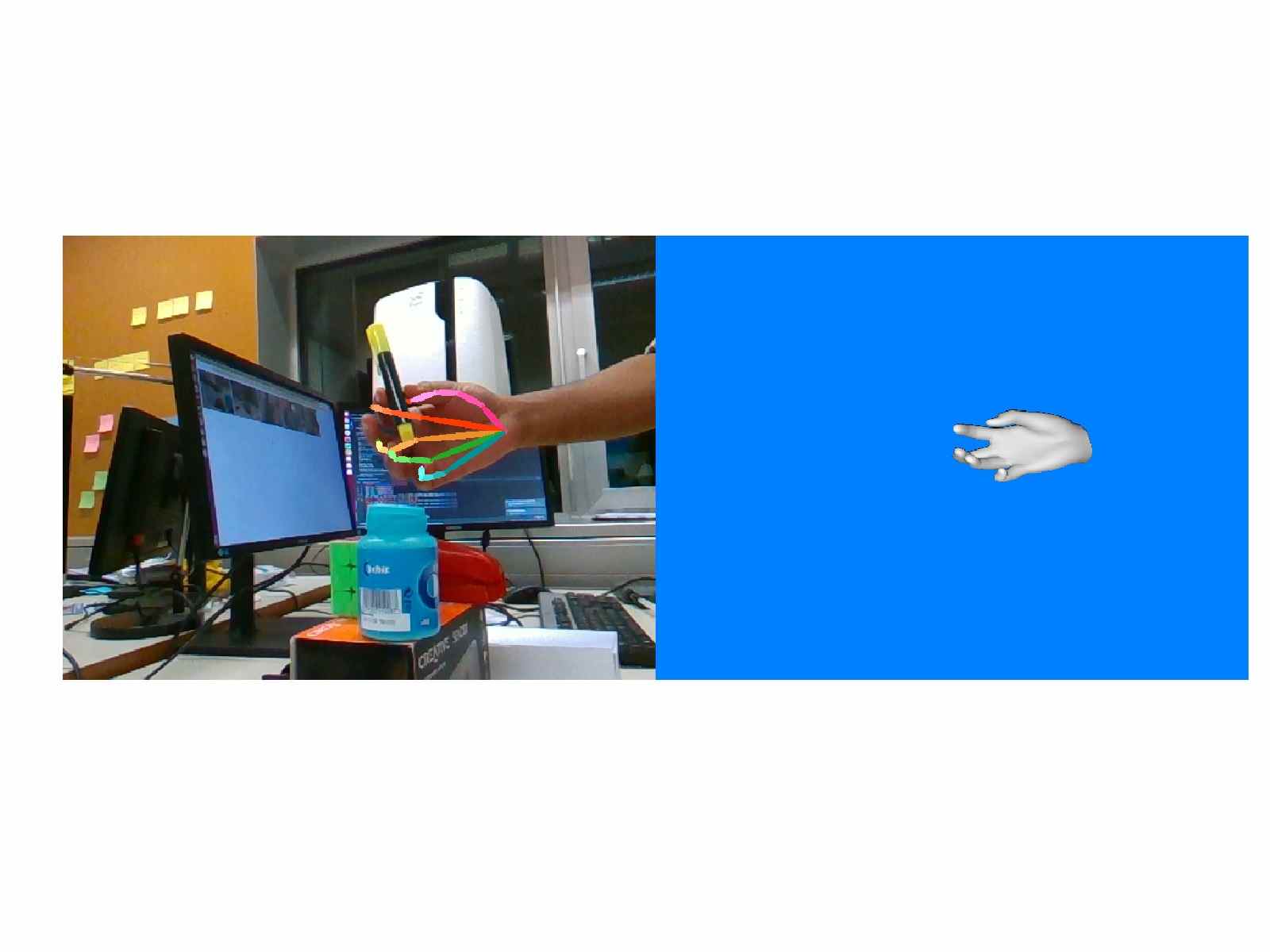} & \hspace{-3mm}\includegraphics[width=0.24\linewidth,trim=0.8cm 3cm 0.8cm 3cm,clip]{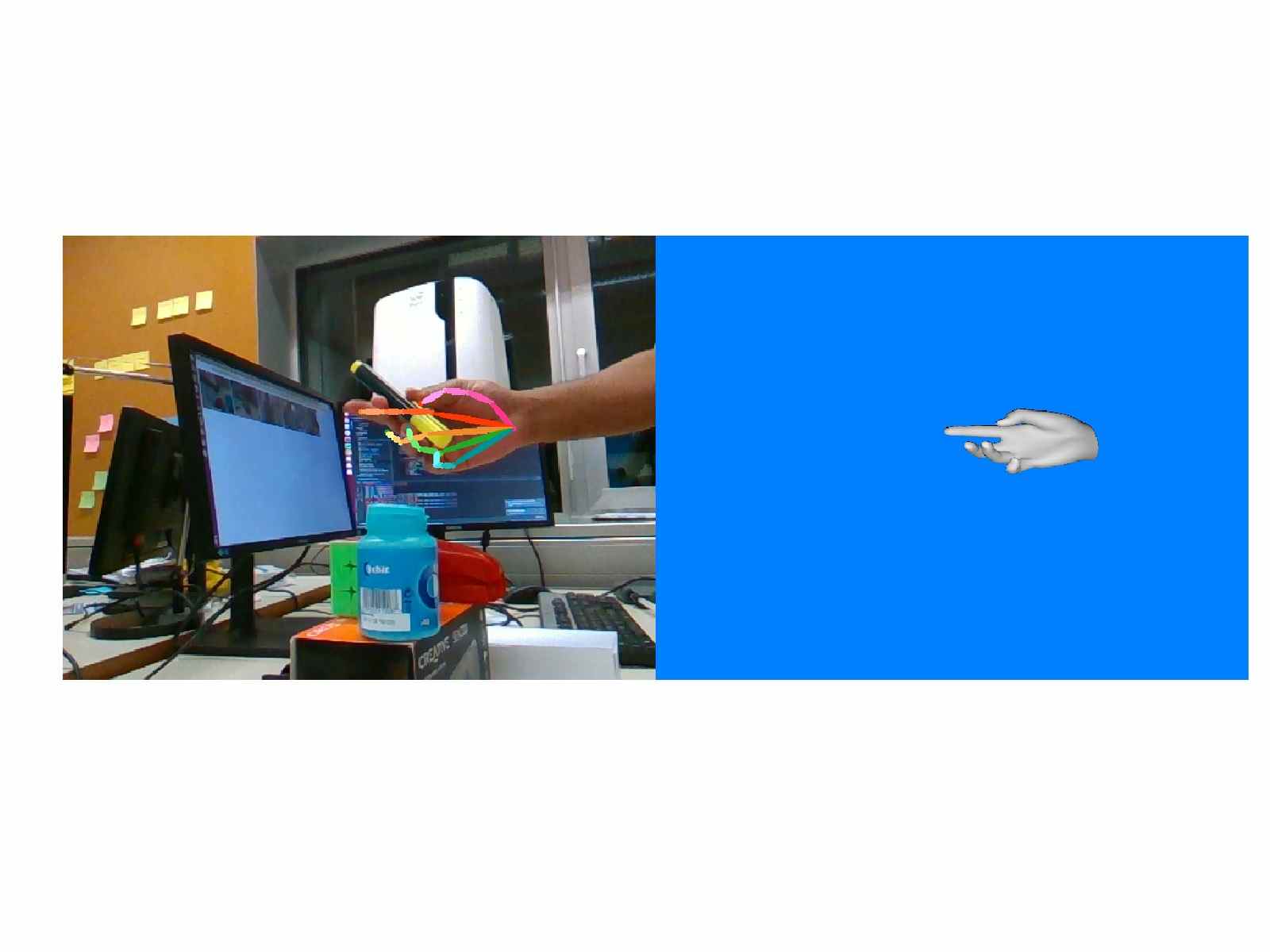} \\

		\end{tabular}
	\end{center}
	\vspace{-2mm}
	\caption{Qualitative results of 3D hand pose estimation of hand manipulating unseen objects. Our pose estimator trained on the \datasetname dataset is still able to predict accurate 3D poses when interacting with new objects.}
	\label{fig:unseen_objects}
\end{figure*}

\begin{figure*}
	\begin{center}
	\begin{tabular}{ccc}
		\includegraphics[width=0.33\linewidth]{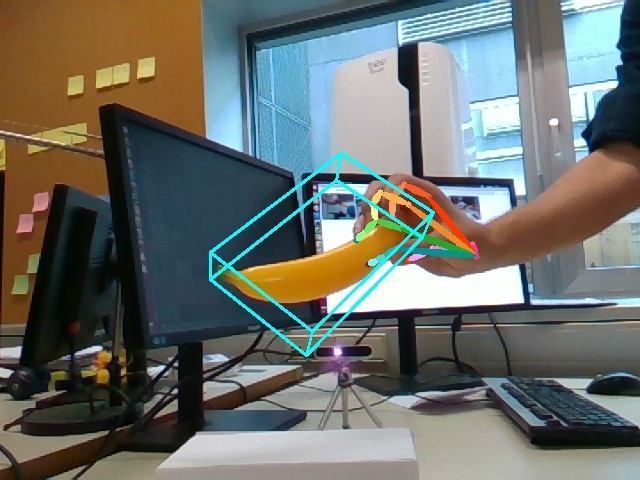} & \hspace{-3mm}
		\includegraphics[width=0.33\linewidth]{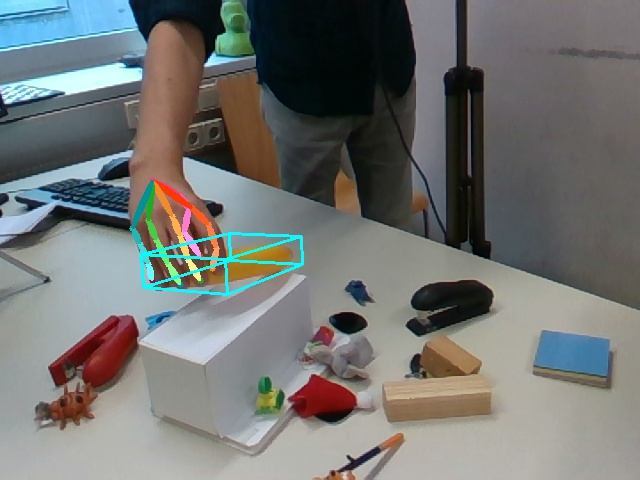} & \hspace{-3mm}
		\includegraphics[width=0.33\linewidth]{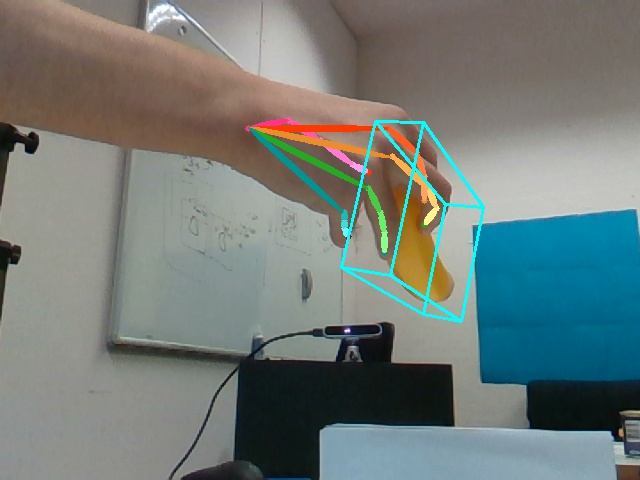} \\

		\includegraphics[width=0.33\linewidth]{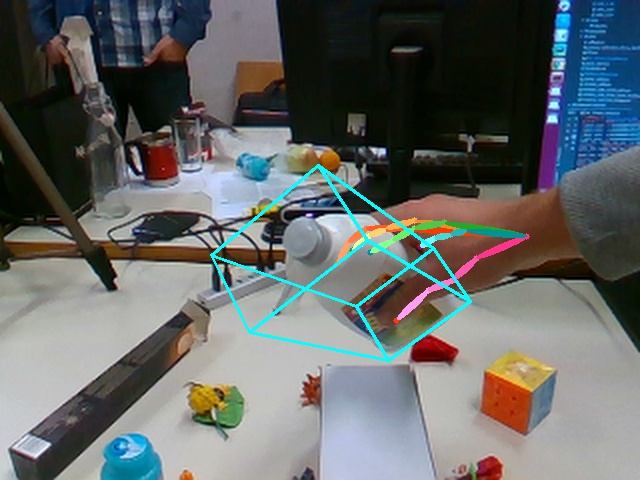} & \hspace{-3mm}
		\includegraphics[width=0.33\linewidth]{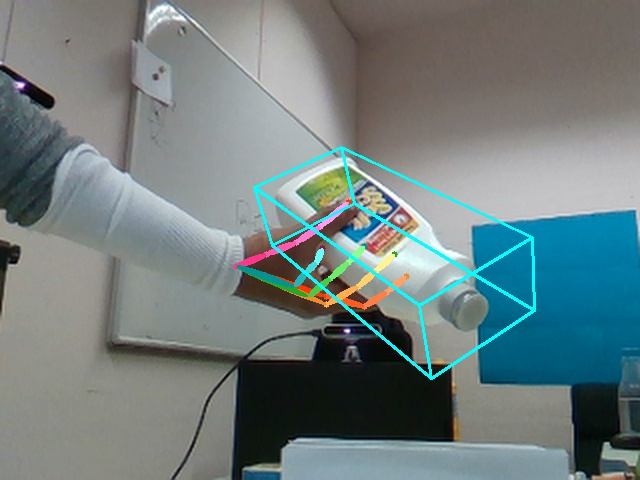} & \hspace{-3mm}
		\includegraphics[width=0.33\linewidth]{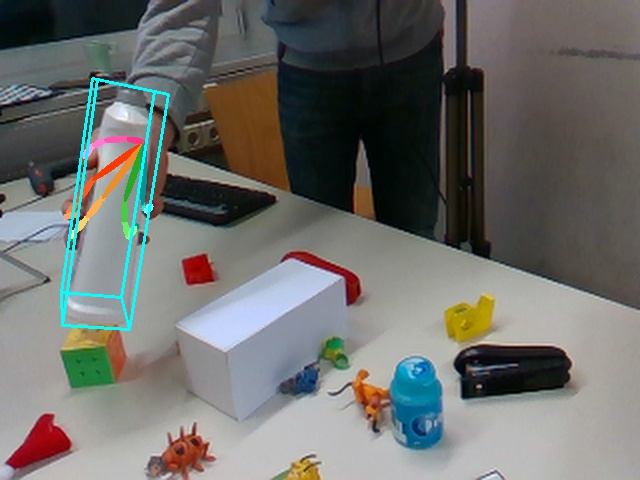} \\

		\includegraphics[width=0.33\linewidth]{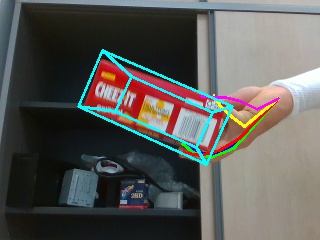} & \hspace{-3mm}
		\includegraphics[width=0.33\linewidth]{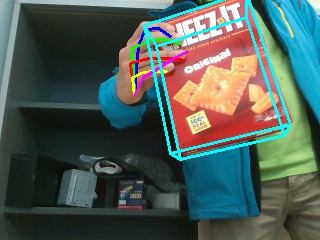} & \hspace{-3mm}
		\includegraphics[width=0.33\linewidth]{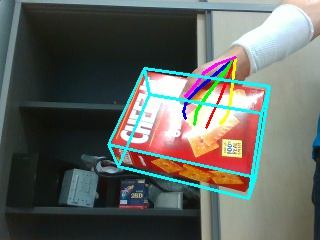} \\

		\includegraphics[width=0.33\linewidth]{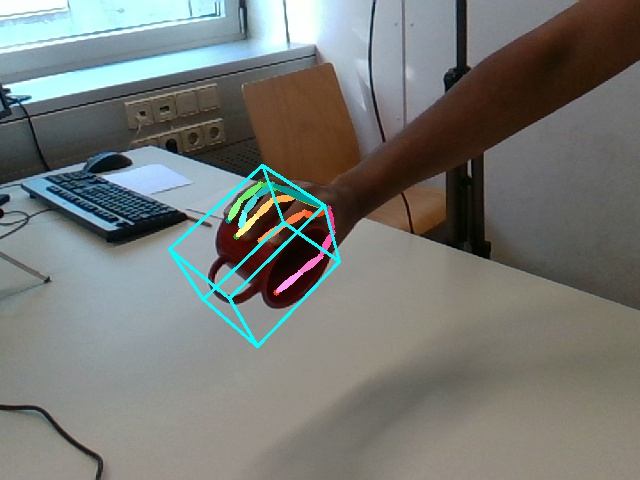} & \hspace{-3mm}
		\includegraphics[width=0.33\linewidth]{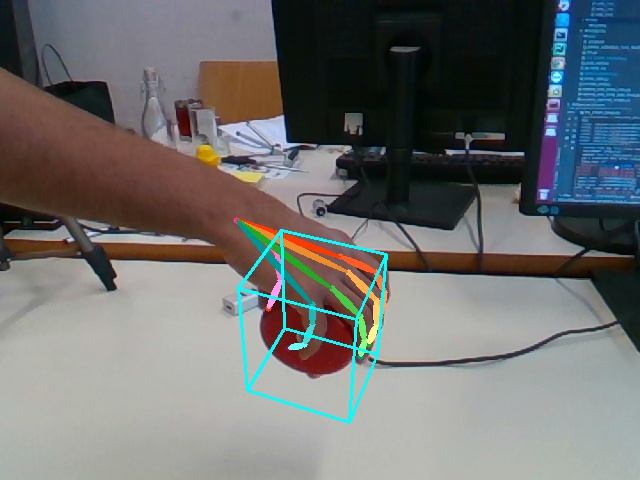} & \hspace{-3mm}
		\includegraphics[width=0.33\linewidth]{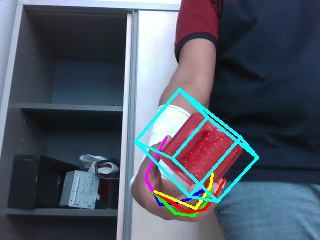} \\

		\includegraphics[width=0.33\linewidth]{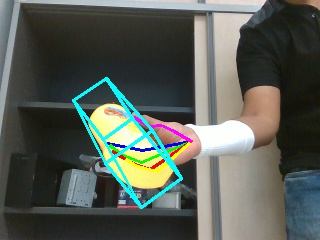} & \hspace{-3mm}
		\includegraphics[width=0.33\linewidth]{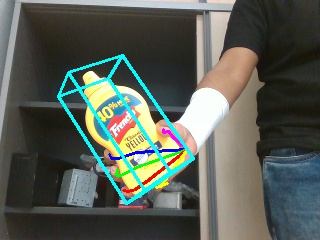} & \hspace{-3mm}
		\includegraphics[width=0.33\linewidth]{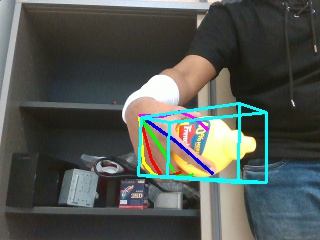} \\

	\end{tabular}
	\end{center}
	\vspace{-2mm}
	\caption{Some examples of the 3D annotated frames for both hand and object from our proposed dataset, HO-3D.}
	\label{fig:HO_3D_1}
\end{figure*}

\begin{figure*}
	\begin{center}
	\begin{tabular}{ccc}
		
		\includegraphics[width=0.33\linewidth]{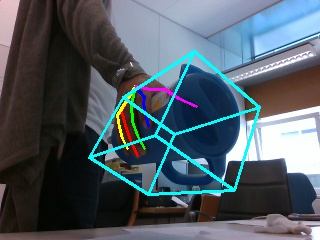} & \hspace{-3mm}
		\includegraphics[width=0.33\linewidth]{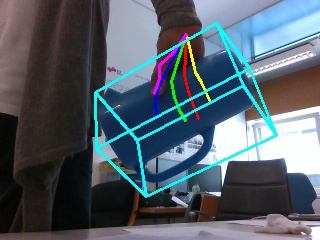} & \hspace{-3mm}
		\includegraphics[width=0.33\linewidth]{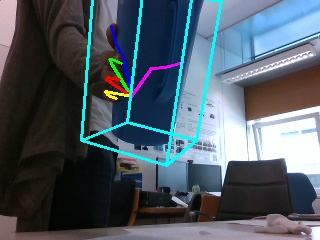} \\

		\includegraphics[width=0.33\linewidth]{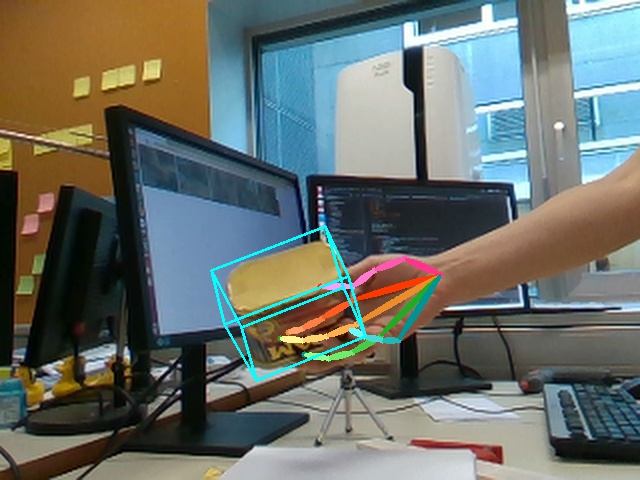} & \hspace{-3mm}
../		\includegraphics[width=0.33\linewidth]{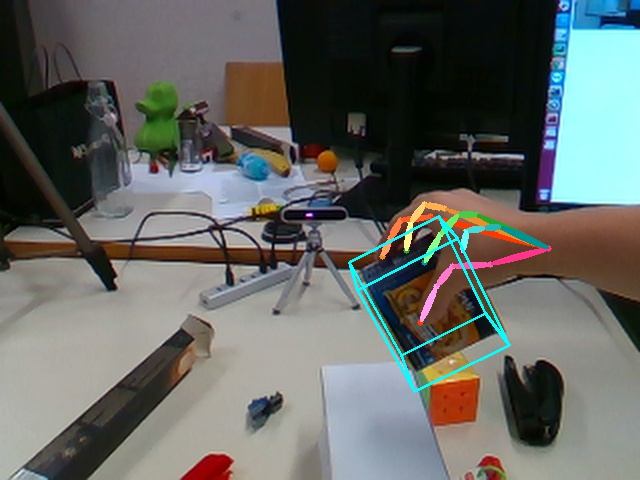} & \hspace{-3mm}
		\includegraphics[width=0.33\linewidth]{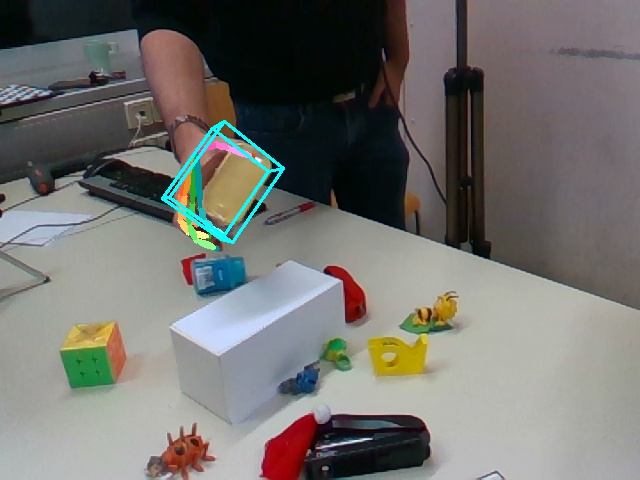} \\

		\includegraphics[width=0.33\linewidth]{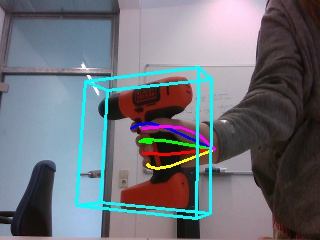} & \hspace{-3mm}
		\includegraphics[width=0.33\linewidth]{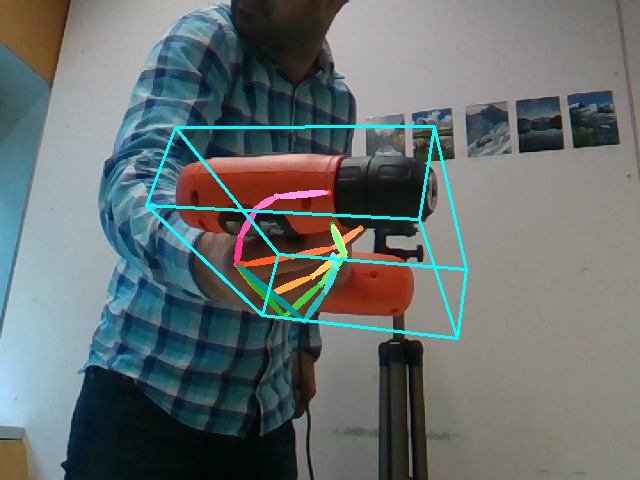} & \hspace{-3mm}
		\includegraphics[width=0.33\linewidth]{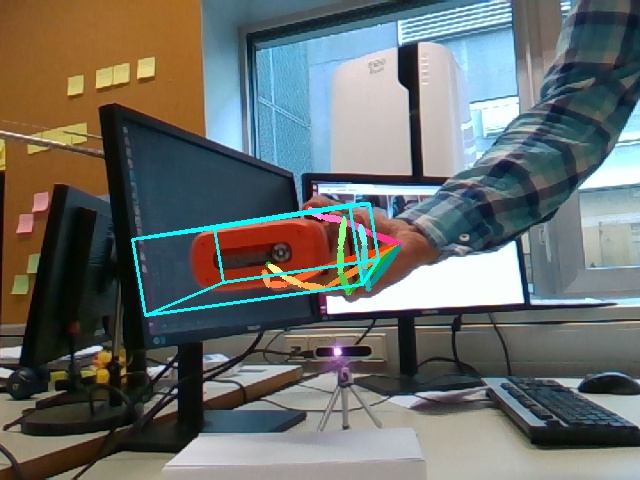} \\

		\includegraphics[width=0.33\linewidth]{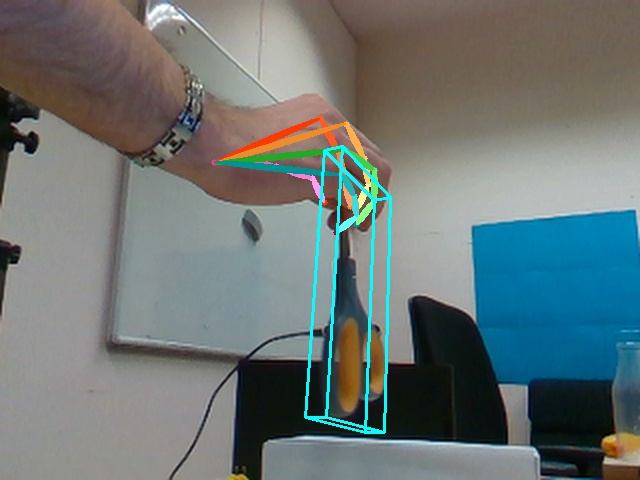} & \hspace{-3mm}
		\includegraphics[width=0.33\linewidth]{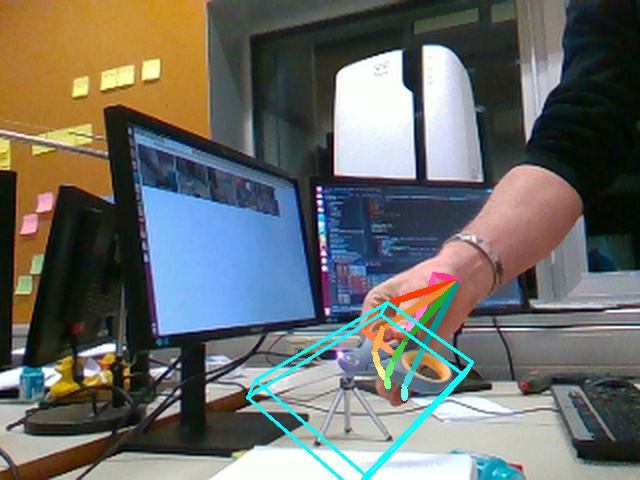} & \hspace{-3mm}
		\includegraphics[width=0.33\linewidth]{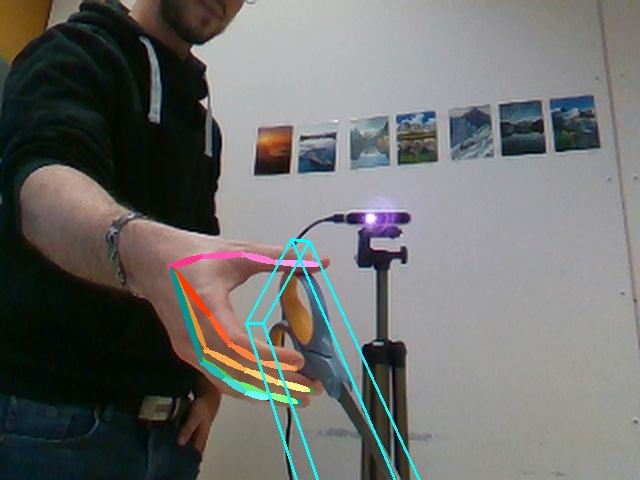} \\

		\includegraphics[width=0.33\linewidth]{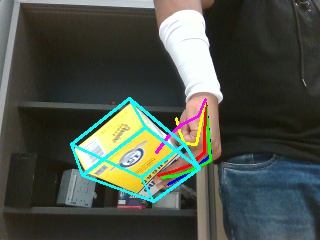} & \hspace{-3mm}
		\includegraphics[width=0.33\linewidth]{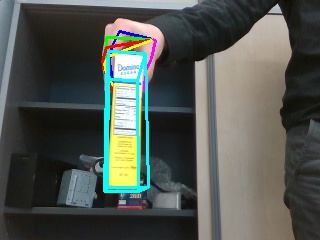} & \hspace{-3mm}
		\includegraphics[width=0.33\linewidth]{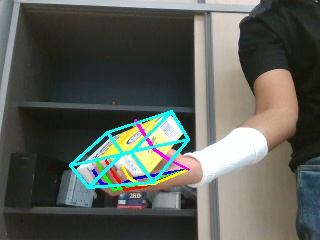} \\
	\end{tabular}
	\end{center}
	\vspace{-2mm}
	\caption{Some examples of the 3D annotated frames for both hand and object from our proposed dataset, HO-3D.}
	\label{fig:HO_3D_2}
\end{figure*}

\begin{figure*}[ht]
	\begin{center}
		\begin{tabular}{ccccc}
			\includegraphics[width=0.18\linewidth]{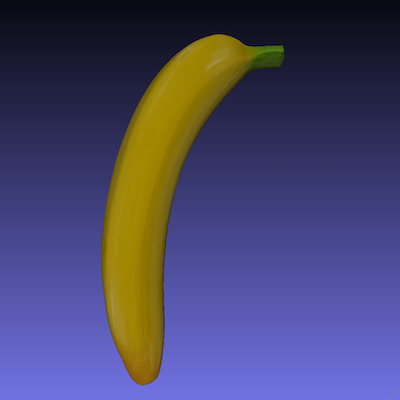} & 
			\includegraphics[width=0.18\linewidth]{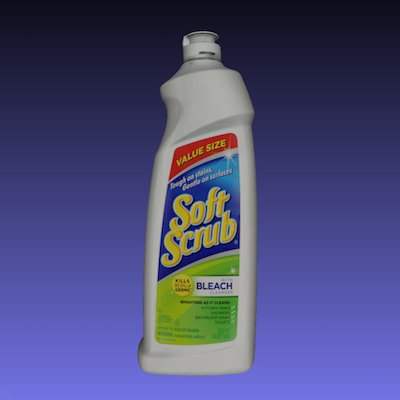} & 
			\includegraphics[width=0.18\linewidth]{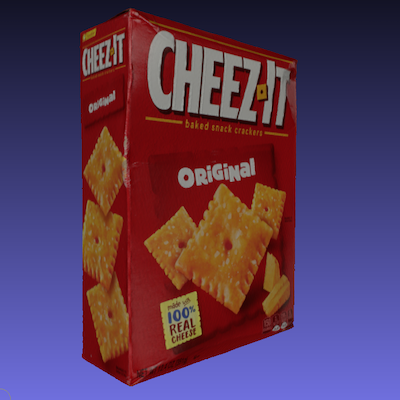} & 
			\includegraphics[width=0.18\linewidth]{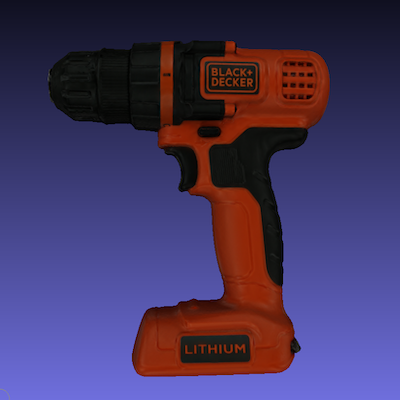} & 
			\includegraphics[width=0.18\linewidth]{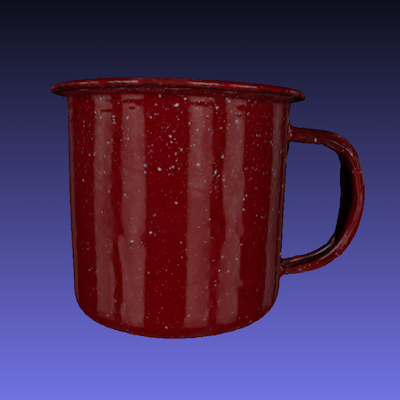} \\
			
			Banana & Bleach & Cracker Box & Driller & Mug \\
			
			\includegraphics[width=0.18\linewidth]{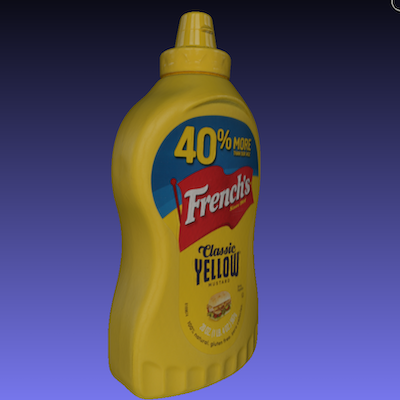} & 
			\includegraphics[width=0.18\linewidth]{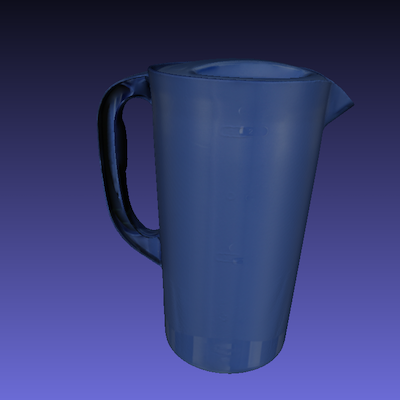} & 
			\includegraphics[width=0.18\linewidth]{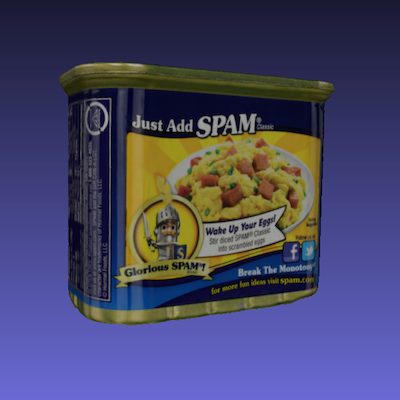} & 
			\includegraphics[width=0.18\linewidth]{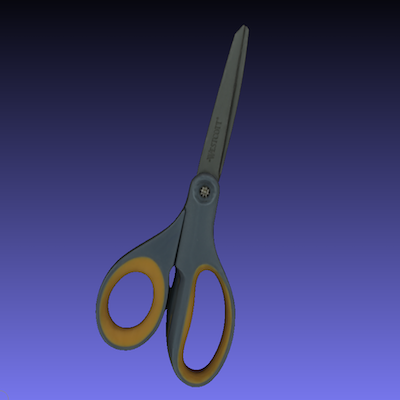} & 
			\includegraphics[width=0.18\linewidth]{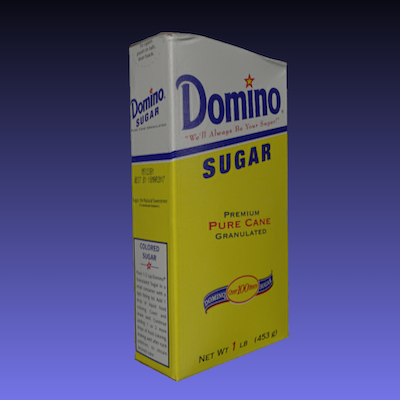} \\
			
			Mustard Bottle & Pitcher & Potted Meat & Scissors & Sugar Box \\
		\end{tabular}
	\end{center}
	\vspace{-2mm}
	\caption{ 10 objects of the YCB dataset~\cite{posecnn2018} that we use for our dataset HO-3D.}
	\label{fig:ycb_objects}
\end{figure*}

%{\small
%\bibliographystyle{ieee_fullname}
%\bibliography{visionSupp}
%}

{\small
\bibliographystyle{ieee_fullname}
\bibliography{vision}

\begin{thebibliography}{10}\itemsep=-1pt

\bibitem{ycbonline}
{YCB Benchmarks – Object and Model Set}.
\newblock http://ycbbenchmarks.org/.

\bibitem{Ballan2012motion}
Luca Ballan, Aparna Taneja, Jürgen Gall, Luc~Van Gool, and Marc Pollefeys.
\newblock Motion capture of hands in action using discriminative salient
  points.
\newblock In {\em European Conference on Computer Vision (ECCV)}, pages
  640--653, Firenze, October 2012.

\bibitem{bambach2015lending}
Sven Bambach, Stefan Lee, David~J. Crandall, and Chen Yu.
\newblock {Lending a Hand: Detecting Hands and Recognizing Activities in
  Complex Egocentric Interactions}.
\newblock In {\em The IEEE International Conference on Computer Vision (ICCV)},
  pages 1949--1957, 2015.

\bibitem{boukhayma20193d}
Adnane Boukhayma, Rodrigo~de Bem, and Philip~HS Torr.
\newblock 3d hand shape and pose from images in the wild.
\newblock In {\em The {IEEE} Conference on Computer Vision and Pattern
  Recognition (CVPR)}, pages 10843--10852, 2019.

\bibitem{Buch2017}
{Anders Glent} Buch, Lilita Kiforenko, and Dirk Kraft.
\newblock Rotational subgroup voting and pose clustering for robust 3d object
  recognition.
\newblock In {\em The IEEE International Conference On Computer Vision (ICCV)},
  pages 4137--4145, United States, 2017.

\bibitem{bullock2015yale}
Ian~M. Bullock, Thomas Feix, and Aaron~M. Dollar.
\newblock {The Yale Human Grasping Dataset: Grasp, Object, and Task Data in
  Household and Machine Shop Environments}.
\newblock {\em The International Journal of Robotics Research}, 34(3):251--255,
  2015.

\bibitem{cai2015scalable}
Minjie Cai, Kris~M. Kitani, and Yoichi Sato.
\newblock {A Scalable Approach for Understanding the Visual Structures of Hand
  Grasps}.
\newblock In {\em The IEEE International Conference on Robotics and Automation
  (ICRA)}, 2015.

\bibitem{Zhang2017}
Qixin Cao and Haoruo Zhang.
\newblock {Combined Holistic and Local Patches for Recovering 6D Object Pose}.
\newblock In {\em The IEEE International Conference on Computer Vision (ICCV)},
  pages 2219--2227, 2017.

\bibitem{Cao2017realtime}
Z. Cao, T. Simon, S.-E. Wei, and Y. Sheikh.
\newblock {Realtime Multi-Person 2D Pose Estimation Using Part Affinity
  Fields}.
\newblock 2017.

\bibitem{deeplabv2017}
Liang{-}Chieh Chen, George Papandreou, Florian Schroff, and Hartwig Adam.
\newblock {Rethinking Atrous Convolution for Semantic Image Segmentation}.
\newblock {\em CoRR}, abs/1706.05587, 2017.

\bibitem{choi2017robust}
Chiho Choi, Sang~Ho Yoon, Chin{-}Ning Chen, and Karthik Ramani.
\newblock {Robust Hand Pose Estimation During the Interaction with an Unknown
  Object}.
\newblock In {\em The {IEEE} International Conference on Computer Vision
  (ICCV)}, pages 3142--3151, 2017.

\bibitem{ConnGoulToin00}
Andrew~R. Conn, Nicholas I.~M. Gould, and {\relax Ph}ilippe~L. Toint.
\newblock {\em {Trust-Region Methods}}.
\newblock SIAM, 2000.

\bibitem{de2011model}
Martin de La~Gorce, David~J. Fleet, and Nikos Paragios.
\newblock {Model-Based 3D Hand Pose Estimation from Monocular Video}.
\newblock {\em {IEEE} Trans. Pattern Anal. Mach. Intell.(PAMI)},
  33(9):1793--1805, 2011.

\bibitem{fathi2011learning}
Alireza Fathi, Xiaofeng Ren, and James~M. Rehg.
\newblock {Learning to Recognize Objects in Egocentric Activities}.
\newblock In {\em The {IEEE} Conference on Computer Vision and Pattern
  Recognition (CVPR)}, pages 3281--3288, 2011.

\bibitem{garcia2018first}
Guillermo Garcia{-}Hernando, Shanxin Yuan, Seungryul Baek, and Tae{-}Kyun Kim.
\newblock {First-Person Hand Action Benchmark with RGB-D Videos and 3D Hand
  Pose Annotations}.
\newblock In {\em The {IEEE} Conference on Computer Vision and Pattern
  Recognition (CVPR)}, pages 409--419, 2018.

\bibitem{ge2018hand}
Liuhao Ge, Yujun Cai, Junwu Weng, and Junsong Yuan.
\newblock {Hand {PointNet}: 3D Hand Pose Estimation Using Point Sets}.
\newblock In {\em The {IEEE} Conference on Computer Vision and Pattern
  Recognition (CVPR)}, pages 8417--8426, 2018.

\bibitem{goudie20173d}
Duncan Goudie and Aphrodite Galata.
\newblock {3D Hand-Object Pose Estimation from Depth with Convolutional Neural
  Networks}.
\newblock In {\em IEEE International Conference on Automatic Face \& Gesture
  Recognition}, 2017.

\bibitem{ho3d_old}
Shreyas Hampali, Markus Oberweger, Mahdi Rad, and Vincent Lepetit.
\newblock {{HO-3D:} a Multi-User, Multi-Object Dataset for Joint 3D Hand-Object
  Pose Estimation}.
\newblock {\em CoRR}, abs/1907.01481, 2019.

\bibitem{hasson2019learning}
Yana Hasson, G{\"{u}}l Varol, Dimitrios Tzionas, Igor Kalevatykh, Michael~J.
  Black, Ivan Laptev, and Cordelia Schmid.
\newblock {Learning Joint Reconstruction of Hands and Manipulated Objects}.
\newblock In {\em The {IEEE} Conference on Computer Vision and Pattern
  Recognition (CVPR)}, pages 11807--11816, 2019.

\bibitem{Henderson2019}
Paul Henderson and Vittorio Ferrari.
\newblock {Learning Single-Image 3D Reconstruction by Generative Modelling of
  Shape, Pose and Shading}.
\newblock {\em International Journal of Computer Vision (IJCV)}, pages
  1573--1405, 2019.

\bibitem{hu2019segpose}
Yinlin Hu, Joachim Hugonot, Pascal Fua, and Mathieu Salzmann.
\newblock {Segmentation-Driven 6d Object Pose Estimation}.
\newblock In {\em The {IEEE} Conference on Computer Vision and Pattern
  Recognition (CVPR)}, pages 3385--3394, 2019.

\bibitem{JooSS18}
Hanbyul Joo, Tomas Simon, and Yaser Sheikh.
\newblock {Total Capture: A 3D Deformation Model for Tracking Faces, Hands, and
  Bodies}.
\newblock In {\em The {IEEE} Conference on Computer Vision and Pattern
  Recognition (CVPR)}, pages 8320--8329, 2018.

\bibitem{Kehl16}
Wadim Kehl, Fausto Milletari, Federico Tombari, Slobodan Ilic, and Nassir
  Navab.
\newblock {Deep Learning of Local RGB-D Patches for 3D Object Detection and 6D
  Pose Estimation}.
\newblock In {\em European Conference on Computer Vision (ECCV)}, pages
  205--220, 2016.

\bibitem{keskin2012hand}
Cem Keskin, Furkan Kira{\c{c}}, Yunus~Emre Kara, and Lale Akarun.
\newblock {Hand Pose Estimation and Hand Shape Classification Using
  Multi-Layered Randomized Decision Forests}.
\newblock In {\em European Conference on Computer Vision (ECCV)}, pages
  852--863, 2012.

\bibitem{Kingma15}
Diederik~P. Kingma and Jimmy Ba.
\newblock {Adam: A Method for Stochastic Optimization}.
\newblock In {\em The International Conference on Learning Representations
  (ICLR)}, 2015.

\bibitem{kokic2019learning}
Mia Kokic, Danica Kragic, and Jeannette Bohg.
\newblock {Learning to Estimate Pose and Shape of Hand-Held Objects from RGB
  Images}.
\newblock In {\em The {IEEE/RSJ} International Conference on Intelligent Robots
  and Systems (IROS)}, pages 3980--3987, 11 2019.

\bibitem{kyriazis2013physically}
Nikolaos Kyriazis and Antonis~A. Argyros.
\newblock {Physically Plausible 3D Scene Tracking: The Single Actor
  Hypothesis}.
\newblock In {\em The {IEEE} Conference on Computer Vision and Pattern
  Recognition (CVPR)}, 2013.

\bibitem{Kyriazis14}
Nikolaos Kyriazis and Antonis~A. Argyros.
\newblock {Scalable 3D Tracking of Multiple Interacting Objects}.
\newblock In {\em The {IEEE} Conference on Computer Vision and Pattern
  Recognition (CVPR)}, 2014.

\bibitem{melax2013dynamics}
Stan Melax, Leonid Keselman, and Sterling Orsten.
\newblock {Dynamics Based 3D Skeletal Hand Tracking}.
\newblock In {\em Proceedings of Graphics Interface}, 2013.

\bibitem{Miller04}
Andrew~T. Miller and Peter~K. Allen.
\newblock {Graspit! a Versatile Simulator for Robotic Grasping}.
\newblock {\em {Robotics Automation Magazine}}, pages 110--122, 2004.

\bibitem{Mitash2017}
Chaitanya Mitash, Abdeslam Boularias, and Kostas~E. Bekris.
\newblock {Improving 6D Pose Estimation of Objects in Clutter via Physics-Aware
  Monte Carlo Tree Search}.
\newblock In {\em The {IEEE} International Conference on Robotics and
  Automation (ICRA)}, pages 1--8, 2018.

\bibitem{Mueller2018}
Franziska Mueller, Florian Bernard, Oleksandr Sotnychenko, Dushyant Mehta,
  Srinath Sridhar, Dan Casas, and Christian Theobalt.
\newblock {Ganerated Hands for Real-Time 3D Hand Tracking from Monocular RGB}.
\newblock In {\em The {IEEE} Conference on Computer Vision and Pattern
  Recognition (CVPR)}, pages 49--59, 2018.

\bibitem{mueller2017real}
Franziska Mueller, Dushyant Mehta, Oleksandr Sotnychenko, Srinath Sridhar, Dan
  Casas, and Christian Theobalt.
\newblock {Real-Time Hand Tracking Under Occlusion from an Egocentric Rgb-D
  Sensor}.
\newblock In {\em The IEEE International Conference on Computer Vision (ICCV)},
  2017.

\bibitem{Myanganbayar18}
Battushig Myanganbayar, Cristina Mata, Gil Dekel, Boris Katz, Guy Ben{-}Yosef,
  and Andrei Barbu.
\newblock {Partially Occluded Hands: A Challenging New Dataset for Single-Image
  Hand Pose Estimation}.
\newblock In {\em The Asian Conference on Computer Vision (ACCV)}, pages
  85--98, 2018.

\bibitem{Neverova17}
Natalia Neverova, Christian Wolf, Florian Nebout, and Graham~W. Taylor.
\newblock {Hand Pose Estimation through Semi-Supervised and Weakly-Supervised
  Learning}.
\newblock {\em Comput. Vis. Image Underst. (CVIU)}, 164:56--67, 2017.

\bibitem{Oberweger17}
Markus Oberweger and Vincent Lepetit.
\newblock {{DeepPrior++}: Improving Fast and Accurate 3D Hand Pose Estimation}.
\newblock In {\em The {IEEE} International Conference on Computer Vision
  Workshops (ICCV)}, pages 585--594, 2017.

\bibitem{Oberweger2018}
Markus Oberweger, Mahdi Rad, and Vincent Lepetit.
\newblock {Making Deep Heatmaps Robust to Partial Occlusions for 3D Object Pose
  Estimation}.
\newblock In {\em European Conference on Computer Vision (ECCV)}, pages
  125--141, 2018.

\bibitem{Oberweger16}
Markus Oberweger, Gernot Riegler, Paul Wohlhart, and Vincent Lepetit.
\newblock {Efficiently Creating 3D Training Data for Fine Hand Pose
  Estimation}.
\newblock In {\em The {IEEE} Conference on Computer Vision and Pattern
  Recognition (CVPR)}, pages 4957--4965, 2016.

\bibitem{Oikonomidis2011full}
Iasonas Oikonomidis, Nikolaos Kyriazis, and Antonis~A. Argyros.
\newblock {Full {DoF} Tracking of a Hand Interacting with an Object by Modeling
  Occlusions and Physical Constraints}.
\newblock In {\em The IEEE International Conference on Computer Vision (ICCV)},
  pages 2088--2095, 2011.

\bibitem{panteleris20153d}
Paschalis Panteleris, Nikolaos Kyriazis, and Antonis~A. Argyros.
\newblock {3D Tracking of Human Hands in Interaction with Unknown Objects}.
\newblock In {\em Proceedings of the British Machine Vision Conference 2015
  (BMVC)}, pages 123.1--123.12, 2015.

\bibitem{Panteleris2018}
Paschalis Panteleris, Iason Oikonomidis, and Antonis~A. Argyros.
\newblock {Using a Single RGB Frame for Real Time 3D Hand Pose Estimation in
  the Wild}.
\newblock In {\em The {IEEE} Winter Conference on Applications of Computer
  Vision (WACV)}, pages 436--445, 2018.

\bibitem{peng2019pvnet}
Sida Peng, Yuan Liu, Qixing Huang, Xiaowei Zhou, and Hujun Bao.
\newblock {Pvnet: Pixel-Wise Voting Network for 6DoF Pose Estimation}.
\newblock In {\em The {IEEE} Conference on Computer Vision and Pattern
  Recognition (CVPR)}, pages 4561--4570, 2019.

\bibitem{pham2018hand}
Tu{-}Hoa Pham, Nikolaos Kyriazis, Antonis~A. Argyros, and Abderrahmane Kheddar.
\newblock {Hand-Object Contact Force Estimation from Markerless Visual
  Tracking}.
\newblock {\em {IEEE} Trans. Pattern Anal. Mach. Intell.(PAMI)},
  40(12):2883--2896, 2018.

\bibitem{pham2015capturing}
Tu-Hoa Pham, Abderrahmane Kheddar, Ammar Qammaz, and Antonis~A. Argyros.
\newblock {Capturing and Reproducing Hand-Object Interactions through
  Vision-Based Force Sensing}.
\newblock In {\em Object Understanding for Interaction}, 2015.

\bibitem{Pham_2015_CVPR}
Tu-Hoa Pham, Abderrahmane Kheddar, Ammar Qammaz, and Antonis~A. Argyros.
\newblock {Towards Force Sensing from Vision: Observing Hand-Object
  Interactions to Infer Manipulation Forces}.
\newblock In {\em The {IEEE} Conference on Computer Vision and Pattern
  Recognition (CVPR)}, 2015.

\bibitem{Qian2014realtime}
Chen Qian, Xiao Sun, Yichen Wei, Xiaoou Tang, and Jian Sun.
\newblock {Realtime and Robust Hand Tracking from Depth}.
\newblock In {\em The {IEEE} Conference on Computer Vision and Pattern
  Recognition (CVPR)}, 2014.

\bibitem{Rad17}
Mahdi Rad and Vincent Lepetit.
\newblock {{BB8}: A Scalable, Accurate, Robust to Partial Occlusion Method for
  Predicting the 3D Poses of Challenging Objects Without Using Depth}.
\newblock In {\em The IEEE International Conference on Computer Vision (ICCV)},
  2017.

\bibitem{Rad2018b}
Mahdi Rad, Markus Oberweger, and Vincent Lepetit.
\newblock {Domain Transfer for 3D Pose Estimation from Color Images Without
  Manual Annotations}.
\newblock In {\em The Asian Conference on Computer Vision (ACCV)}, 2018.

\bibitem{rogez2015understanding}
Gr{\'{e}}gory Rogez, James Steven~Supancic III, and Deva Ramanan.
\newblock {Understanding Everyday Hands in Action from RGB-D Images}.
\newblock In {\em The IEEE International Conference on Computer Vision (ICCV)},
  2015.

\bibitem{rogez20143d}
Gr{\'{e}}gory Rogez, Maryam Khademi, James Steven~Supancic III, J.~M.~M.
  Montiel, and Deva Ramanan.
\newblock {3D Hand Pose Detection in Egocentric RGB-D Images}.
\newblock In {\em European Conference on Computer Vision (ECCV)}, 2014.

\bibitem{romero2017embodied}
Javier Romero, Dimitrios Tzionas, and Michael~J. Black.
\newblock {Embodied Hands: Modeling and Capturing Hands and Bodies Together}.
\newblock {\em {ACM} Trans. Graph.}, 36(6):245, 2017.

\bibitem{sharp2015accurate}
Toby Sharp, Cem Keskin, Duncan~P. Robertson, Jonathan Taylor, Jamie Shotton,
  David Kim, Christoph Rhemann, Ido Leichter, Alon Vinnikov, Yichen Wei, Daniel
  Freedman, Pushmeet Kohli, Eyal Krupka, Andrew~W. Fitzgibbon, and Shahram
  Izadi.
\newblock {Accurate, Robust, and Flexible Real-Time Hand Tracking}.
\newblock In {\em Proceedings of the 33rd Annual {ACM} Conference on Human
  Factors in Computing Systems, {CHI}}, pages 3633--3642, 2015.

\bibitem{Simonyan14c}
K. Simonyan and A. Zisserman.
\newblock {Very Deep Convolutional Networks for Large-Scale Image Recognition}.
\newblock {\em CoRR}, abs/1409.1556, 2014.

\bibitem{RealtimeHO_ECCV2016}
Srinath Sridhar, Franziska Mueller, Michael Zollh{\"{o}}fer, Dan Casas, Antti
  Oulasvirta, and Christian Theobalt.
\newblock {Real-Time Joint Tracking of a Hand Manipulating an Object from RGB-D
  Input}.
\newblock In {\em European Conference on Computer Vision (ECCV)}, volume 9906,
  pages 294--310, 2016.

\bibitem{sridhar2013interactive}
Srinath Sridhar, Antti Oulasvirta, and Christian Theobalt.
\newblock {Interactive Markerless Articulated Hand Motion Tracking Using RGB
  and Depth Data}.
\newblock In {\em The IEEE International Conference on Computer Vision (ICCV)},
  pages 2456--2463, 2013.

\bibitem{Sun2015}
Xiao Sun, Yichen Wei, Shuang Liang, Xiaoou Tang, and Jian Sun.
\newblock {Cascaded Hand Pose Regression}.
\newblock In {\em The {IEEE} Conference on Computer Vision and Pattern
  Recognition (CVPR)}, pages 824--832, 2015.

\bibitem{sundermeyer2018implicit}
Martin Sundermeyer, Zoltan{-}Csaba Marton, Maximilian Durner, Manuel Brucker,
  and Rudolph Triebel.
\newblock {Implicit 3D Orientation Learning for 6D Object Detection from RGB
  Images}.
\newblock In {\em European Conference on Computer Vision (ECCV)}, pages
  712--729, 2018.

\bibitem{Tan_2016_CVPR}
David~Joseph Tan, Thomas~J. Cashman, Jonathan Taylor, Andrew~W. Fitzgibbon,
  Daniel Tarlow, Sameh Khamis, Shahram Izadi, and Jamie Shotton.
\newblock {Fits Like a Glove: Rapid and Reliable Hand Shape Personalization}.
\newblock In {\em The {IEEE} Conference on Computer Vision and Pattern
  Recognition (CVPR)}, pages 5610--5619, June 2016.

\bibitem{Tang2014}
Danhang Tang, Hyung~Jin Chang, Alykhan Tejani, and Tae{-}Kyun Kim.
\newblock {Latent Regression Forest: Structured Estimation of 3D Articulated
  Hand Posture}.
\newblock In {\em The {IEEE} Conference on Computer Vision and Pattern
  Recognition (CVPR)}, pages 3786--3793, 2014.

\bibitem{tekin2019h}
Bugra Tekin, Federica Bogo, and Marc Pollefeys.
\newblock {{H+O}: Unified Egocentric Recognition of 3D Hand-Object Poses and
  Interactions}.
\newblock In {\em The {IEEE} Conference on Computer Vision and Pattern
  Recognition (CVPR)}, pages 4511--4520, 2019.

\bibitem{Tompson14b}
Jonathan Tompson, Murphy Stein, Yann LeCun, and Ken Perlin.
\newblock {Real-Time Continuous Pose Recovery of Human Hands Using
  Convolutional Networks}.
\newblock {\em {ACM} Trans. Graph.}, 33:169:1--169:10, 2014.

\bibitem{Tsoli_2018_ECCV}
Aggeliki Tsoli and Antonis~A. Argyros.
\newblock {Joint 3D Tracking of a Deformable Object in Interaction with a
  Hand}.
\newblock In {\em European Conference on Computer Vision (ECCV)}, pages
  504--520, 2018.

\bibitem{Tzionas2016}
Dimitrios Tzionas, Luca Ballan, Abhilash Srikantha, Pablo Aponte, Marc
  Pollefeys, and Juergen Gall.
\newblock {Capturing Hands in Action Using Discriminative Salient Points and
  Physics Simulation}.
\newblock {\em International Journal of Computer Vision (IJCV)},
  118(2):172--193, 2016.

\bibitem{Tzionas_2015_ICCV}
Dimitrios Tzionas and Juergen Gall.
\newblock {3D Object Reconstruction from Hand-Object Interactions}.
\newblock In {\em The IEEE International Conference on Computer Vision (ICCV)},
  pages 729--737, 2015.

\bibitem{Tzionas2014capturing}
Dimitrios Tzionas, Abhilash Srikantha, Pablo Aponte, and Juergen Gall.
\newblock {Capturing Hand Motion with an RGB-D Sensor, Fusing a Generative
  Model with Salient Points}.
\newblock In {\em German Conference on Pattern Recognition (GCPR)}, pages
  277--289, 2014.

\bibitem{Wan_2019_CVPR}
Chengde Wan, Thomas Probst, Luc~Van Gool, and Angela Yao.
\newblock {Self-Supervised 3D Hand Pose Estimation through Training by
  Fitting}.
\newblock In {\em The {IEEE} Conference on Computer Vision and Pattern
  Recognition (CVPR)}, pages 10853--10862, June 2019.

\bibitem{Wang11d}
Robert~Y. Wang, Sylvain Paris, and Jovan Popovic.
\newblock {6D Hands: Markerless Hand-Tracking for Computer Aided Design}.
\newblock In {\em ACM Symposium on User Interface Software and Technology},
  pages 549--558, 2011.

\bibitem{Wei16}
Shih{-}En Wei, Varun Ramakrishna, Takeo Kanade, and Yaser Sheikh.
\newblock {Convolutional Pose Machines}.
\newblock In {\em The {IEEE} Conference on Computer Vision and Pattern
  Recognition (CVPR)}, pages 4724--4732, 2016.

\bibitem{xiang2019monocular}
Donglai Xiang, Hanbyul Joo, and Yaser Sheikh.
\newblock {Monocular Total Capture: Posing Face, Body, and Hands in the Wild}.
\newblock In {\em The {IEEE} Conference on Computer Vision and Pattern
  Recognition (CVPR)}, pages 10965--10974, 2019.

\bibitem{posecnn2018}
Yu Xiang, Tanner Schmidt, Venkatraman Narayanan, and Dieter Fox.
\newblock {{PoseCNN}: A Convolutional Neural Network for 6D Object Pose
  Estimation in Cluttered Scenes}.
\newblock In {\em Robotics: Science and Systems XIV (RSS)}, 2018.

\bibitem{Xu16}
Chi Xu, Lakshmi~Narasimhan Govindarajan, Yu Zhang, and Li Cheng.
\newblock {Lie-X: Depth Image Based Articulated Object Pose Estimation,
  Tracking, and Action Recognition on Lie Groups}.
\newblock {\em International Journal of Computer Vision (IJCV)}, 123:454--478,
  2017.

\bibitem{ye2016spatial}
Qi Ye, Shanxin Yuan, and Tae{-}Kyun Kim.
\newblock {Spatial Attention Deep Net with Partial {PSO} for Hierarchical
  Hybrid Hand Pose Estimation}.
\newblock In {\em European Conference on Computer Vision (ECCV)}, pages
  346--361, 2016.

\bibitem{Yuan2017}
Shanxin Yuan, Qi Ye, Bjorn Stenger, Siddhant Jain, and Tae{-}Kyun Kim.
\newblock {Big Hand 2.2m Benchmark: Hand Pose Data Set and State of the Art
  Analysis}.
\newblock In {\em The {IEEE} Conference on Computer Vision and Pattern
  Recognition (CVPR)}, 2017.

\bibitem{Zhou16c}
Xingyi Zhou, Qingfu Wan, Wei Zhang, Xiangyang Xue, and Yichen Wei.
\newblock Model-based deep hand pose estimation.
\newblock In {\em The International Joint Conference on Artificial Intelligence
  (IJCAI)}, pages 2421--2427, 2016.

\bibitem{Zimmermann2017}
Christian Zimmermann and Thomas Brox.
\newblock {Learning to Estimate 3D Hand Pose from Single RGB Images}.
\newblock In {\em The IEEE International Conference on Computer Vision (ICCV)},
  pages 4913--4921, 2017.

\bibitem{zimmermann2019freihand}
Christian Zimmermann, Duygu Ceylan, Jimei Yang, Bryan~C. Russell, Max~J. Argus,
  and Thomas Brox.
\newblock {FreiHAND: A Dataset for Markerless Capture of Hand Pose and Shape
  from Single RGB Images}.
\newblock In {\em The IEEE International Conference on Computer Vision (ICCV)},
  pages 813--822, 2019.

\end{thebibliography}
}

\end{document}